\documentclass{article}

\usepackage[nonatbib, final]{neurips_data_2022}
\usepackage[numbers]{natbib}

\usepackage[utf8]{inputenc} % allow utf-8 input
\usepackage[T1]{fontenc}    % use 8-bit T1 fonts
\usepackage{hyperref}       % hyperlinks
\usepackage{url}            % simple URL typesetting
\usepackage{booktabs}       % professional-quality tables
\usepackage{amsfonts}       % blackboard math symbols
\usepackage{nicefrac}       % compact symbols for 1/2, etc.
\usepackage{microtype}      % microtypography
\usepackage{xcolor}         % colors
\usepackage{amsmath,amssymb}
\usepackage{amsthm,latexsym,paralist}
\usepackage{wrapfig}
\usepackage{graphicx}
\usepackage{subcaption}
\usepackage{multirow}
\usepackage{bm}
\usepackage{floatrow}
\usepackage{xspace}
\usepackage{soul}
\usepackage{listings}
\usepackage[finalizecache=false, frozencache=true, cachedir=mintedcache]{minted}
\usepackage{tcolorbox}
\usepackage{etoolbox}

\definecolor{codegreen}{rgb}{0,0.6,0}
\definecolor{codegray}{rgb}{0.5,0.5,0.5}
\definecolor{codeblue}{rgb}{0.2,0,0.7}
\definecolor{codepurple}{rgb}{0.58,0,0.82}
\definecolor{backcolor}{rgb}{0.98,0.98,0.98}

\lstdefinestyle{mystyle}{
    backgroundcolor=\color{backcolor},   
    commentstyle=\color{codegray},
    keywordstyle=\color{orange},
    numberstyle=\tiny\color{codeblue},
    stringstyle=\color{codegreen},
    basicstyle=\ttfamily\footnotesize,
    breakatwhitespace=false,         
    breaklines=true,                 
    captionpos=b,                    
    keepspaces=true,                 
    numbers=left,                    
    numbersep=5pt,                  
    showspaces=false,                
    showstringspaces=false,
    showtabs=false,                  
    tabsize=2
}

\makeatletter
\DeclareRobustCommand\onedot{\futurelet\@let@token\@onedot}
\def\@onedot{\ifx\@let@token.\else.\null\fi\xspace}

\def\eg{\emph{e.g}\onedot} 
\def\ie{\emph{i.e}\onedot}

\newcommand{\printfnsymbol}[1]{%
  \textsuperscript{\@fnsymbol{#1}}%
}
\makeatother

\title{GOOD: A Graph Out-of-Distribution Benchmark}

\author{%
  Shurui Gui\thanks{Equal contributions},\quad Xiner Li\printfnsymbol{1},\quad Limei Wang,\quad Shuiwang Ji\\
  Texas A\&M University\\
  College Station, TX 77843 \\
  \texttt{\{shurui.gui,lxe,limei,sji\}@tamu.edu} \\
  \AND
}

\begin{document}

\maketitle

\vspace{-1cm}

\begin{abstract}
Out-of-distribution (OOD) learning deals with scenarios in which training and test data follow different distributions. Although general OOD problems have been intensively studied in machine learning, graph OOD is only an emerging area of research. Currently, there lacks a systematic benchmark tailored to graph OOD method evaluation. In this work, we aim at developing an OOD benchmark, known as GOOD, for graphs specifically. We explicitly make distinctions between covariate and concept shifts and design data splits that accurately reflect different shifts. We consider both graph and node prediction tasks as there are key differences in designing shifts. Overall, GOOD contains 11 datasets with 17 domain selections. When combined with covariate, concept, and no shifts, we obtain 51 different splits. We provide performance results on 10 commonly used baseline methods with 10 random runs. This results in 510 dataset-model combinations in total. Our results show significant performance gaps between in-distribution and OOD settings. Our results also shed light on different performance trends between covariate and concept shifts by different methods. Our GOOD benchmark is a growing project and expects to expand in both quantity and variety of resources as the area develops. The GOOD benchmark can be accessed via \url{https://github.com/divelab/GOOD/}.
\end{abstract}

\section{Introduction} \label{sec:introduction}

In machine learning, training and test data are commonly assumed to be i.i.d.. Models designed under this assumption may not perform well when the i.i.d. assumption does not hold. The area of out-of-distribution (OOD) learning deals with scenarios in which training and test data follow different distributions. Two commonly studied OOD settings are
covariate shift and concept shift. 
Over the years, multiple OOD methods have been proposed~\cite{ganin2016domain,sun2016deep,arjovsky2019invariant,sagawa2019distributionally,krueger2021out}. To facilitate evaluations, several benchmarks have been curated, including DomainBed~\cite{gulrajani2020search}, OoD-Bench~\cite{ye2021ood}, and WILDS~\cite{koh2021wilds}.
Although both general OOD problems and graph analysis~\cite{kipf2017semi,gao2019graph,velickovic2018graph,xu2018how,Liu:NC-GNN}
have been intensively studied, graph OOD is only an emerging area of research~\cite{wu2022dir, wu2022handling, zhu2021shift, bevilacqua2021size}.
Some initial attempts have been made to curate graph OOD benchmarks~\cite{ji2022drugood, ding2021closer}. However, existing benchmarks lack in several aspects, as detailed in Section~\ref{sec:related}.

\begin{figure}[!t]
    \centering
    % \vspace{-2cm}
    \scalebox{0.296}{
        \includegraphics{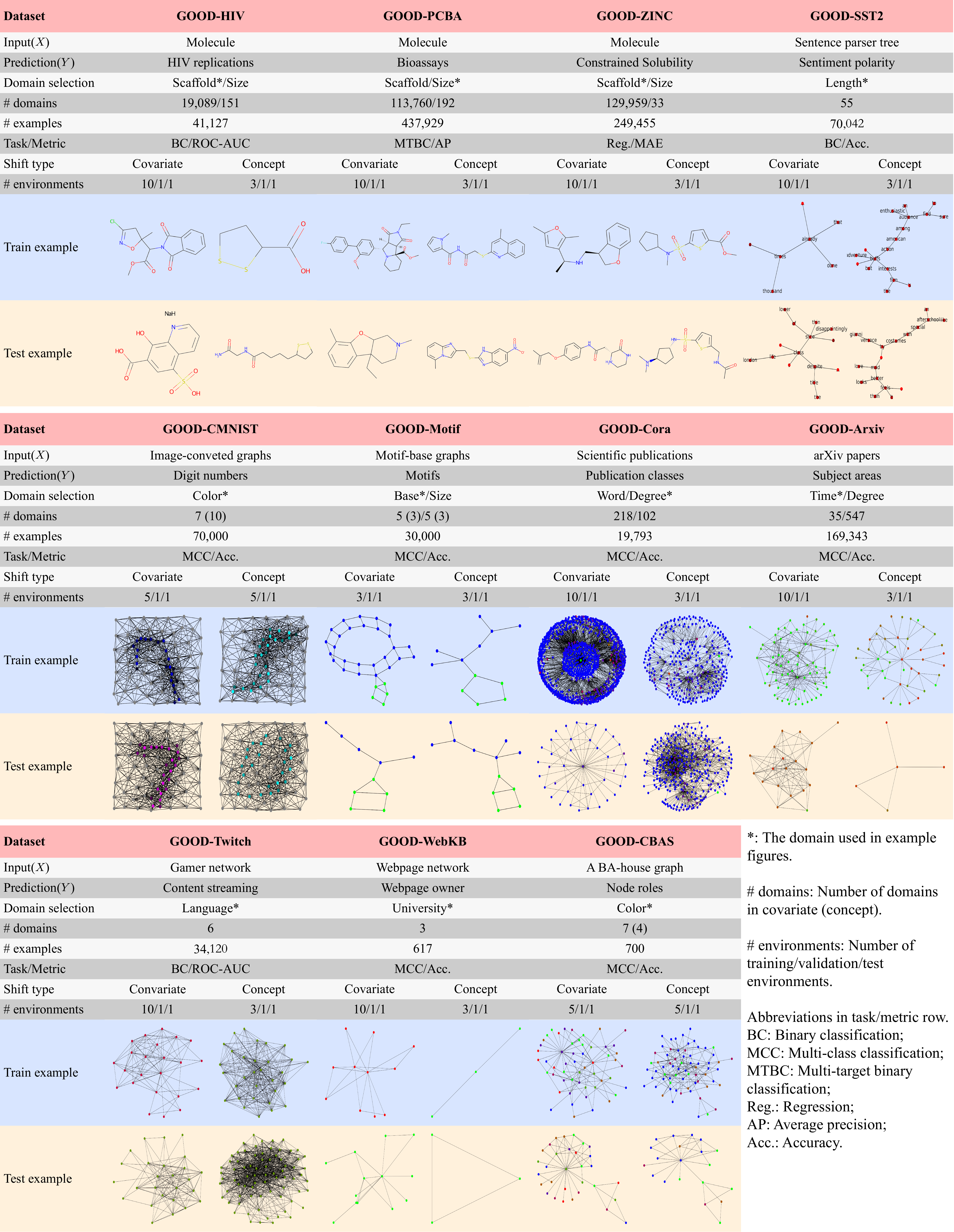}}
        % \vspace{-0.2cm}
    % \setlength{\belowcaptionskip}{-10pt}
    
    \caption{
    A summary of datasets included in the proposed benchmark. 
    For a covariate shift, the training and test examples are from different domains. For a concept shift, examples are chosen from the same domain with different labels to show the different domain-output correlations.
    }
    % \vspace{-0.2cm}
    \label{fig:datasets}
\end{figure}

\textbf{Covariate shift and concept shift.} Distribution shifts can generally be defined as two types; \ie, covariate shift and concept shift (drift)~\cite{quinonero2008dataset, MORENOTORRES2012521, widmer1996learning}. Formally, in supervised learning, a model is trained to predict an output $Y\in\mathcal{Y}$ given an input $X\in\mathcal{X}$, also known as a covariate variable. The output $Y$ is categorical in classification and continuous in regression problems. In multi-task learning, the output $Y$ becomes a vector, and we consider each task separately. Since the joint distribution $P(Y, X)$ can be written as $P(Y|X)P(X)$, two types of OOD problems are commonly considered, namely covariate and concept shifts. In covariate shift, the input distributions have been shifted between training and test data. Formally, $P^\text{train}(X)\neq P^\text{test}(X)$ and $P^\text{train}(Y|X) = P^\text{test}(Y|X)$, where $P^\text{train}(\cdot)$ and $P^\text{test}(\cdot)$ denote training and test distributions, respectively. In contrast, in concept shift, the conditional distribution $P(Y|X)$ has been shifted as
$P^\text{train}(Y|X)\neq P^\text{test}(Y|X)$ and $P^\text{train}(X)=P^\text{test}(X)$. In this work, we explicitly make distinctions and consider both shifts.

\textbf{Differences between graph OOD and general OOD.} Traditional OOD methods commonly focus on simple structure equation models~\cite{arjovsky2019invariant, ahuja2021invariance, rosenfeld2020risks, lu2021invariant} or computer vision tasks~\cite{lu2021invariant, tzeng2017adversarial, ganin2016domain, sun2016deep}. In these tasks, the inputs are variables or image features, denoted as $F$. However, graph data possess the complex nature of irregularity and connectivity in topology. A graph is commonly represented as an input pair $(F, A)$, where $F$ are node/edge features, and $A$ is the adjacency matrix. Consequently, graph OOD problems focus not only on general feature distribution shifts but also on structure distribution shifts. Graph neural networks are designed based on $F$ and $A$ to pass messages, demonstrating that structures and features carry different perspectives of input information of a graph.
The uniqueness of graph data prompts the development of graph-specific OOD methods~\cite{wu2022dir, wu2022handling, bevilacqua2021size, zhu2021shift, fan2021generalizing, li2021ood} and calls for graph OOD benchmarks.
\footnote{For simplicity, we use $X$ instead of $(F, A)$ to denote inputs of graphs in our benchmark design. Therefore, a feature in $X$ may refer to not only a node feature but also a specific graph structure.}

\textbf{Contributions and novelty.} In this work, we develop a systematic graph OOD benchmark, known as GOOD.  
As design principles, we strive to (1) create non-trivial performance gaps between training and test data; and (2) provide carefully designed data environments to ensure that the induced distribution shifts are potentially solvable for models.
Specifically, GOOD contains 6 graph-level datasets and 5 node-level datasets as shown in Fig.~\ref{fig:datasets}. 
For each dataset, we select one or two types of domains. Given a domain, we generate no-shift, covariate shift, and concept shift splits for ease of comparison among 10 baselines.
We summarize our novel contributions as follows.
(1) To the best of our knowledge, no existing OOD benchmark provides both covariate and concept shifts comparison for the same domain selection. This is important as it sheds light on the differences between various shifts with proper variable control. For example, from our experiments, DIR~\cite{wu2022dir} performs favorably against other methods in concept shift of GOOD-HIV size split while failing in the corresponding covariate shift.
(2) In terms of graph OOD, we are the first benchmark to include not only graph classification, but also graph regression and node-level datasets, which improves the diversity of graph OOD tasks. 
(3) GOOD provides numerous comparisons for 51 different dataset splits and 10 OOD methods, providing solid baselines for future method developments.

\section{Related Work}\label{sec:related}

OOD or distribution shift is a longstanding problem in machine learning and artificial intelligence~\cite{hand2006classifier,quinonero2008dataset,MORENOTORRES2012521, shen2021towards}.
To address OOD problem substantially, several benchmarks have been curated~\cite{gulrajani2020search, ye2021ood, he2021towards, koh2021wilds, zhang2022nico++} to evaluate different algorithms~\cite{ganin2016domain, sun2016deep, peters2016causal, arjovsky2019invariant, krueger2021out, sagawa2019distributionally, zhang2017mixup, ahuja2021invariance}. 
DomainBed~\cite{gulrajani2020search} is an early OOD benchmark in computer vision.
Following DomainBed, OoD-Bench~\cite{ye2021ood} collects datasets and categorizes them into diversity and correlation shifts.
WILDS~\cite{koh2021wilds} collects real-world data from wild and studies domain generalization and subpopulation shift.
Specifically, domain generalization focuses on disjoint training and test domains, while subpopulation shift considers shifts between majority and minority groups, which leads to insufficient training for minority data.
With the success of graph neural networks~\cite{kipf2017semi, xu2018how, velickovic2018graph, gao2019graph, liu2021dig, gao2021topology, liu2020deep,liu2022spherical}, graph OOD problems are gaining growing attention~\cite{wu2022dir, wu2022handling, bevilacqua2021size, zhu2021shift, fan2021generalizing, li2021ood, li2022out, chen2022invariance}. 
GDS~\cite{ding2021closer} collects several datasets to compare the performance of well-known baselines and data augmentation methods. DrugOOD~\cite{ji2022drugood} is a recent benchmark specifically designed for molecular graph OOD problems. It is curated based on a large-scale bioassay dataset ChEMBL~\cite{mendez2019chembl} and includes an automated pipeline for obtaining more datasets.

\textbf{Differences with existing (graph) OOD benchmarks.} Generalization abilities of OOD algorithms for covariate shift~\cite{shimodaira2000improving} and concept shift~\cite{widmer1996learning, hand2006classifier,alaiz2008assessing, krueger2021out} differ fundamentally. However,
existing benchmarks, including but not limited to graph OOD benchmarks, either ignore one of the shifts or fail to compare the two shifts of the same feature on the same dataset.
Firstly, most existing OOD benchmarks include only one type of shift. For example, DrugOOD~\cite{ji2022drugood} focuses on domain generalization for molecules, exclusively considering covariate shift. WILDS~\cite{koh2021wilds} includes domain generalization and subpopulation shift, which are two cases of covariate shift, while still ignoring concept shift. 
Secondly, a few benchmarks~\cite{ye2021ood} involve both shifts but simply categorize each dataset as one of these two shifts. GDS~\cite{ding2021closer} collects eight datasets but makes no distinctions between the two shifts; among their datasets, we can categorize ColoredMNIST as concept shift and others as covariate shift. In contrast, our benchmark proposes novel dataset splitting methods to generate both shifts for the same domain selection on the same dataset to enable comparison between shifts. This variable-controlled comparison enables a more thorough analysis of shifts given any specific domain, leading to a more comprehensive OOD benchmark.
Furthermore, we curate more diverse graph datasets with diverse tasks, including single/multi-task graph classification, graph regression, and node classification, while neither GDS nor DrugOOD includes graph regression or node-level tasks. In addition, while GDS and DrugOOD do not benchmark any graph-specific OOD methods, we evaluate 4 graph-specific OOD methods, shedding light on further graph OOD research.

\section{The GOOD Benchmark Design}

When training and test samples are assumed to be i.i.d., random split is commonly used to split datasets into training and test sets. In contrast, 
splits in OOD problems should be carefully designed in order to accurately assess the generalization ability of algorithms. In GOOD, we consider both covariate and concept shifts and meticulously design data splits to ensure these shifts are reflected.
Formally, following prior invariant learning work~\cite{arjovsky2019invariant, chen2022invariance, ahuja2021invariance, rosenfeld2020risks, lu2021invariant}, as shown in Fig.~\ref{fig:causal_graph}, $C, S_1, S_2\in \mathcal{Z}$ are the latent variables that causes target $Y\in\mathcal{Y}$, is non-causally associated with $Y$, and is independent to $Y$, respectively. $\rightarrow$ denotes the causal mapping. $S_2$ is commonly caused by target-irrelevant environments.\footnote{We do not explicitly consider unobserved confounders in this paper.} In the input feature space, given input features $X\in \mathcal{X}$, we assume the invariant features $X_\text{inv}\in X$ are projected by an injective function from $C$.\footnote{This assumption is for clearer input distinction, and will not introduce any side effect to further analysis, since the shift split design focuses on selecting $X_\text{ind}$ instead of distinguishing $X_\text{inv}$ and $X_\text{ass}$.} Therefore, $X_\text{inv}$ can fully determine $Y$. $X_\text{ass}$ denotes input features associated with $Y$ by confounding and anti-causal associations through $C$ and $S_1$. $X_\text{ind}$ are input features independent to $Y$ and are only caused by $S_2$. 
In practice, it might be hard to strictly separate $X_\text{inv}$, $X_\text{ass}$, and $X_\text{ind}$. We try to only select and shift part of $X_\text{ind}$, but since our shift splits contain significant dataset shifts, the selection of parts of $X_\text{ass}$ won't affect the benchmarking performance. Though, we use $X_\text{ind}$ throughout this paper for simplicity.

\subsection{Covariate shifts}\label{sec:covariate}

Domain generalization methods follow the covariate shift assumption~\cite{ben2010theory} and assume that the covariate distribution $P(X)$ shifts across splits, while the concept distribution $P(Y|X)$ remains the same.
This implies that a shift of $P(X)$ should not cause corresponding shift in $P(Y|X)$.
That is, covariate shift can only happen on input features that are not associated with $Y$. Therefore, with prior knowledge, we can manually select and shift one or several of these irrelevant features, $X_\text{ind}$, to build covariate splits. Different $X_\text{ind}$ feature values indicate different domains, and each domain can be viewed as a split.
For instance, in the graph ColoredMNIST dataset in which we distinguish hand-written digits with colors, the color is irrelevant with labels. Thus, in our covariate splits, digits with different colors belong to corresponding color domains, and each domain becomes a split.

Formally, possible values of $X_\text{ind}$ are discrete and finite. Therefore, we define each domain by its unique $X_\text{ind}$ value, forming its unique input distribution $P(X)$. Then, a dataset can be viewed as a mixture of $|\mathcal{D}|$ domains as $\mathcal{D}=\{d_1, \ldots, d_{|\mathcal{D}|}\}$. 
For a domain $d_i$, we represent its input distribution as $P_{d_i}(X)$. Specifically, $P_{d_i}(X_\text{inv}, X_\text{ass})$ is fixed while $P_{d_i}(X_\text{ind}=\mathbf{x}_i)=1$, where $\mathbf{x}_i$ is the value of $X_\text{ind}$ in domain $d_i$. Since $X_\text{ind}$ is independent to $Y$, $P_{d_i}(Y|X)=P(Y|X)$. The data distribution is
\begin{equation}\label{eq:covariate}
\setlength{\abovedisplayskip}{0.5pt}
\setlength{\belowdisplayskip}{0.5pt}
P(Y, X)=\sum\limits^{|\mathcal{D}|}_{i=1}w_i P_{d_i}(Y, X)=\sum\limits^{|\mathcal{D}|}_{i=1}w_i P_{d_i}(X)P(Y|X), 
\end{equation}
where $w_i$ is the mixture weight for domain $d_i$.

\begin{figure}[!tb]
    \centering
    \begin{tabular}[c]{cc}
\begin{subfigure}[c]{0.52\textwidth}
    \includegraphics[width=\textwidth]{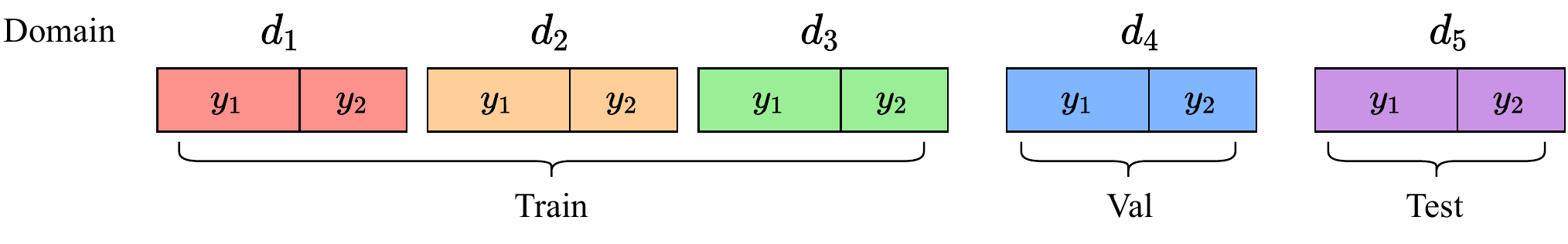}
    \vspace{-0.5cm}
    \caption{Covariate shift split}
    \label{fig:split1}
\end{subfigure}&
% \hfill

\multirow{2}{*}{ % [14pt]
    \begin{subfigure}{0.26\textwidth}
      \includegraphics[width=\textwidth]{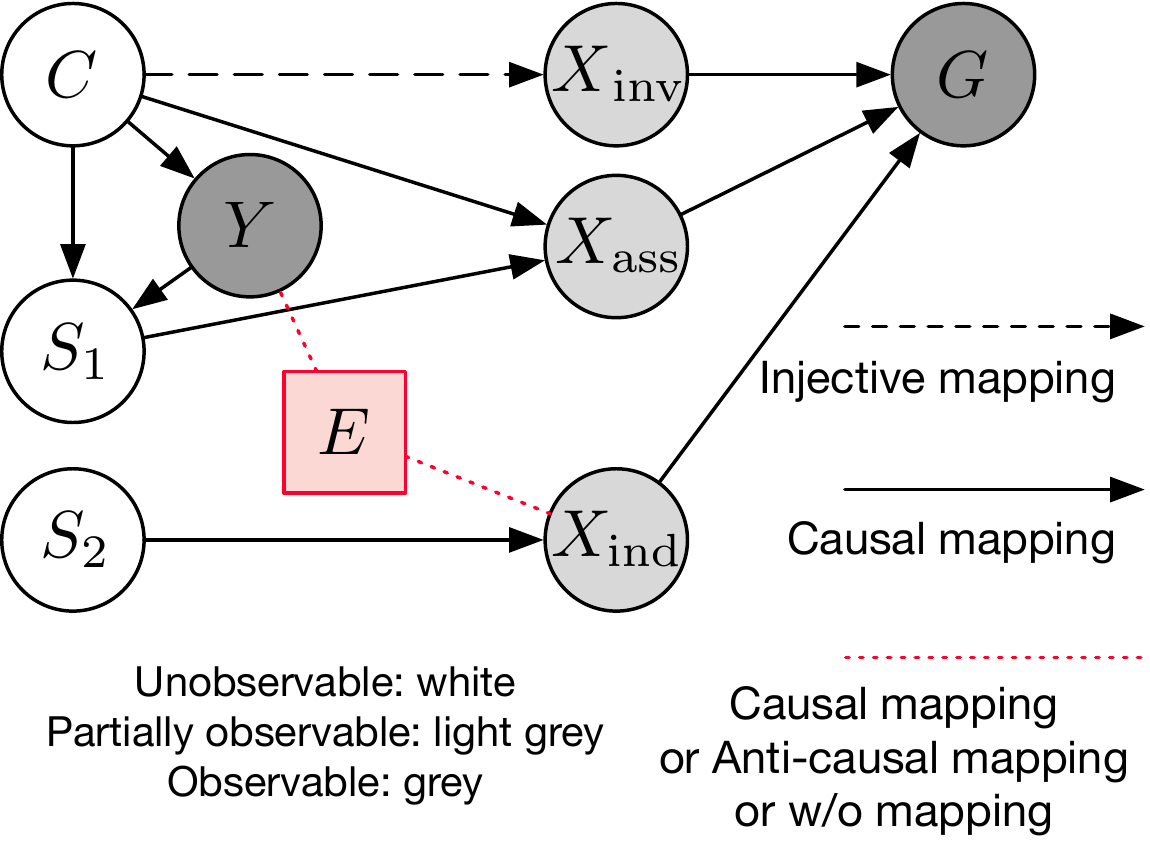} %height=2cm, 
      \caption{Causal graph}
    %   \vspace{-0.2cm}
      \label{fig:causal_graph}
    \end{subfigure}
} \\

    \begin{subfigure}[c]{0.7\textwidth}
    \includegraphics[width=\textwidth]{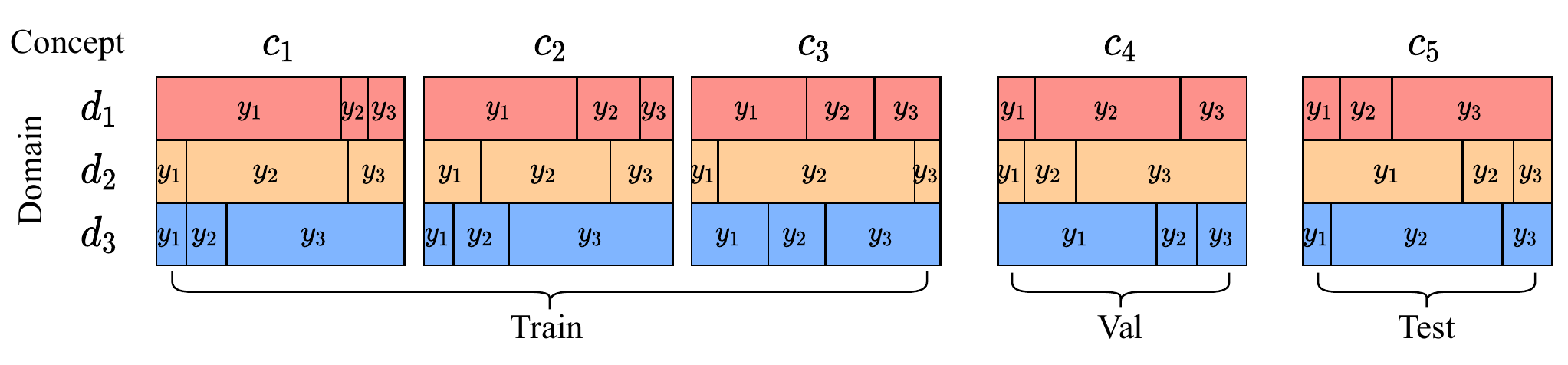}
    \caption{Concept shift split}
    \label{fig:split2}
\end{subfigure}

\end{tabular}    
\caption{(a) Illustration of covariate shift split. Five domains are denoted as different colors, where each domain includes outputs of the same distribution. We sort the dataset according to the domain $d_i$, then group them into train/validation/test sets. (c) Illustration of concept shift split. Each concept includes all three domains, and each domain has spurious correlations with a specific output in a concept. For example, in concept $c_1$, the domain colored in red is highly associated with $y_1$, but this domain corresponds to $y_2$ in concept $c_4$. Note that the distributions of concepts in training are similar. (b) Illustration of causal graph~\cite{pearl2009causality, peters2016causal} for dataset generation and observation. (Left) $C$, $S_1$, $S_2$ locate in the latent space and are not observable. (Middle) $X_\text{inv}$, $X_\text{ass}$, and $X_\text{ind}$ are input features that can be partially observed and selected manually, such as motif shapes or molecule scaffolds. (Right) $G$ is the graph data input including node features and adjacency matrices. $E\in \mathcal{E}$ is the environment variable that can determine or be determined by $X_\text{inv}$ and $Y$ according to different types of datasets and shifts. Detailed discussions can be found in Appendix~\ref{sec:A}.}
    \label{fig:summary}
\end{figure}

\textbf{Comparison of covariate split design on 3 types of datasets.} We perform covariate shift splits on synthetic, semi-artificial, and real-world datasets. For a synthetic dataset, given the $X_\text{ind}$ feature values of each domain, we generate graphs according to its domain distributions, respectively. For a semi-artificial dataset, given a graph, we generate extra variant features to produce modified graphs that follow the domain distribution. Since we cannot create or modify graphs for real-world datasets, we use carefully designed data splits. As shown in Fig.~\ref{fig:split1}, we sort the graphs by their domain $d_i\in \mathcal{D}$ and then divide the dataset into five domain splits with a specific split ratio, \eg, $20\%$, for each domain. Finally, the training, validation, and test sets are obtained based on domains without intersections. The difference between synthetic/semi-artificial datasets and real-world datasets is that the domain of graphs in artificial datasets can be defined arbitrarily before feature modifications, but the domain of graphs in real-world datasets should be defined strictly according to its features.

\subsection{Concept shifts}

In contrast to covariate shift, concept shift considers the scenario in which the concept distribution $P(Y|X)$ is shifted across splits.
%, while the covariate distribution $P(X)$ remains the same. 
Since $X_\text{inv}$ can fully determine $Y$, $P(Y|X_\text{inv})$ is invariant. Thus, shifts of $P(Y|X)$ can only happen with shifts of $P(Y|X_\text{ass})$ and $P(Y|X_\text{ind})$. Since $X_\text{ind}$ is irrelevant with $Y$, the correlation $P(Y|X_\text{ind})$ between $Y$ and $X_\text{ind}$ in each domain is spurious correlation. However, the association built between $Y$ and $X_\text{ind}$ will also connect $X_\text{inv}$ to $X_\text{ind}$ through $Y$ according to Fig.~\ref{fig:causal_graph}, leading to the change of $P(X_\text{inv}|X_\text{ind})$, so that $P(X)$ will be changed inevitably. Therefore, we can only build major concept shifts with necessary covariate shifts. We will still call it concept shift for simplicity and distinction. Therefore, given the selected domain features $X_\text{ind}$, we can build concept shift splits by manually creating such spurious correlations of certain rates. For example, the spurious correlation rate between the domain $d_i$ and the output value $y_j$ can be set as $P(Y=y_j| X_\text{ind}=\mathbf{x}_i)=90\%$. Specifically, different spurious correlation rates define different concepts, and each concept can be viewed as a split. We use the graph ColoredMNIST dataset as an example, as shown in Fig.~\ref{fig:split2}. In every concept, each color domain is highly correlated with a label. Therefore, in our concept splits, different spurious color-label correlation rates determine different concepts, and each concept becomes a split.

Formally, a dataset can be viewed as a mixture of $|\mathcal{C}|$ concepts $\mathcal{C}=\{c_1, \ldots, c_{|\mathcal{C}|} \}$ in our concept shift split. 
We use $P_{y_j,d_i}(Y)$ to represent a certain output distribution on value $y_j$ given domain $d_i$, defined as $P(Y=y_j|X_\text{ind}=\mathbf{x_i})=1$.
We first consider the classification case in which $Y$ is categorical.
Given a concept ${c_k}$, we formulate its 2-D conditional distribution $P_{c_k}(Y|X)$
% the set of domain-output correlations, 
by describing multiple 1-D distributions; that is, for each domain $d_i$,
\begin{equation}\label{eq:concept1}
\setlength{\abovedisplayskip}{0.5pt}
\setlength{\belowdisplayskip}{0.5pt}
P_{c_k}(Y|X_\text{ind}=\mathbf{x_i})=\sum\limits^{|\mathcal{Y}|}_{j=1} q_{i, j}^k P_{y_j,d_i}(Y), 
\end{equation}
where 
$q_{i, j}^k$ is the rate of the spurious correlation in concept ${c_k}$ between domain $d_i$ and output $y_j$. In the regression case where $Y$ is continuous, the sum becomes integral. In the multi-task case, $Y$ and $y_j$ become vectors.
The overall dataset distribution can be written as 
\begin{equation}\label{eq:concept2}
\setlength{\abovedisplayskip}{0.5pt}
\setlength{\belowdisplayskip}{1pt}
P(Y, X) = \sum\limits_{k=1}^{|\mathcal{C}|}w_k P_{c_k}(Y, X)=\sum\limits_{k=1}^{|\mathcal{C}|}w_k P(X)_{c_k}P_{c_k}(Y|X), 
\end{equation}
where $w_k$ is the mixture weight for concept $c_k$.

\textbf{Comparisons of concept split design on 3 types of datasets.} In practice, we create significant concept shifts between training, validation, and test sets, in which the domain-output correlations are completely different. 
Note that mixing different domain-output correlations can weaken the spuriousness of these correlations. Thus, concepts within the training set are designed to have similar domain-output correlations to guarantee the concept shift between training and test.
Concretely, we perform concept shift splits on synthetic, semi-artificial, and real-world datasets.
For synthetic datasets, we generate graphs where the domain feature is highly correlated with a specific output according to the preset correlation in the concept. For semi-artificial datasets, given a graph and a concept, we generate extra features as domains to build spurious domain-output correlations. For real-world datasets, we cannot create or modify data. Thus, we propose a screening approach to scan and select graphs in the dataset. 
Each graph has a probability to be included in a concept $c_k$ according to the value of $q_{i, j}^k$. 
To conclude, the difference in concept splits between artificial datasets and real-world datasets is located in whether we can \textit{arbitrarily} determine the concept of a graph. We can define the concept of a graph in artificial datasets, but the concept of a graph in real-world datasets is defined with a probability according to $X_\text{ind}$ and target $Y$.

\subsection{Environments}\label{sec:env}

Many current OOD learning algorithms~\cite{sagawa2019distributionally, krueger2021out} follow the framework of invariant causal predictor (ICP)~\cite{peters2016causal} and invariant risk minimization (IRM)~\cite{arjovsky2019invariant}, assuming that the training data form groups by distributions, known as environments. This framework assumes the data are similar within an environment and dissimilar across different environments. Since OOD problems are complicated and multi-perspective, works under this framework explicitly or implicitly injects environment information to models to figure out the specific generalization direction for better OOD performances.
Specifically, the shift between training and test data, though more significant, should be similarly reflected among different training environments, so that OOD models can potentially grasp the shift between training and test data by learning the shifts among different training environments.
Following this strategy, to enhance the OOD generalization ability of models, we use the distribution shift information provided by the difference of training environments to convey the types of shifts expected between training and test data.
In covariate shift, environments take the form of domains. During training, models can learn from $P^\text{train}(X_\text{ind})$, which varies across domains, that $X_\text{ind}$ is not causally related to labels, thereby preventing the unknown $P^\text{test}(X_\text{ind})$ from misleading predictions during test.
In concept shifts, environments take the form of concepts.
By learning from different spurious correlations across training concepts, models can learn that the domain-output correlations $P^\text{train}(Y|X_\text{ind})$ are spurious, thereby avoiding being misled by the new spurious correlation $P^\text{test}(Y|X_\text{ind})$ during test.

Formally, we consider a dataset with a set of $|\mathcal{E}|$ environments $\mathcal{E}=\{e_1, \ldots, e_{|\mathcal{E}|}\}$, each with distribution $P_e(Y,X)$ for $e \in \mathcal{E}$ (Fig~\ref{fig:causal_graph}). In this case, the dataset distribution $P(Y,X)=\sum_{e}P_e(Y,X)$.
Specifically, for both covariate and concept shifts, the distributions $P^\text{train}$ and $P^\text{test}$ are weighted combinations of environment distributions $P_e(Y,X)$. With the training and test environments $\mathcal{E}^\text {train}, \mathcal{E}^\text{test} \subset \mathcal{E}$, we express $P\textsuperscript{train}=\sum_{e\in \mathcal{E}^\text{train}}w_e^\text{train}P_e(Y, X)$ and $P\textsuperscript{test}=\sum_{e\in \mathcal{E}^\text{test}}w_e^\text{test}P_e(Y, X)$, where $w_e^\text{train}$ and $w_e^\text{test}$ are the weights for each training and test environment, respectively. 

\section{The GOOD Datasets}\label{sec:dataset}
In this section, we introduce the datasets in GOOD. The benchmark contains 11 datasets, covering multiple tasks and data sources. 
For each dataset, we select one or two domain features. Then we apply covariate and concept shift splits per domain to create diverse distribution shifts between training, OOD validation, and OOD test sets. 
Finally, we shuffle the training set and divide it into final training set, in-domain (ID) validation set, and in-domain (ID) test set. 
Summary statistics of datasets are given in Fig.~\ref{fig:datasets}. Other details and data processing details are included in Appendix~\ref{sec:A}.

\subsection{Graph prediction tasks}

\textbf{GOOD-HIV} is a small-scale real-world molecular dataset adapted from MoleculeNet~\cite{wu2018moleculenet}. The inputs are molecular graphs in which nodes are atoms, and edges are chemical bonds. The task is to predict whether the molecule can inhibit HIV replication. 
We design splits based on two domain selections, namely, scaffold and size. 
The first one is Bemis-Murcko scaffold~\cite{bemis1996properties} which is the two-dimensional structural base of a molecule. The second one is the number of nodes in a molecular graph, an inevitable structural feature of a graph. Both features should not determine the label, therefore, both can become major sources of distribution shifts. 
For each domain selection, the value space for the feature is very large, therefore we cluster graphs with similar domain values into one environment, improving the OOD learning procedure and reducing the training time complexity.

\textbf{GOOD-PCBA} is a real-world molecular dataset from~\citet{wu2018moleculenet}. It includes 128 bioassays, forming 128 binary classification tasks. Due to the extremely unbalanced classes (only 1.4\% positive labels), we use the Average Precision (AP) averaged over the tasks as the evaluation metric. GOOD-PCBA uses the same domain selections as GOOD-HIV.

\textbf{GOOD-ZINC} is a real-world molecular property regression dataset from ZINC database~\cite{gomez2018automatic}. The inputs are molecular graphs with up to 38 heavy atoms, and the task is to predict the constrained solubility~\cite{jin2018junction, kusner2017grammar} of molecules. GOOD-ZINC uses the same domain selections as GOOD-HIV. 

\textbf{GOOD-SST2} is a real-world natural language sentimental analysis dataset adapted from~\citet{yuan2020explainability}. Each sentence is transformed into a grammar tree graph, where each node represents a word with corresponding word embeddings as node features. The dataset forms a binary classification task to predict the sentiment polarity of a sentence. We select sentence lengths as domains since the length of a sentence should not affect the sentimental polarity.

\textbf{GOOD-CMNIST} is a semi-artificial dataset designed for node feature shifts. It contains graphs of hand-written digits transformed from MNIST database using superpixel techniques~\cite{monti2017geometric}. Following \citet{arjovsky2019invariant}, we color digits according to their domains and concepts.
Specifically, 
in covariate shift split, we color digits with 7 different colors, and digits with the first 5 colors, the 6th color, and the 7th color are categorized into training, validation, and test sets.
In concept shift split, we color digits with 10 colors. Each color is highly correlated with one digit label in the training set, while colors have weak correlations and no correlation with labels in validation and test sets, respectively.

\textbf{GOOD-Motif} is a synthetic dataset motivated by Spurious-Motif~\cite{wu2022dir} and is designed for structure shifts. Particularly, GOOD-CMNIST and GOOD-Motif compose an OOD algorithm check for both feature/structure shifts. Each graph in the dataset is generated by connecting a base graph and a motif, and the label is determined by the motif solely. 
Instead of combining the base-label spurious correlations and size covariate shift together as in \citet{wu2022dir}, we study covariate and concept shifts separately. Specifically, we generate graphs using five label irrelevant base graphs (wheel, tree, ladder, star, and path) and three label determining motifs (house, cycle, and crane). To create covariate and concept splits, we select the base graph type and the size as domain features.

\subsection{Node prediction tasks}

\textbf{GOOD-Cora} is a citation network adapted from the full Cora dataset~\cite{bojchevski2017deep}.  
The input is a small-scale citation network graph, in which nodes represent scientific publications and edges are citation links. The task is a 70-class classification of publication types.
We generate splits based on two domain selections, namely, word and degree. The first one is the word diversity defined by the selected-word-count of a publication, purely irrelevant with the label. The second one is node degree in the graph, implying that the popularity of a paper should not determine the class of a paper.

\textbf{GOOD-Arxiv} is a citation dataset adapted from OGB~\cite{hu2020open}. The input is a directed graph representing the citation network among the computer science (CS) arXiv papers. Nodes in the graph represent arXiv papers, and directed edges represent citations. 
The task is predicting the subject area of arXiv CS papers, forming a 40-class classification problem. 
We generate splits based on two domain selections; \ie, time (publication year) and node degree. 

\textbf{GOOD-Twitch} is a gamer network dataset. The nodes represent gamers with games as node features, and the edge represents the friendship connection of gamers. The binary classification task is to predict whether a user streams mature content. The domain of GOOD-Twitch splits includes user language, implying that the prediction target should not be biased by the language a user uses.

\textbf{GOOD-WebKB} is a university webpage network dataset. A node in the network represents a webpage, with words appearing in the webpage as node features, and edges are hyperlinks between webpages. Its 5-class prediction task is to predict the classes of webpages. We split GOOD-WebKB according to the domain university, suggesting that classified webpages are based on word contents and link connections instead of university features.

\textbf{GOOD-CBAS} is a synthetic dataset modified from BA-Shapes~\cite{ying2019gnnexplainer}. The input is a graph created by attaching 80 house-like motifs to a 300-node Barabási–Albert base graph, and the task is to predict the role of nodes, including the top/middle/bottom node of a house-like motif or the node from the base graph, forming a 4-class classification task. Instead of using constant node features, we generate colored features as in GOOD-CMNIST so that OOD algorithms need to tackle node color differences in covariate splits and color-label correlations in concept splits.

\section{Experimental Studies}

We conduct experiments on 11 datasets with 10 baseline methods.
For each dataset, we use the same GNN backbone for all baseline methods for fair comparisons.
Specifically, we use GIN-Virtual~\cite{xu2018how, gilmer2017neural} and GCN~\cite{kipf2017semi,Zeng2020GraphSAINT:} as GNN backbones for graph and node prediction tasks, respectively.
Note that for GOOD-Motif, we adopt GIN~\cite{xu2018how} as the GNN backbone since adding virtual nodes does not improve the performance.
For all experiments, we select the best checkpoints for ID and OOD tests according to results on ID and OOD validation sets, respectively. Experimental details and hyper-parameter selections are provided in Appendix~\ref{sec:B}.
All the datasets, implementation codes, and best checkpoints to reproduce the results in this paper are available at \url{https://github.com/divelab/GOOD/}.

\subsection{In-distribution versus out-of-distribution performance gap}

As introduced in Sec.~\ref{sec:introduction}, one principle for designing GOOD is to create non-trivial distribution shifts and performance gaps between training and test data. Equivalently, we expect distinct performance gaps between ID and OOD settings.
To verify performance gaps, we run experiments using empirical risk minimization (ERM) and summarize the results in Table~\ref{table:gap}.
The differences between ID\textsubscript{ID} and OOD\textsubscript{ID} or OOD\textsubscript{OOD}for each domain selection and distribution shift show the substantial and consistent performance gap between the ID and OOD settings. In addition, for most splits, OOD\textsubscript{OOD} is better than OOD\textsubscript{ID}. This implies that OOD validation sets outperform ID validation sets in selecting models with better generalization ability.

\begin{table*}[t]
% \setlength{\abovecaptionskip}{0.1cm}
% \footnotesize
% \scalebox{0.622}{
% % \setlength{\tabcolsep}{0.3mm}{
% \centering
% \setlength{\textfloatsep}{3pt}
% \setlength{\floatsep}{1ex}
\centering
\resizebox{\textwidth}{!}
{
\begin{tabular}{lllcccccccccccccc}
\toprule[2pt]

& & & \multicolumn{7}{c}{domain selection 1} & \multicolumn{7}{c}{domain selection 2}\\
\cmidrule(r){4-10} \cmidrule(r){11-17}
& & & \multicolumn{3}{c}{covariate} & \multicolumn{3}{c}{concept} & \multicolumn{1}{c}{no-shift} & \multicolumn{3}{c}{covariate} & \multicolumn{3}{c}{concept} & \multicolumn{1}{c}{no-shift} \\
\cmidrule(r){4-6} \cmidrule(r){7-9} \cmidrule(r){10-10} \cmidrule(r){11-13} \cmidrule(r){14-16} \cmidrule(r){17-17}

& & & \multicolumn{1}{c}{ID\textsubscript{ID}} & \multicolumn{1}{c}{OOD\textsubscript{ID}} & \multicolumn{1}{c}{OOD\textsubscript{OOD}} & \multicolumn{1}{c}{ID\textsubscript{ID}} & \multicolumn{1}{c}{OOD\textsubscript{ID}} & \multicolumn{1}{c}{OOD\textsubscript{OOD}} & \multicolumn{1}{c}{ID\textsubscript{ID}} & \multicolumn{1}{c}{ID\textsubscript{ID}} & \multicolumn{1}{c}{OOD\textsubscript{ID}} & \multicolumn{1}{c}{OOD\textsubscript{OOD}} & \multicolumn{1}{c}{ID\textsubscript{ID}} & \multicolumn{1}{c}{OOD\textsubscript{ID}} & \multicolumn{1}{c}{OOD\textsubscript{OOD}} & \multicolumn{1}{c}{ID\textsubscript{ID}} \\
\midrule[1pt]

\multicolumn{3}{l}{GOOD-HIV\textuparrow} & 82.79 & 68.86 & 69.58 & 84.22 & 65.31 & 72.33 & 80.86 & 83.72 & 58.41 & 59.94 & 88.05 & 44.75 & 63.26 & 80.86\\
\multicolumn{3}{l}{GOOD-PCBA\textuparrow} & 33.45 & 16.87 & 16.89 & 25.95 & 21.34 & 21.63 & 33.77 & 34.31 & 17.81 & 17.86 & 32.54 & 14.83 & 15.36 & 33.77\\
\multicolumn{3}{l}{GOOD-ZINC\textdownarrow} & 0.1224 & 0.1825 & 0.1995 & 0.1222 & 0.1328 & 0.1306 & 0.1233 & 0.1199 & 0.2569 & 0.2427 & 0.1315 & 0.1418 & 0.1403 & 0.1233\\
\multicolumn{3}{l}{GOOD-SST2\textuparrow} & 89.82 & 77.76 & 81.30 & 94.43 & 67.26 & 72.43 & 91.61 & -- & -- & -- & -- & -- & -- & --\\
\multicolumn{3}{l}{GOOD-CMNIST\textuparrow} & 77.96 & 26.90 & 28.60 & 90.00 & 40.80 & 42.87 & 77.30 & -- & -- & -- & -- & -- & -- & --\\
\multicolumn{3}{l}{GOOD-Motif\textuparrow} & 92.60 & 69.97 & 68.66 & 92.02 & 80.87 & 81.44 & 92.09 & 92.28 & 51.28 & 51.74 & 91.73 & 69.41 & 70.75 & 92.09\\
\multicolumn{3}{l}{GOOD-Cora\textuparrow} & 70.43 & 64.44 & 64.86 & 66.05 & 64.20 & 64.60 & 69.41 & 72.27 & 55.76 & 56.30 & 68.71 & 60.38 & 60.54 & 69.42\\
\multicolumn{3}{l}{GOOD-Arxiv\textuparrow} & 72.69 & 70.64 & 71.08 & 74.76 & 65.70 & 67.32 & 73.02 & 77.47 & 58.53 & 58.91 & 75.27 & 61.77 & 62.99 & 72.99\\
\multicolumn{3}{l}{GOOD-Twitch\textuparrow} & 70.66 & 47.73 & 48.95 & 80.29 & 48.57 & 57.32 & 68.05 & -- & -- & -- & -- & -- & -- & --\\
\multicolumn{3}{l}{GOOD-WebKB\textuparrow} & 38.25 & 11.64 & 14.29 & 65.00 & 24.77 & 27.83 & 47.85 & -- & -- & -- & -- & -- & -- & --\\
\multicolumn{3}{l}{GOOD-CBAS\textuparrow} & 89.29 & 77.57 & 76.00 & 89.79 & 82.22 & 82.36 & 99.43 & -- & -- & -- & -- & -- & -- & --\\
\bottomrule[2pt]
\end{tabular}
}
% }
\caption{
ID and OOD performance gaps learned with ERM across 51 splits. The metric and domain selections for each dataset are in Fig.~\ref{fig:datasets}. \textuparrow { }indicates higher values correspond to better performance while \textdownarrow { }indicates lower for better. ID test results with ID validations are denoted as ID\textsubscript{ID}, while OOD test results with ID/OOD validations are written as OOD\textsubscript{ID} and OOD\textsubscript{OOD}, respectively. Note that the no-shift random split only has the ID setting. We report the average values over 10 runs. The standard deviations are listed in Appendix~\ref{sec:D}.
}
\label{table:gap}
\end{table*}

\subsection{Performance of baseline algorithms}
In our benchmark, we conduct experiments with 10 baseline methods. Based on the comparison results, we provide an analysis of the learning strategy of OOD methods.

\begin{table*}[t]
\setlength{\abovecaptionskip}{0.1cm}
\footnotesize
\centering
\resizebox{\textwidth}{!}
{
% \scalebox{0.59}{
% \setlength{\tabcolsep}{0.3mm}{
\begin{tabular}{lllccccccccccccccc}
\toprule[2pt]
\multicolumn{3}{c}{\multirow{3}{*}{covariate}} & \multicolumn{4}{c}{GOOD-HIV\textuparrow} & \multicolumn{4}{c}{GOOD-PCBA\textuparrow} & \multicolumn{4}{c}{GOOD-ZINC\textdownarrow} & \multicolumn{2}{c}{GOOD-CMNIST\textuparrow} & \multirow{11}{*}{graph}\\
\cmidrule(r){4-7} \cmidrule(r){8-11} \cmidrule(r){12-15} \cmidrule(r){16-17}

& & & \multicolumn{2}{c}{scaffold} & \multicolumn{2}{c}{size} & \multicolumn{2}{c}{scaffold} & \multicolumn{2}{c}{size} & \multicolumn{2}{c}{scaffold} & \multicolumn{2}{c}{size} & \multicolumn{2}{c}{color} &\\
\cmidrule(r){4-5} \cmidrule(r){6-7} \cmidrule(r){8-9} \cmidrule(r){10-11} \cmidrule(r){12-13} \cmidrule(r){14-15} \cmidrule(r){16-17}
& & & \multicolumn{1}{c}{ID\textsubscript{ID}} & \multicolumn{1}{c}{OOD\textsubscript{OOD}} & \multicolumn{1}{c}{ID\textsubscript{ID}} & \multicolumn{1}{c}{OOD\textsubscript{OOD}} & \multicolumn{1}{c}{ID\textsubscript{ID}} & \multicolumn{1}{c}{OOD\textsubscript{OOD}} & \multicolumn{1}{c}{ID\textsubscript{ID}} & \multicolumn{1}{c}{OOD\textsubscript{OOD}} & \multicolumn{1}{c}{ID\textsubscript{ID}} & \multicolumn{1}{c}{OOD\textsubscript{OOD}} & \multicolumn{1}{c}{ID\textsubscript{ID}} & \multicolumn{1}{c}{OOD\textsubscript{OOD}} & \multicolumn{1}{c}{ID\textsubscript{ID}} & \multicolumn{1}{c}{OOD\textsubscript{OOD}} &\\
\cmidrule(r){1-17}

% & & & \multicolumn{7}{c}{scaffold} & \multicolumn{7}{c}{size}\\
% \cmidrule(r){1-17}
\multicolumn{3}{l}{ERM} & \textbf{82.79} & 69.58 & 83.72  & {59.94} & 33.45 & 16.89 & {34.31} & 17.86 & 0.1224  & {0.1995} & 0.1199 & 0.2427 & 77.96  & 28.60 &\\
\multicolumn{3}{l}{IRM} & 81.35 & 67.97 & 81.33 & 59.00 & 33.56 & 16.90 & 34.28 & \textbf{18.05} & 0.1213 & 0.2025 & 0.1222 & {0.2403} & 77.92 & 27.83 &\\
\multicolumn{3}{l}{VREx} & 82.11 & \textbf{70.77} & 83.47 & 58.53 & \textbf{33.88} & \textbf{16.98} & 34.09 & 17.79 & 0.1211 & 0.2094 & 0.1234 & \textbf{0.2384} & 77.98 & 28.48 &\\
\multicolumn{3}{l}{GroupDRO} & {82.60} & {70.64} & 83.79 & 58.98 & {33.81} & {16.98} & 33.95 & 17.59 & \textbf{0.1168} & \textbf{0.1934} & {0.1180} & 0.2423 & 77.98 & {29.07} &\\
\multicolumn{3}{l}{DANN} & 81.18 & 70.63 & {83.90} & 58.68 & 33.63 & 16.90 & 34.17 & 17.86 & 0.1186 & 0.2004 & 0.1188 & 0.2439 & {78.00} & \textbf{29.14} &\\
\multicolumn{3}{l}{Deep Coral} & 82.53 & 68.61 & \textbf{84.70} & \textbf{60.11} & 33.47 & 16.93 & \textbf{34.49} & {17.94} & {0.1185} & 0.2036 & \textbf{0.1134} & 0.2505 & \textbf{78.64} & 29.05 &\\
\cmidrule(r){1-17}
\multicolumn{3}{l}{Mixup} & 82.29 & 68.88 & 83.16 & 59.03 & 30.22 & 16.59 & 30.63 & 17.06 & 0.1279 & 0.2240 & 0.1255 & 0.2748 & 77.40 & 26.47 &\\
\multicolumn{3}{l}{DIR} & 82.54 & 67.47 & 80.46 & 57.11 & 32.55 & 14.98 & 32.89 & 16.61 & 0.3799 & 0.6493 & 0.1541 & 0.5482 & 31.09 & 20.60 &\\

\midrule[1.5pt]
\multicolumn{3}{c}{\multirow{3}{*}{covariate}} & \multicolumn{4}{c}{GOOD-Motif\textuparrow} & \multicolumn{2}{c}{GOOD-SST2\textuparrow} & & \multicolumn{1}{c}{\multirow{3}{*}{concept}} & \multicolumn{4}{c}{GOOD-Motif\textuparrow} & \multicolumn{2}{c}{GOOD-SST2\textuparrow} & \multirow{11}{*}{graph}\\
\cmidrule(r){4-7} \cmidrule(r){8-9} \cmidrule(r){12-15}\cmidrule(r){16-17}

& & & \multicolumn{2}{c}{base} & \multicolumn{2}{c}{size} & \multicolumn{2}{c}{length} & & & \multicolumn{2}{c}{base} & \multicolumn{2}{c}{size} & \multicolumn{2}{c}{length} &\\
\cmidrule(r){4-5} \cmidrule(r){6-7} \cmidrule(r){8-9} \cmidrule(r){12-13}\cmidrule(r){14-15}\cmidrule(r){16-17}
& & & \multicolumn{1}{c}{ID\textsubscript{ID}} & \multicolumn{1}{c}{OOD\textsubscript{OOD}} & \multicolumn{1}{c}{ID\textsubscript{ID}} & \multicolumn{1}{c}{OOD\textsubscript{OOD}} & \multicolumn{1}{c}{ID\textsubscript{ID}} & \multicolumn{1}{c}{OOD\textsubscript{OOD}} & & & \multicolumn{1}{c}{ID\textsubscript{ID}} & \multicolumn{1}{c}{OOD\textsubscript{OOD}} & \multicolumn{1}{c}{ID\textsubscript{ID}} & \multicolumn{1}{c}{OOD\textsubscript{OOD}} & \multicolumn{1}{c}{ID\textsubscript{ID}} & \multicolumn{1}{c}{OOD\textsubscript{OOD}} &\\
\cmidrule(r){1-9} \cmidrule(r){11-17}

\multicolumn{3}{l}{ERM} & 92.60 & 68.66 & {92.28} & 51.74 & \textbf{89.82} & 81.30 & & \multicolumn{1}{l}{ERM} & {92.02} & {81.44} & {91.73} & \textbf{70.75} & \textbf{94.43} & 72.43 &\\
\multicolumn{3}{l}{IRM} & 92.60 & {70.65} & 92.18 & 51.41 & 89.41 & 79.91 & & \multicolumn{1}{l}{IRM} & 92.00 & 80.71 & 91.68 & 69.77 & 94.10 & \textbf{77.47} &\\
\multicolumn{3}{l}{VREx} & 92.60 & \textbf{71.47} & 92.25 & \textbf{52.67} & 89.51 & 80.64 & & \multicolumn{1}{l}{VREx} & \textbf{92.05} & \textbf{81.56} & 91.67 & 70.24 & 94.26 & 73.16 &\\
\multicolumn{3}{l}{GroupDRO} & {92.61} & 68.24 & \textbf{92.29} & {51.95} & 89.59 & \textbf{81.35} & & \multicolumn{1}{l}{GroupDRO} & 92.01 & 81.43 & 91.67 & 69.98 & 94.41 & 71.86 &\\
\multicolumn{3}{l}{DANN} & 92.60 & 65.47 & 92.23 & 51.46 & 89.60 & 79.71 & & \multicolumn{1}{l}{DANN} & {92.02} & 81.33 & \textbf{91.81} & {70.72} & 94.02 & 76.03 &\\
\multicolumn{3}{l}{Deep Coral} & {92.61} & 68.88 & 92.22 & 50.97 & 89.68 & 79.81 & & \multicolumn{1}{l}{Deep Coral} & 92.01 & 81.37 & 91.68 & 70.49 & 94.25 & 72.34 &\\
\cmidrule(r){1-9} \cmidrule(r){11-17}
\multicolumn{3}{l}{Mixup} & \textbf{92.68} & 70.08 & 92.02 & 51.48 & 89.78 & 80.88 & & \multicolumn{1}{l}{Mixup} & 91.89 & 77.63 & 91.45 & 67.81 & 94.12 & 73.34 &\\
\multicolumn{3}{l}{DIR} & 87.73 & {61.50} & 84.53 & 50.41 & 84.30 & 77.65 & & \multicolumn{1}{l}{DIR} & 91.60 & 72.14 & 73.10 & 56.28 & 93.71 & 68.76 &\\

\midrule[1.5pt]
\multicolumn{3}{c}{\multirow{3}{*}{concept}} & \multicolumn{4}{c}{GOOD-HIV\textuparrow} & \multicolumn{4}{c}{GOOD-PCBA\textuparrow} & \multicolumn{4}{c}{GOOD-ZINC\textdownarrow} & \multicolumn{2}{c}{GOOD-CMNIST\textuparrow} & \multirow{11}{*}{graph}\\
\cmidrule(r){4-7} \cmidrule(r){8-11} \cmidrule(r){12-15} \cmidrule(r){16-17}

& & & \multicolumn{2}{c}{scaffold} & \multicolumn{2}{c}{size} & \multicolumn{2}{c}{scaffold} & \multicolumn{2}{c}{size} & \multicolumn{2}{c}{scaffold} & \multicolumn{2}{c}{size} & \multicolumn{2}{c}{color}  &\\
\cmidrule(r){4-5} \cmidrule(r){6-7} \cmidrule(r){8-9} \cmidrule(r){10-11} \cmidrule(r){12-13} \cmidrule(r){14-15} \cmidrule(r){16-17}
& & & \multicolumn{1}{c}{ID\textsubscript{ID}} & \multicolumn{1}{c}{OOD\textsubscript{OOD}} & \multicolumn{1}{c}{ID\textsubscript{ID}} & \multicolumn{1}{c}{OOD\textsubscript{OOD}} & \multicolumn{1}{c}{ID\textsubscript{ID}} & \multicolumn{1}{c}{OOD\textsubscript{OOD}} & \multicolumn{1}{c}{ID\textsubscript{ID}} & \multicolumn{1}{c}{OOD\textsubscript{OOD}} & \multicolumn{1}{c}{ID\textsubscript{ID}} & \multicolumn{1}{c}{OOD\textsubscript{OOD}} & \multicolumn{1}{c}{ID\textsubscript{ID}} & \multicolumn{1}{c}{OOD\textsubscript{OOD}} & \multicolumn{1}{c}{ID\textsubscript{ID}} & \multicolumn{1}{c}{OOD\textsubscript{OOD}}  &\\
\cmidrule(r){1-17}

\multicolumn{3}{l}{ERM} & {84.22} & 72.33 & 88.05 & 63.26 & 25.95 & 21.63 & 32.54 & 15.36 & 0.1222 & 0.1306 & 0.1315 & 0.1403 & 90.00 & 42.87  &\\
\multicolumn{3}{l}{IRM} & 82.89 & 72.59 & \textbf{88.62} & 59.90 & 25.89 & 21.22 & {32.99} & {16.07} & 0.1225 & 0.1314 & 0.1278 & {0.1368} & {90.02} & 42.80  &\\
\multicolumn{3}{l}{VREx} & 83.84 & 72.60 & {88.28} & 60.23 & \textbf{26.62} & {22.02} & 32.49 & 15.59 & {0.1186} & {0.1270} & 0.1309 & 0.1419 & 89.99 & {43.31}  &\\
\multicolumn{3}{l}{GroupDRO} & 83.40 & \textbf{73.64} & {88.28} & 61.37 & 26.32 & 21.83 & \textbf{33.03} & 15.99 & 0.1207 & 0.1281 & \textbf{0.1251} & 0.1369 & \textbf{90.02} & \textbf{43.32}  &\\
\multicolumn{3}{l}{DANN} & 83.87 & 71.92 & 87.28 & {65.27} & 26.07 & 21.64 & 32.74 & 15.78 & \textbf{0.1172} & \textbf{0.1256} & {0.1253} & \textbf{0.1339} & 89.94 & 43.11  &\\
\multicolumn{3}{l}{Deep Coral} & \textbf{84.65} & {72.97} & 87.88 & 62.28 & {26.38} & {21.95} & 32.67 & {16.20} & 0.1187 & 0.1279 & 0.1287 & 0.1370 & 89.94 & 43.16  &\\
\cmidrule(r){1-17}
\multicolumn{3}{l}{Mixup} & 82.36 & 72.03 & 87.64 & {64.87} & 23.73 & 19.78 & 30.23 & 13.36 & 0.1353 & 0.1475 & 0.1423 & 0.1522 & 89.95 & 40.96  &\\
\multicolumn{3}{l}{DIR} & 83.28 & 69.05 & 79.19 & \textbf{72.61} & 25.85 & \textbf{22.20} & 30.53 & \textbf{16.86} & 0.3501 & 0.3865 & 0.2348 & 0.2871 & 86.76 & 22.69  &\\

\midrule[1pt]
\midrule[1pt]
\multicolumn{3}{c}{\multirow{3}{*}{covariate}} & \multicolumn{4}{c}{GOOD-Cora\textuparrow} & \multicolumn{4}{c}{GOOD-Arxiv\textuparrow} & \multicolumn{2}{c}{GOOD-CBAS\textuparrow} & \multicolumn{2}{c}{GOOD-Twitch\textuparrow} & \multicolumn{2}{c}{GOOD-WebKB\textuparrow} & \multirow{11}{*}{node}\\
\cmidrule(r){4-7} \cmidrule(r){8-11} \cmidrule(r){12-13} \cmidrule(r){14-15} \cmidrule(r){16-17}

& & & \multicolumn{2}{c}{word} & \multicolumn{2}{c}{degree} & \multicolumn{2}{c}{time} & \multicolumn{2}{c}{degree} & \multicolumn{2}{c}{color} & \multicolumn{2}{c}{language} & \multicolumn{2}{c}{university}  &\\
\cmidrule(r){4-5} \cmidrule(r){6-7} \cmidrule(r){8-9} \cmidrule(r){10-11} \cmidrule(r){12-13} \cmidrule(r){14-15} \cmidrule(r){16-17}
& & & \multicolumn{1}{c}{ID\textsubscript{ID}} & \multicolumn{1}{c}{OOD\textsubscript{OOD}} & \multicolumn{1}{c}{ID\textsubscript{ID}} & \multicolumn{1}{c}{OOD\textsubscript{OOD}} & \multicolumn{1}{c}{ID\textsubscript{ID}} & \multicolumn{1}{c}{OOD\textsubscript{OOD}} & \multicolumn{1}{c}{ID\textsubscript{ID}} & \multicolumn{1}{c}{OOD\textsubscript{OOD}} & \multicolumn{1}{c}{ID\textsubscript{ID}} & \multicolumn{1}{c}{OOD\textsubscript{OOD}} & \multicolumn{1}{c}{ID\textsubscript{ID}} & \multicolumn{1}{c}{OOD\textsubscript{OOD}} & \multicolumn{1}{c}{ID\textsubscript{ID}} & \multicolumn{1}{c}{OOD\textsubscript{OOD}}  &\\
\cmidrule(r){1-17}

\multicolumn{3}{l}{ERM} & 70.43 & {64.86} & 72.27 & 56.30 & {72.69} & 71.08 & 77.47 & 58.91 & 89.29 & 76.00 & 70.66 & 48.95 & 38.25 & 14.29 &\\
\multicolumn{3}{l}{IRM} & 70.27 & 64.77 & {72.64} & 56.28 & 72.66 & 71.04 & 77.50 & 58.98 & 91.00 & 76.00 & 69.75 & 47.21 & 39.34 & 13.49 &\\
\multicolumn{3}{l}{VREx} & 70.47 & 64.80 & 72.25 & 56.30 & 72.66 & 71.12 & 77.49 & 58.99 & \textbf{91.14} & 77.14 & 70.66 & 48.99 & 39.34 & 14.29 &\\
\multicolumn{3}{l}{GroupDRO} & 70.41 & 64.72 & 72.18 & 56.29 & 72.68 & {71.15} & 77.46 & \textbf{59.08} & 90.86 & 76.14 & 70.84 & 47.20 & 39.89 & 17.20 &\\
\multicolumn{3}{l}{DANN} & {70.66} & 64.77 & 72.47 & 56.10 & \textbf{72.74} & 71.05 & {77.51} & {59.00} & 90.14 & \textbf{77.57} & 70.67 & 48.98 & 39.89 & 15.08 &\\
\multicolumn{3}{l}{Deep Coral} & 70.47 & 64.72 & 72.16 & {56.35} & 72.66 & 71.07 & 77.48 & 58.97 & \textbf{91.14} & 75.86 & 70.67 & 49.64 & 38.25 & 13.76 &\\
\cmidrule(r){1-17}
\multicolumn{3}{l}{Mixup} & \textbf{71.54} & \textbf{65.23} & \textbf{74.57} & \textbf{58.20} & 72.49 & \textbf{71.34} & \textbf{77.61} & 57.60 & 73.57 & 70.57 & 71.30 & \textbf{52.27} & \textbf{54.65} & 17.46 &\\
\multicolumn{3}{l}{EERM} & 68.79 & 61.98 & 73.32 & 56.88 & OOM & OOM & OOM & OOM & 67.62 & 52.86 & \textbf{73.87} & 51.34 & 46.99 & \textbf{24.61} &\\
\multicolumn{3}{l}{SRGNN} & 70.27 & 64.66 & 71.37 & 54.78 & 72.50 & 70.83 & 75.96 & 57.52 & 77.62 & 74.29 & 70.58 & 47.30 & 39.89 & 13.23 &\\

\midrule[1.5pt]
\multicolumn{3}{c}{\multirow{3}{*}{concept}} & \multicolumn{4}{c}{GOOD-Cora\textuparrow} & \multicolumn{4}{c}{GOOD-Arxiv\textuparrow} & \multicolumn{2}{c}{GOOD-CBAS\textuparrow} & \multicolumn{2}{c}{GOOD-Twitch\textuparrow} & \multicolumn{2}{c}{GOOD-WebKB\textuparrow} & \multirow{11}{*}{node}\\
\cmidrule(r){4-7} \cmidrule(r){8-11} \cmidrule(r){12-13} \cmidrule(r){14-15} \cmidrule(r){16-17}

& & & \multicolumn{2}{c}{word} & \multicolumn{2}{c}{degree} & \multicolumn{2}{c}{time} & \multicolumn{2}{c}{degree} & \multicolumn{2}{c}{color} & \multicolumn{2}{c}{language} & \multicolumn{2}{c}{university}  &\\
\cmidrule(r){4-5} \cmidrule(r){6-7} \cmidrule(r){8-9} \cmidrule(r){10-11} \cmidrule(r){12-13} \cmidrule(r){14-15} \cmidrule(r){16-17}
& & & \multicolumn{1}{c}{ID\textsubscript{ID}} & \multicolumn{1}{c}{OOD\textsubscript{OOD}} & \multicolumn{1}{c}{ID\textsubscript{ID}} & \multicolumn{1}{c}{OOD\textsubscript{OOD}} & \multicolumn{1}{c}{ID\textsubscript{ID}} & \multicolumn{1}{c}{OOD\textsubscript{OOD}} & \multicolumn{1}{c}{ID\textsubscript{ID}} & \multicolumn{1}{c}{OOD\textsubscript{OOD}} & \multicolumn{1}{c}{ID\textsubscript{ID}} & \multicolumn{1}{c}{OOD\textsubscript{OOD}} & \multicolumn{1}{c}{ID\textsubscript{ID}} & \multicolumn{1}{c}{OOD\textsubscript{OOD}} & \multicolumn{1}{c}{ID\textsubscript{ID}} & \multicolumn{1}{c}{OOD\textsubscript{OOD}}  &\\
\cmidrule(r){1-17}
\multicolumn{3}{l}{ERM} & 66.05 & {64.60} & {68.71} & 60.54 & 74.76 & 67.32 & \textbf{75.27} & {62.99} & 89.79 & 82.36 & 80.29 & 57.32 & 65.00 & 27.83 &\\
\multicolumn{3}{l}{IRM} & 66.09 & {64.60} & 68.58 & {61.23} & 74.67 & 67.41 & 75.23 & 62.97 & 90.71 & \textbf{83.21} & 77.05 & 59.17 & 65.56 & 27.52 &\\
\multicolumn{3}{l}{VREx} & 66.00 & 64.57 & 68.45 & 60.58 & {74.80} & 67.37 & 75.19 & \textbf{63.00} & 89.50 & 82.86 & 80.29 & 57.37 & 65.00 & 27.83 &\\
\multicolumn{3}{l}{GroupDRO} & {66.17} & \textbf{64.62} & 68.37 & 60.65 & 74.73 & \textbf{67.45} & 75.19 & 62.88 & 90.36 & 82.00 & 81.95 & \textbf{60.27} & 65.00 & 28.14 &\\
\multicolumn{3}{l}{DANN} & 66.16 & 64.51 & 68.08 & 60.78 & 74.73 & 67.28 & {75.25} & 62.91 & 89.93 & 82.50 & 80.28 & 57.46 & 65.00 & 26.91 &\\
\multicolumn{3}{l}{Deep Coral} & 66.13 & 64.58 & 68.38 & 60.58 & 74.77 & {67.42} & 75.16 & 62.85 & 89.36 & 82.64 & 80.14 & 56.97 & 65.00 & 28.75 &\\
\cmidrule(r){1-17}
\multicolumn{3}{l}{Mixup} & \textbf{69.66} & 64.44 & \textbf{70.32} & \textbf{63.65} & \textbf{74.92} & 64.84 & 72.75 & 61.28 & \textbf{93.64} & 64.57 & 78.89 & 55.28 & \textbf{67.22} & \textbf{31.19} &\\
\multicolumn{3}{l}{EERM} & 65.75 & 63.09 & 66.50 & 58.38 & OOM & OOM & OOM & OOM & 78.33 & 64.29 & \textbf{83.91} & 51.94 & 61.67 & 27.83 &\\
\multicolumn{3}{l}{SRGNN} & 66.45 & \textbf{64.62} & 68.34 & 61.08 & 74.64 & 67.17 & 74.83 & 62.09 & 88.57 & 81.43 & 80.21 & 56.05 & 61.67 & 27.52 &\\

\bottomrule[2pt]
\end{tabular}
}
\vspace{0.4cm}
% }
\caption{ID\textsubscript{ID} and OOD\textsubscript{OOD} performances of 10 baselines on 11 datasets.
All numerical results are averages across 3 to 10 random runs. Numbers in \textbf{bold} represent the best results. OOM denotes out of memory. Additional results are in Appendix~\ref{sec:D}. More empirical results and analysis are in Appendix~\ref{sec:C}.
}
\label{table:OODalgs}
% \vspace{-10pt}
\end{table*}

\subsubsection{Baseline methods}
We consider empirical risk minimization (ERM) and 9 OOD algorithms as baselines, among which 4 are graph-specific methods.
Firstly, we choose two domain adaptation algorithms that target minimizing feature discrepancies.
DANN~\cite{ganin2016domain} adversarially trains the regular classifier and a domain classifier to make features indistinguishable.
Deep Coral~\cite{sun2016deep} encourages features in different domains to be similar by minimizing the deviation of covariant matrices from different domains.
Furthermore, we adopt two invariant learning baselines following the invariant prediction assumption~\cite{peters2016causal}.
IRM~\cite{arjovsky2019invariant} searches for data representations that perform well across all environments by penalizing feature distributions that have different optimal linear classifiers for each environment.
VREx~\cite{krueger2021out} targets both covariate robustness and the invariant prediction. It specifically reduces the variance of risks in test environments by minimizing the risk variances of training environments. 
By applying fair optimization, GroupDRO~\cite{sagawa2019distributionally} tackles the problem that the distribution minority lacks sufficient training. This method, known as risk interpolation~\cite{krueger2021out}, is achieved by explicitly minimizing the loss in the worst training environment.
To evaluate the performance of current OOD methods specifically designed for graphs, we include the following 4 graph OOD methods.
We incorporate the data augmentation method Mixup~\cite{zhang2017mixup} following the implementation of Mixup-For-Graph\citet{wang2021mixup} designed for graph data, which improves model generalization abilities.
DIR~\cite{wu2022dir} selects a subset of graph representations as causal rationales and conducts interventional data augmentation to create multiple distributions.
EERM~\cite{wu2022handling} tries to generate environments by a REINFORCE algorithm to maximize loss variance between environments, while its main loss minimization requires adversarially minimizing loss variance.
SRGNN~\cite{zhu2021shift} aims at converting the biased training data to the given unbiased distribution, performed through a central moment discrepancy regularizer and the kernel mean matching technique by solving a quadratic problem.

\subsubsection{Quantitative comparison and analysis}

Table~\ref{table:OODalgs} shows the OOD\textsubscript{OOD} and ID\textsubscript{ID} results of 10 baselines for all splits. Most OOD algorithms have comparable performances with ERM, while many OOD algorithms outperform ERM with certain patterns. Specifically, we observe that the risk interpolation (GroupDRO) and extrapolation (VREx) perform favorably against other methods on multiple datasets and shift splits. VREx outperforms other methods on 7 out of 34 OOD splits, evidencing its learning invariance and robustness, especially for covariate shifts in graph prediction tasks. GroupDRO outperforms 8 out of 34 OOD splits, showing its advantage in fair optimization. The two feature discrepancy minimization methods, DANN and Deep Coral, do not perform well enough. DANN outperforms on 4 splits, and it is especially suitable for graph concept shift splits. Deep Coral outperforms on 1 OOD split but usually has advantages on ID tests. Finally, IRM performs similarly to ERM and outperforms on 3 of the OOD results, showing the difficulty of achieving invariant prediction in non-linear settings.

Graph OOD methods make extra effort to interpolate the irregularity and connectivity of graph topology, and certain improvements are achieved. Mixup-For-Graph exclusively excels at node prediction tasks, yielding consistent gains across datasets, which can attribute to its node-specific design~\cite{wang2021mixup}. It outperforms 6 out of 14 node-task OOD splits. However, it fails at graph prediction tasks due to the simple graph representation mixup strategy. DIR specifically solves concept shifts for graph classification tasks and outperforms on 3 splits, indicating that interventional augmentation on representations weakens spurious correlations by diversifying the distribution. Its benefit on concept shift does not apply to covariate shifts since DIR only expands the combination of representations without creating new domains; it also fails on regression tasks which require a more delicate learning process. EERM and SRGNN generally have average performances, outperforming only on a few splits. EERM reveals that while environment generation is learnable with REINFORCE, this adversarial training is difficult and needs to be perfected. SRGNN makes use of our OOD validation data to draw the training data closer to an OOD distribution; however, without sufficient generalization, it can seldom perform well in tests since OOD validation data cannot exactly reflect OOD test data. To conclude, while these graph-specific methods apply well to graph topology, other flaws in the methodology design create a performance bottleneck.

\section{Discussions}\label{sec:discussion}

GOOD aims to facilitate the development of graph OOD and general OOD algorithms. 
Our results and comparisons show that current OOD algorithms can improve generalization abilities, but not significantly. In addition, an algorithm might improve performance on one type of shift, but not both. With these observations, future OOD methods can focus on solving one of covariate and concept shifts to improve the specific generalization ability. The improvement might be achieved by managing well-designed model architectures, optimization schemes, or data augmentation strategies.
Moreover, models cannot be expected to solve unknown distribution shifts. Thus, we believe using the given environment information to convey the types of shifts expected during testing is a promising direction.

Our GOOD benchmark is a growing project. We expect to include more methods as the OOD field develops especially graph-specific algorithms and include datasets and domain selections of a larger quantity and variety.
In addition, the current benchmark does not consider link prediction tasks~\cite{zhou2022ood}, which will be added as the project develops.

% \section*{Reproducibility}

% \newpage
\begin{ack}
We thank Jundong Li and Jing Ma for insightful discussions. This work was supported in part by National Science Foundation
grants IIS-1955189, IIS-1908198, and IIS-1908220.
\end{ack}

\newpage

\bibliographystyle{plainnat}
\bibliography{graphood}

%%%%%%%%%%%%%%%%%%%%%%%%%%%%%%%%%%%%%%%%%%%%%%%%%%%%%%%%%%%%
\newpage

\appendix

\section{GOOD Dataset Details}\label{sec:A}

GOOD provides 11 datasets with 17 domain selections. For each domain selection, we provide two shift splits and a no shift split, leading to 51 splits. For each covariate/concept shift split, we split data into 5 subsets, namely, training set, in-distribution (ID) validation set, in-distribution (ID) test set, out-of-distribution (OOD) validation set, and out-of-distribution (OOD) test set. For no shift splits, we split data into training, ID validation, and ID test sets. The statistics of splits are listed in Table~\ref{tab:split}.
Meanwhile, the datasets in GOOD consist of 8 real-world datasets, 1 semi-artificial dataset, and 2 synthetic datasets, and we specify the details of splits for each category in the following three paragraphs. The 8 real-world datasets are public datasets~\cite{hu2020open, wu2018moleculenet, gomez2018automatic, bojchevski2017deep} and we closely follow the license rules, which are specified in the Appendix~\ref{sec:F}.

\textbf{Real-world datasets.}
For covariate shift splits, given a domain selection, we sort the graphs/nodes by their domains and divide the data into a certain number of domains by specifying the split ratio. Then training, validation, and test sets consist of one or several domains. Also the independence between $Y$ and $X_\text{ind}$ guarantees that covariate shift design does not contain concept shift theoretically. 
For concept shift splits, we adopt a screening process to build the splits. We first explain this screening process for graph prediction datasets. Each concept has specific domain-label correlations, which come in the form of a set of domain-label probabilities. Consequently, to build a specific concept, each graph has a domain-label probability to be included in this concept. Therefore, we build each concept by scanning the whole dataset and selecting graphs to be included according to their probabilities. The selected graphs form the current concept and are excluded from the dataset scanning. We repeat this procedure to form each of the concepts sequentially, and the last concept includes all the remaining graphs. 
Similarly, in node classification tasks, we apply the screening process to nodes instead of graphs. That is, we build node selection masks instead of collecting graphs out of datasets. 
Note that the selection probabilities are relatively similar for those concepts within the training set, while largely dissimilar between the training, validation, and test sets. 
Also, it is difficult to specify all domain-label probabilities for tasks like 70-classes classification in GOOD-Cora, and impossible for regression tasks. Therefore, we design to group labels as only two categories, namely high/low labels. Then we can build concept splits in a clear sense. For example, we assign a high probability for domain $d_i$ with label $0$ in training concepts, while a high probability for domain $d_i$ with label $1$  in test concepts.

\begin{table*}[!t]
% \setlength{\abovecaptionskip}{0.1cm}
% \footnotesize
% \scalebox{0.622}{
% % \setlength{\tabcolsep}{0.3mm}{
% \centering
\centering
\resizebox{\textwidth}{!}
{
\begin{tabular}{llccccccccccc}
\toprule[2pt]

\multicolumn{2}{l}{Dataset} & \multicolumn{1}{c}{Shift} & \multicolumn{1}{c}{Train} & \multicolumn{1}{c}{ID validation} & \multicolumn{1}{c}{ID test} & \multicolumn{1}{c}{OOD validation} & \multicolumn{1}{c}{OOD test} & \multicolumn{1}{c}{Train} & \multicolumn{1}{c}{OOD validation} & \multicolumn{1}{c}{ID validation} & \multicolumn{1}{c}{ID test} & \multicolumn{1}{c}{OOD test} \\
\midrule[1pt]
& & &  \multicolumn{5}{c}{Scaffold} & \multicolumn{4}{c}{Size}\\ 
\cmidrule(r){4-8} \cmidrule(r){9-13}
\multicolumn{2}{l}{\multirow{3}{*}{GOOD-HIV}} & covariate & 24682 & 4112 & 4112 & 4113 & 4108  & 26169 & 4112 & 4112 & 2773 & 3961 \\
& & concept & 15209 & 3258 & 3258 & 9365 & 10037  & 14454 & 3096 & 3096 & 9956 & 10525 \\
& & no shift & 24676 & 8225 & 8226 & - & - & 24676 & 8225 & 8226 & - & -  \\
\midrule[1pt]
& & & \multicolumn{5}{c}{Scaffold} & \multicolumn{4}{c}{Size}\\ 
\cmidrule(r){4-8} \cmidrule(r){9-13}
\multicolumn{2}{l}{\multirow{3}{*}{GOOD-PCBA}} & covariate & 262764 & 43792 & 43792 & 44019 & 43562   & 269990 & 43792 & 43792 & 48430 & 31925  \\
& & concept & 159158 & 34105 & 34105 & 90740 & 119821   & 150121 & 32168 & 32168 & 108267 & 115205  \\
& & no shift & 262757 & 87586 & 87586 & - & - & 262757 & 87586 & 87586 & - & -   \\
\midrule[1pt]
& & & \multicolumn{5}{c}{Scaffold} & \multicolumn{4}{c}{Size}\\ 
\cmidrule(r){4-8} \cmidrule(r){9-13}
\multicolumn{2}{l}{\multirow{3}{*}{GOOD-ZINC}} & covariate & 149674 & 24945 & 24945 & 24945 & 24946 & 161893 & 24945 & 24945 & 20270 & 17402 \\
& & concept & 101867 & 21828 & 21828 & 43539 & 60393 & 89418 & 19161 & 19161 & 51409 & 70306 \\
& & no shift & 149673 & 49891 & 49891 & - & - & 149673 & 49891 & 49891 & - & -  \\
\midrule[1pt]
& & & \multicolumn{5}{c}{Length} & \\ 
\cmidrule(r){4-8} 
\multicolumn{2}{l}{\multirow{3}{*}{GOOD-SST2}} & covariate & 24744 & 5301 & 5301 & 17206 & 17490  &  &  &  &  & \\
& & concept & 27270 & 5843 & 5843 & 15142 & 15944 &  &  &  &  & \\
& & no shift & 42025 & 14008 & 14009 & - & - &  &  &  &  & \\
\midrule[1pt]
& & & \multicolumn{5}{c}{Color} & \\ 
\cmidrule(r){4-8} 
\multicolumn{2}{l}{\multirow{3}{*}{GOOD-CMNIST}} & covariate & 42000 & 7000 & 7000 & 7000 & 7000  &  &  &  &  & \\
& & concept & 29400 & 6300 & 6300 & 14000 & 14000 &  &  &  &  & \\
& & no shift & 42000 & 14000 & 14000 & - & - &  &  &  &  & \\
\midrule[1pt]
& & & \multicolumn{5}{c}{Base} & \multicolumn{4}{c}{Size}\\ 
\cmidrule(r){4-8} \cmidrule(r){9-13}
\multicolumn{2}{l}{\multirow{3}{*}{GOOD-Motif}} & covariate & 18000 & 3000 & 3000 & 3000 & 3000  & 18000 & 3000 & 3000 & 3000 & 3000 \\
& & concept & 12600 & 2700 & 2700 & 6000 & 6000  & 12600 & 2700 & 2700 & 6000 & 6000 \\
& & no shift & 18000 & 6000 & 6000 & - & - & 18000 & 6000 & 6000 & - & - \\
\midrule[1pt]
& & & \multicolumn{5}{c}{Word} & \multicolumn{4}{c}{Degree}\\ 
\cmidrule(r){4-8} \cmidrule(r){9-13}
\multicolumn{2}{l}{\multirow{3}{*}{GOOD-Cora}} & covariate & 9378& 1979& 1979& 3003& 3454  & 8213& 1979& 1979& 3841& 3781 \\
& & concept & 7273& 1558& 1558& 3807& 5597  & 7281& 1560& 1560& 3706& 5686 \\
& & no shift & 11875 & 3959 & 3959 & - & - & 11875 & 3959 & 3959 & - & -  \\
\midrule[1pt]
& & & \multicolumn{5}{c}{Time} & \multicolumn{4}{c}{Degree}\\ 
\cmidrule(r){4-8} \cmidrule(r){9-13}
\multicolumn{2}{l}{\multirow{3}{*}{GOOD-Arxiv}} & covariate & 57073& 16934& 16934& 29799& 48603 & 68607& 16934& 16934& 46264& 20604   \\
& & concept & 62083& 13303& 13303& 32560& 48094 & 58619& 12561& 12561& 34222& 51380 \\
& & no shift & 101605 & 33869 & 33869 & - & - & 101605 & 33869 & 33869 & - & -  \\
\midrule[1pt]
& & & \multicolumn{5}{c}{Language} & \\ 
\cmidrule(r){4-8} 
\multicolumn{2}{l}{\multirow{3}{*}{GOOD-Twitch}} & covariate & 14448& 3412& 3412& 6551& 6297  &  &  &  &  & \\
& & concept & 13605& 2914& 2914& 6762& 7925  &  &  &  &  & \\
& & no shift & 13648 & 10236 & 10236 & - & -   &  &  &  &  & \\
\midrule[1pt]
& & & \multicolumn{5}{c}{University} & \\ 
\cmidrule(r){4-8} 
\multicolumn{2}{l}{\multirow{3}{*}{GOOD-WebKB}} & covariate & 244& 61& 61& 125& 126  &  &  &  &  & \\
& & concept & 282& 60& 60& 106& 109  &  &  &  &  & \\
& & no shift & 370 & 123 & 124 & - & -   &  &  &  &  & \\
\midrule[1pt]
& & & \multicolumn{5}{c}{Color} & \\ 
\cmidrule(r){4-8} 
\multicolumn{2}{l}{\multirow{3}{*}{GOOD-CBAS}} & covariate & 420& 70& 70& 70& 70  &  &  &  &  & \\
& & concept & 140& 140& 140& 140& 140  &  &  &  &  & \\
& & no shift & 420 & 140 & 140 & - & -   &  &  &  &  & \\

\bottomrule[2pt]
\end{tabular}
}
\caption{Numbers of graphs/nodes in training, ID validation, ID test, OOD validation, and OOD test sets for the 11 datasets.}\label{tab:split}
\label{table:split}
\end{table*}

\textbf{Semi-artificial datasets.}
For semi-artificial datasets, we firstly define a domain/concept and modify graph attributes according to the assigned domain/concept. Due to the difficulty in modifying graph structures without breaking the semantics of the original graphs, we choose to modify or append the features of nodes in graphs. GOOD currently corporate one semi-artificial dataset GOOD-CMNIST. For GOOD-CMNIST, since the original colors of graphs are in gray-scale, we color graphs by setting node features as 3-channel RGB colors such as red, blue, and cyan. For covariate shift, the training graphs contain 5 color domains, thus forming 5 environments. Other than these 5 colors, the validation and test set graphs are in different colors, respectively. Therefore, in total, we produce 7 different node color features. For concept shift, each digit label is associated with one color, \eg, 0 with red, 1 with green, 2 with blue, etc. Hence we generate 10 different node color features to match the 10-class labels.

\textbf{Synthetic datasets.}
For GOOD-Motif, we generate graphs using five label-independent base graphs (wheel, tree, ladder, star, and path) and three label-dependent motifs (house, cycle, and crane). We select the base graph type and the size as domain features to create covariate and concept splits. In the covariate shift splits with base domain, the training set includes graphs with the first three bases, while the validation and the test sets include graphs with base star and path, respectively. In the concept splits with base domain, for 3 different concepts in the training set, each motif is highly correlated to a specific base graph with different correlation rates; \ie, the house-wheel, cycle-tree, and crane-ladder correlations in the training concepts have high probabilities of $99\%$, $97\%$, and $95\%$. In contrast, in the validation and the test sets, these correlations are weak and nonexistent, respectively. Note that only three base graphs are used in this concept shift. 
In both shift splits with size domain, the base graphs match motifs randomly, while the sizes of base graphs differ. Given five size ranges, in covariate splits, the training set contains three small sizes, while the validation and the test sets include the middle and the largest size ranges, respectively. In concept splits, there are three size ranges which have high, weak, and no correlations with labels for the training, validation, and test sets, respectively. 
GOOD-CBAS is a color domain dataset with a similar color strategy as GOOD-CMNIST. The main difference in the coloring process is that GOOD-CBAS adopts 4-channel RGBA colors instead of 3-channel colors.

More details about split algorithms can be found in \url{https://github.com/divelab/GOOD/tree/main/GOOD/data/good_datasets}.

\textbf{Discussions of the environment variable in the causal graph.} In covariate shift, the environment variable $E$ is only associated with $X_\text{ind}$. According to the split processes in Section~3, $E\rightarrow X_\text{ind}$ in synthetic/semi-artificial datasets, while $E\leftarrow X_\text{ind}$ in real-world datasets, where $\rightarrow$ denotes a causal mapping. In concept shift, $E$ is correlated with both $X_\text{ind}$ and $Y$. In synthetic datasets, $E$ is a confounder, \ie, $Y\leftarrow E \rightarrow X_\text{ind}$. In semi-artificial datasets, $Y\rightarrow E \rightarrow X_\text{ind}$. In real-world datasets, $Y\rightarrow E \leftarrow X_\text{ind}$.

\section{Experimental Details}\label{sec:B}
We conduct experiments on 11 datasets, 51 shift splits, with 10 baseline methods. 
For graph prediction and node prediction tasks, we respectively select strong and commonly acknowledged GNN backbones.  
For each dataset, we use the same GNN backbone for all baseline methods for fair comparison.
For graph prediction tasks, we use GIN-Virtual Node~\cite{xu2018how, gilmer2017neural} as the GNN backbone. As an exception, for GOOD-Motif we adopt GIN~\cite{xu2018how} as the GNN backbone, since we observe from experiments that the global information provided by virtual nodes would interrupt the training process here.
For node prediction tasks, we adopt GraphSAINT~\cite{Zeng2020GraphSAINT:} and use GCN~\cite{kipf2017semi} as the GNN backbone.
Note that the GNN backbone for Mixup is a modified GCN according to the implementation of~\citet{wang2021mixup}. 

Our code is implemented based on PyTorch Geometric~\cite{fey2019fast}.
For all the experiments, we use the Adam optimizer, with a weight decay of 0 and a dropout rate of 0.5. The GNN model and the number of convolutional layers for each dataset are specified in Table~\ref{table:para1}. We use mean global pooling and the RELU activation function, and the dimension of the hidden layer is 300. The batch size, the maximum number of epochs, (the number of iterations per epoch for node prediction tasks,) and initial learning rate are also specified in Table~\ref{table:para1}. In the training process, all models are trained to converge. For computation, we generally use one NVIDIA GeForce RTX 2080 Ti for each single experiment. However, the graph OOD algorithm EERM encounters CUDA out of memory, due to its high memory requirement.
\begin{table*}[t]
\setlength{\abovecaptionskip}{0.1cm}
\footnotesize
\scalebox{0.78}{
\centering
\begin{tabular}{lcccccc}
\toprule[2pt]
Dataset & model & \# model layers & batch size & \# max epochs & \# iterations per epoch & initial learning rate\\
\midrule[1pt]
GOOD-HIV & GIN-Virtual & 3 & 32 & 200 & -- & 1e-3\\
GOOD-PCBA & GIN-Virtual & 5 & 32 & 200 & -- & 1e-3\\
GOOD-ZINC & GIN-Virtual & 3 & 32 & 200 & -- & 1e-3\\
GOOD-SST2 & GIN-Virtual & 3 & 32 & 200/100 & -- & 1e-3\\
GOOD-CMNIST & GIN-Virtual & 5 & 128 & 500 & -- & 1e-3\\
GOOD-Motif & GIN & 3 & 32 & 200 & -- & 1e-3\\
GOOD-Cora & GCN & 3 & 4096 & 100 & 10 & 1e-3\\
GOOD-Arxiv & GCN & 3 & 4096 & 100 & 100 & 1e-3\\
GOOD-Twitch & GCN & 3 & 4096 & 100 & 10 & 1e-3/5e-3\\
GOOD-WebKB & GCN & 3 & 4096 & 100 & 10 & 1e-3/5e-3\\
GOOD-CBAS & GCN & 3 & 1000 & 200 & 10 & 3e-3\\
\bottomrule[2pt]
\end{tabular}
}
% }
\caption{General model and hyperparameters for 11 datasets. Specifically, GOOD-SST2 uses 100 max epochs for DIR and 200 for the rest of the methods; GOOD-Twitch and GOOD-WebKB uses an initial learning rate of 5e-3 for EERM and 1e-3 for the rest of the methods.}
\label{table:para1}
\end{table*}

\textbf{Hyperparameters for OOD algorithms.}
For each OOD algorithm, we choose one or two algorithm-specific hyperparameter to tune. For IRM and Deep Coral, we tune the weight for penalty loss. For VREx, we tune the weight for VREx's loss variance penalty. For GroupDRO, we tune the step size. For DANN, we tune the weight for domain classification penalty loss. For Mixup, we tune the alpha value of its Beta function. The Beta function is used to randomize the lamda weight, which is the weight for mixing two instances up. For DIR, we tune the causal ratio for selecting causal edges. For EERM, we tune the learning rate for reinforcement learning and the beta value to trade off between mean and variance. For SRGNN, we tune the weight for shift-robust loss calculated by central moment discrepancy.
For each split of a dataset and each OOD algorithm, we search from a hyperparameter set of 3 to 8 values and select the optimal one based on validation metric scores. 
The hyperparameter sets and the optimal hyperparameters are listed in Appendix~\ref{sec:E}.

\textbf{Reproducibility.} For all experiments, we select the best checkpoints for ID and OOD tests according to ID and OOD validation sets, and report the results. All the datasets, codes, and best checkpoints to reproduce the results in this paper are available at \url{https://github.com/divelab/GOOD/}. Simple usage guideline and examples are as Appendix~\ref{sec:F}. For coding details and instructions, please refer to the GOOD package documents~\url{https://good.readthedocs.io}.

\section{Empirical Results and Analysis}\label{sec:C}
We analyze empirical results based on the numerical results in Appendix~\ref{sec:D}. 
Notations are the same as in the main paper.
By comparing ID\textsubscript{ID} with OOD\textsubscript{ID}, and ID\textsubscript{OOD} with OOD\textsubscript{OOD} results, we can observe substantial and consistent gaps between both pairs of ID/OOD performances. In all cases, the OOD performance is significantly worse than the corresponding ID performance,  demonstrating that all our splits meaningfully produce distribution shifts. 
For most splits, the OOD\textsubscript{ID} performance is worse than the OOD\textsubscript{OOD} performance. This implies that OOD validation sets outperform ID validation sets in selecting models with better generalization ability, since the OOD validation set contains similar distribution shifts as the OOD test set. 
However, this is not always the case, since models do not possess sufficient generalization ability, and cannot always deal with distribution shifts during test even these shifts are similar to that during validation.
In addition, for the no shift random split, where only ID setting exists, performances are comparable with covariate/concept ID\textsubscript{ID} settings but constantly a bit worse; this is explainable in the sense that no shift splits include more unfiltered OOD data, and the greater diversity of data adds to training difficulty.

In most cases, algorithms have comparable performances on the same split. Many OOD algorithms outperform ERM with certain patterns, and the number of outperforming cases reveals essential information about the generalization ability of an algorithm. As mentioned in Section 5.2 of the main paper, the risk interpolation (GroupDRO) and extrapolation (VREx) perform favorably against other methods on multiple datasets and shift splits.VREx outperforms other methods on 7 out of 34 OOD splits, evidencing its learning invariance and robustness, especially for covariate shifts in graph prediction tasks. GroupDRO outperforms on 8 out of 34 OOD splits, showing its advantage in fair optimization. The two feature discrepancy minimization methods, DANN and Deep Coral, do not perform well enough. DANN outperforms on 4 splits, and it is especially suitable for graph concept shift splits. Deep Coral outperforms on 1 OOD split but usually has advantages on ID tests. Finally, IRM performs similarly to ERM and outperforms on 3 of the OOD results, showing the difficulty of achieving invariant prediction in non-linear settings.

Graph OOD methods make extra effort to interpolate the irregularity and connectivity of graph topology, and certain improvements are achieved. Mixup-For-Graph exclusively excels at node prediction tasks, yielding consistent gains across datasets, which can attribute to its node-specific design~\cite{wang2021mixup}. It outperforms 6 out of 14 node-task OOD splits. However, it fails at graph prediction tasks due to the simple graph representation mixup strategy. DIR specifically solves concept shifts for graph classification tasks and outperforms on 3 splits, indicating that interventional augmentation on representations weakens spurious correlations by diversifying the distribution. Its benefit on concept shift does not apply to covariate shifts since DIR only expands the combination of representations without creating new domains; it also fails on regression tasks which require a more delicate learning process. EERM and SRGNN generally have average performances, outperforming only on a few splits. EERM reveals that while environment generation is learnable with REINFORCE, this adversarial training is difficult and needs to be perfected. SRGNN makes use of our OOD validation data to draw the training data closer to an OOD distribution; however, without sufficient generalization, it can seldom perform well in tests since OOD validation data cannot exactly reflect OOD test data. To conclude, while these graph-specific methods apply well to graph topology, other flaws in the methodology design create a performance bottleneck.

When OOD algorithms achieve good performance on certain splits, they usually cannot perform equally well in the corresponding ID settings. This phenomenon reveals the OOD-specific generalization ability of these algorithms. In contrast, Mixup, the data augmentation method, performs equally well in both OOD and ID settings. This indicates its data augmentation nature that benefits the model's generalization ability by making overall progress in learning. Also, the deviation minimization of feature covariant matrices benefits Deep Coral's performances in ID settings.

\textbf{Insights on future OOD method development.} Our results and comparisons show that current OOD algorithms can improve generalization abilities, but not significantly, underscoring the need for OOD methods that are more robust and better-performing in practice. 
Additionally, in practice models cannot be expected to solve unknown distribution shifts. Thus, we believe using the given environment information in training to convey the types of shifts expected during testing is a promising direction. Similarly, we suggest using OOD validation containing possible distribution shift types of OOD test set to select models that are potentially better at our target generalization abilities.
Moreover, we observe distinct performance difference on covariate and concept shifts for OOD algorithms, demonstrating that OOD algorithms might need shift-specific design to maximize generalization ability for one type of shift. In this case, future OOD methods can focus on solving one of covariate or concept shift. Inspired by a recent work~\cite{zhang2022nico++}, we expect to evaluate covariate and concept shifts using shift-specific metrics. Therefore, covariate and concept shifts can be viewed, solved and evaluated separately.
On top of that, OOD generalization abilities can be improved by managing well-designed model architectures, optimization schemes, or data augmentation strategies. To solve graph OOD problems, it is critical that methods should be specifically designed for graphs. For example, Mixup's specifically designed node prediction network~\cite{wang2021mixup} is quite well-performing while the graph prediction network~\cite{wang2021mixup} adopted directly from image field~\cite{zhang2017mixup} shows no advantage. One possible reason is that functional data augmentation for graphs should consider the complex structure of graphs, so simple strategies like direct graph representation Mixup can cause topological mismatch.

\section{Complete Dataset Results}\label{sec:D}

\subsection{Complete numerical results}
We report the complete results of ID/OOD test performances from ID/OOD validation for 10 baselines on 11 datasets in a series of tables, as shown in Table~\ref{table:res1}-\ref{table:res16}.

\begin{table*}[!htbp]
% \setlength{\abovecaptionskip}{0.1cm}
% \footnotesize
% \scalebox{0.622}{
% % \setlength{\tabcolsep}{0.3mm}{
% \centering
\centering
\resizebox{\textwidth}{!}
{
\begin{tabular}{lllccccccccc}
\toprule[2pt]
\multicolumn{3}{c}{\multirow{2}{*}{GOOD-HIV}} & \multicolumn{9}{c}{scaffold}\\ 
\cmidrule(r){4-12}
& & & \multicolumn{4}{c|}{covariate} & \multicolumn{4}{c|}{concept} & \multicolumn{1}{c}{no shift} \\
\cmidrule(r){1-12}
\multicolumn{3}{c}{\multirow{2}{*}{ROC-AUC}} & \multicolumn{2}{c|}{ID validation} & \multicolumn{2}{c|}{OOD validation} & \multicolumn{2}{c|}{ID validation} & \multicolumn{2}{c|}{OOD validation} & \multicolumn{1}{c}{ID validation} \\
\cmidrule(r){4-12}
& & & \multicolumn{1}{c|}{ID test} & \multicolumn{1}{c|}{OOD test} & \multicolumn{1}{c|}{ID test} & \multicolumn{1}{c|}{OOD test} & \multicolumn{1}{c|}{ID test} & \multicolumn{1}{c|}{OOD test} & \multicolumn{1}{c|}{ID test} & \multicolumn{1}{c|}{OOD test} & ID test \\
\midrule[1pt]

\multicolumn{3}{l}{ERM} & \textbf{82.79}\textpm1.10 & 68.86\textpm2.10 & 80.84\textpm0.57 & 69.58\textpm1.99 & 84.22\textpm0.85 & 65.31\textpm3.49 & 82.64\textpm1.58 & 72.33\textpm1.04 & 80.86\textpm0.63 \\
\multicolumn{3}{l}{IRM} & 81.35\textpm0.83 & 67.31\textpm1.94 & 80.74\textpm0.87 & 67.97\textpm2.46 & 82.89\textpm1.27 & 66.06\textpm3.06 & 81.93\textpm1.11 & 72.59\textpm0.45 & \textbf{81.06}\textpm0.61 \\
\multicolumn{3}{l}{VREx} & 82.11\textpm1.48 & 69.25\textpm1.84 & 81.09\textpm1.56 & \textbf{70.77}\textpm1.35 & 83.84\textpm1.09 & 66.48\textpm2.16 & 82.55\textpm1.09 & 72.60\textpm0.82 & 80.57\textpm0.65 \\
\multicolumn{3}{l}{GroupDRO} & 82.60\textpm1.25 & 69.24\textpm2.20 & 81.60\textpm1.40 & 70.64\textpm1.72 & 83.40\textpm0.67 & 65.89\textpm2.78 & 82.01\textpm1.28 & \textbf{73.64}\textpm0.86 & 80.27\textpm0.90 \\
\multicolumn{3}{l}{DANN} & 81.18\textpm1.37 & 70.05\textpm1.02 & 80.85\textpm1.42 & 70.63\textpm1.82 & 83.87\textpm0.99 & \textbf{66.57}\textpm2.30 & 82.58\textpm1.14 & 71.92\textpm1.23 & 80.82\textpm0.64 \\
\multicolumn{3}{l}{Deep Coral} & 82.53\textpm1.01 & 68.00\textpm2.62 & \textbf{82.02}\textpm0.69 & 68.61\textpm1.70 & \textbf{84.65}\textpm1.73 & 65.74\textpm3.49 & \textbf{82.99}\textpm2.09 & 72.97\textpm1.04 & 80.73\textpm0.83 \\
\cmidrule(r){1-12}
\multicolumn{3}{l}{Mixup} & 82.29\textpm1.34 & \textbf{70.66}\textpm3.56 & 81.27\textpm1.83 & 68.88\textpm2.40 & 82.36\textpm1.94 & 65.94\textpm2.96 & 80.81\textpm2.26 & 72.03\textpm0.53 & 80.28\textpm1.27 \\
\multicolumn{3}{l}{DIR} & {82.54}\textpm0.17 & {66.71}\textpm2.38 & {76.75}\textpm1.52 & {67.47}\textpm2.61 & {83.28}\textpm0.48 & {65.13}\textpm2.46 & {81.71}\textpm1.30 & {69.05}\textpm0.92 & {79.40}\textpm0.60 \\

\bottomrule[2pt]
\end{tabular}
}
% }
\caption{ID/OOD test performances from ID/OOD validation on GOOD-HIV with scaffold domain. 
Numerical results are average \textpm { }standard deviation across 10 random runs. Numbers in \textbf{bold} represent the best results. The metric and domain selections for each dataset are listed in each table. Note that the no shift random split only has the ID setting.}
\label{table:res1}
\vspace{-5pt}
\end{table*}

\begin{table*}[!htbp]
% \setlength{\abovecaptionskip}{0.1cm}
% \footnotesize
% \scalebox{0.622}{
% % \setlength{\tabcolsep}{0.3mm}{
% \centering
\centering
\resizebox{\textwidth}{!}
{
\begin{tabular}{lllccccccccc}
\toprule[2pt]
\multicolumn{3}{c}{\multirow{2}{*}{GOOD-HIV}} & \multicolumn{9}{c}{size}\\ 
\cmidrule(r){4-12}
& & & \multicolumn{4}{c|}{covariate} & \multicolumn{4}{c|}{concept} & \multicolumn{1}{c}{no shift} \\
\cmidrule(r){1-12}
\multicolumn{3}{c}{\multirow{2}{*}{ROC-AUC}} & \multicolumn{2}{c|}{ID validation} & \multicolumn{2}{c|}{OOD validation} & \multicolumn{2}{c|}{ID validation} & \multicolumn{2}{c|}{OOD validation} & \multicolumn{1}{c}{ID validation} \\
\cmidrule(r){4-12}
& & & \multicolumn{1}{c|}{ID test} & \multicolumn{1}{c|}{OOD test} & \multicolumn{1}{c|}{ID test} & \multicolumn{1}{c|}{OOD test} & \multicolumn{1}{c|}{ID test} & \multicolumn{1}{c|}{OOD test} & \multicolumn{1}{c|}{ID test} & \multicolumn{1}{c|}{OOD test} & ID test \\
\midrule[1pt]

\multicolumn{3}{l}{ERM} & 83.72\textpm1.06 & 58.41\textpm2.53 & 82.94\textpm1.65 & 59.94\textpm2.86 & 88.05\textpm0.67 & 44.75\textpm2.92 & 82.97\textpm2.73 & 63.26\textpm2.47 & 80.86\textpm0.63 \\
\multicolumn{3}{l}{IRM} & 81.33\textpm1.13 & 58.41\textpm1.79 & 79.93\textpm1.00 & 59.00\textpm2.74 & \textbf{88.62}\textpm0.86 & 44.17\textpm4.58 & \textbf{85.67}\textpm1.20 & 59.90\textpm3.15 & \textbf{81.06}\textpm0.61 \\
\multicolumn{3}{l}{VREx} & 83.47\textpm1.11 & \textbf{60.24}\textpm2.54 & 83.20\textpm1.35 & 58.53\textpm2.22 & 88.28\textpm0.88 & 44.43\textpm3.77 & 84.93\textpm1.32 & 60.23\textpm1.70 & 80.57\textpm0.65 \\
\multicolumn{3}{l}{GroupDRO} & 83.79\textpm0.68 & 59.50\textpm2.21 & 82.03\textpm1.45 & 58.98\textpm1.84 & 88.28\textpm0.84 & 45.42\textpm3.34 & 84.41\textpm1.72 & 61.37\textpm2.79 & 80.27\textpm0.90 \\
\multicolumn{3}{l}{DANN} & 83.90\textpm0.68 & 58.68\textpm3.02 & 82.17\textpm2.49 & 58.68\textpm1.83 & 87.28\textpm1.12 & 43.26\textpm3.68 & 81.83\textpm2.56 & {65.27}\textpm3.75 & 80.82\textpm0.64 \\
\multicolumn{3}{l}{Deep Coral} & \textbf{84.70}\textpm1.17 & 59.72\textpm3.66 & \textbf{83.89}\textpm0.83 & \textbf{60.11}\textpm3.53 & 87.88\textpm0.57 & {47.56}\textpm3.55 & 84.80\textpm1.17 & 62.28\textpm1.42 & 80.73\textpm0.83 \\
\cmidrule(r){1-12}
\multicolumn{3}{l}{Mixup} & 83.16\textpm1.12 & 60.13\textpm2.06 & 82.03\textpm1.72 & 59.03\textpm3.07 & 87.64\textpm0.81 & 46.19\textpm4.40 & 81.20\textpm1.97 & 64.87\textpm1.77 & 80.28\textpm1.27 \\
\multicolumn{3}{l}{DIR} & {80.46}\textpm0.55 & {56.88}\textpm1.54 & {79.98}\textpm1.36 & {57.11}\textpm1.43 & {79.19}\textpm0.76 & \textbf{68.33}\textpm2.02 & {78.41}\textpm3.08 & \textbf{72.61}\textpm2.03 & {79.40}\textpm0.60 \\

\bottomrule[2pt]
\end{tabular}
}
% }
\caption{Performance on GOOD-HIV with size domain.}
\label{table:res2}
\vspace{-5pt}
\end{table*}

\begin{table*}[!htbp]
% \setlength{\abovecaptionskip}{0.1cm}
% \footnotesize
% \scalebox{0.622}{
% % \setlength{\tabcolsep}{0.3mm}{
% \centering
\centering
\resizebox{\textwidth}{!}
{
\begin{tabular}{lllccccccccc}
\toprule[2pt]
\multicolumn{3}{c}{\multirow{2}{*}{GOOD-PCBA}} & \multicolumn{9}{c}{scaffold}\\ 
\cmidrule(r){4-12}
& & & \multicolumn{4}{c|}{covariate} & \multicolumn{4}{c|}{concept} & \multicolumn{1}{c}{no shift} \\
\cmidrule(r){1-12}
\multicolumn{3}{c}{\multirow{2}{*}{AP}} & \multicolumn{2}{c|}{ID validation} & \multicolumn{2}{c|}{OOD validation} & \multicolumn{2}{c|}{ID validation} & \multicolumn{2}{c|}{OOD validation} & \multicolumn{1}{c}{ID validation} \\
\cmidrule(r){4-12}
& & & \multicolumn{1}{c|}{ID test} & \multicolumn{1}{c|}{OOD test} & \multicolumn{1}{c|}{ID test} & \multicolumn{1}{c|}{OOD test} & \multicolumn{1}{c|}{ID test} & \multicolumn{1}{c|}{OOD test} & \multicolumn{1}{c|}{ID test} & \multicolumn{1}{c|}{OOD test} & ID test \\
\midrule[1pt]

\multicolumn{3}{l}{ERM} & 33.45\textpm0.42 & 16.87\textpm0.49 & 32.62\textpm1.02 & 16.89\textpm0.55 & 25.95\textpm0.94 & 21.34\textpm0.89 & 25.95\textpm1.06 & 21.63\textpm0.97 & 33.77\textpm0.31 \\
\multicolumn{3}{l}{IRM} & 33.56\textpm0.57 & 16.94\textpm0.35 & 32.86\textpm0.65 & 16.90\textpm0.42 & 25.89\textpm0.42 & 21.05\textpm0.39 & 25.78\textpm0.62 & 21.22\textpm0.39 & 33.36\textpm0.31 \\
\multicolumn{3}{l}{VREx} & \textbf{33.88}\textpm0.74 & 17.01\textpm0.27 & \textbf{33.27}\textpm1.18 & \textbf{16.98}\textpm0.29 & \textbf{26.62}\textpm0.64 & {21.98}\textpm0.86 & 26.45\textpm0.73 & {22.02}\textpm0.88 & 33.61\textpm0.49 \\
\multicolumn{3}{l}{GroupDRO} & 33.81\textpm0.55 & \textbf{17.06}\textpm0.28 & 32.32\textpm0.88 & 16.98\textpm0.26 & 26.32\textpm0.41 & 21.61\textpm0.53 & 26.03\textpm0.75 & 21.83\textpm0.61 & 33.35\textpm0.53 \\
\multicolumn{3}{l}{DANN} & 33.63\textpm0.46 & 16.86\textpm0.46 & 32.62\textpm0.90 & 16.90\textpm0.33 & 26.07\textpm0.29 & 21.23\textpm0.44 & 25.99\textpm0.46 & 21.64\textpm0.37 & 33.47\textpm0.32 \\
\multicolumn{3}{l}{Deep Coral} & 33.47\textpm0.57 & 16.84\textpm0.55 & 32.50\textpm1.49 & 16.93\textpm0.59 & 26.38\textpm0.82 & 21.70\textpm0.66 & \textbf{26.46}\textpm0.83 & 21.95\textpm0.76 & \textbf{33.77}\textpm0.48 \\
\cmidrule(r){1-12}
\multicolumn{3}{l}{Mixup} & 30.22\textpm0.33 & 16.68\textpm0.37 & 29.92\textpm0.46 & 16.59\textpm0.42 & 23.73\textpm0.53 & 19.58\textpm0.56 & 23.25\textpm0.79 & 19.78\textpm0.44 & 30.35\textpm0.26 \\
\multicolumn{3}{l}{DIR} & {32.55}\textpm0.17 & {14.97}\textpm0.35 & {30.58}\textpm0.34 & {14.98}\textpm0.32 & {25.85}\textpm0.37 & \textbf{22.26}\textpm0.50 & {25.55}\textpm0.25 & \textbf{22.20}\textpm0.43 & {30.50}\textpm0.69 \\

\bottomrule[2pt]
\end{tabular}
}
% }
\caption{Performance on GOOD-PCBA with scaffold domain.}
\label{table:res3}
\vspace{-5pt}
\end{table*}

\begin{table*}[!htbp]
% \setlength{\abovecaptionskip}{0.1cm}
% \footnotesize
% \scalebox{0.622}{
% % \setlength{\tabcolsep}{0.3mm}{
% \centering
\centering
\resizebox{\textwidth}{!}
{
\begin{tabular}{lllccccccccc}
\toprule[2pt]
\multicolumn{3}{c}{\multirow{2}{*}{GOOD-PCBA}} & \multicolumn{9}{c}{size}\\ 
\cmidrule(r){4-12}
& & & \multicolumn{4}{c|}{covariate} & \multicolumn{4}{c|}{concept} & \multicolumn{1}{c}{no shift} \\
\cmidrule(r){1-12}
\multicolumn{3}{c}{\multirow{2}{*}{AP}} & \multicolumn{2}{c|}{ID validation} & \multicolumn{2}{c|}{OOD validation} & \multicolumn{2}{c|}{ID validation} & \multicolumn{2}{c|}{OOD validation} & \multicolumn{1}{c}{ID validation} \\
\cmidrule(r){4-12}
& & & \multicolumn{1}{c|}{ID test} & \multicolumn{1}{c|}{OOD test} & \multicolumn{1}{c|}{ID test} & \multicolumn{1}{c|}{OOD test} & \multicolumn{1}{c|}{ID test} & \multicolumn{1}{c|}{OOD test} & \multicolumn{1}{c|}{ID test} & \multicolumn{1}{c|}{OOD test} & ID test \\
\midrule[1pt]

\multicolumn{3}{l}{ERM} & 34.31\textpm0.57 & 17.81\textpm0.43 & 34.29\textpm0.56 & 17.86\textpm0.38 & 32.54\textpm0.83 & 14.83\textpm0.61 & 31.96\textpm0.93 & 15.36\textpm0.54 & 33.77\textpm0.31 \\
\multicolumn{3}{l}{IRM} & 34.28\textpm0.46 & \textbf{17.94}\textpm0.30 & 34.29\textpm0.54 & \textbf{18.05}\textpm0.29 & 32.99\textpm0.89 & {15.76}\textpm0.54 & 32.55\textpm0.89 & 16.07\textpm0.52 & 33.36\textpm0.31 \\
\multicolumn{3}{l}{VREx} & 34.09\textpm0.29 & 17.76\textpm0.43 & 34.07\textpm0.28 & 17.79\textpm0.41 & 32.49\textpm0.76 & 15.22\textpm0.53 & 32.06\textpm0.74 & 15.59\textpm0.57 & 33.61\textpm0.49 \\
\multicolumn{3}{l}{GroupDRO} & 33.95\textpm0.51 & 17.49\textpm0.46 & 33.92\textpm0.45 & 17.59\textpm0.46 & \textbf{33.03}\textpm0.32 & 15.62\textpm0.53 & \textbf{32.58}\textpm0.45 & 15.99\textpm0.43 & 33.35\textpm0.53 \\
\multicolumn{3}{l}{DANN} & 34.17\textpm0.34 & 17.86\textpm0.47 & 34.09\textpm0.34 & 17.86\textpm0.48 & 32.74\textpm0.50 & 15.40\textpm0.46 & 32.25\textpm0.77 & 15.78\textpm0.39 & 33.47\textpm0.32 \\
\multicolumn{3}{l}{Deep Coral} & \textbf{34.49}\textpm0.43 & 17.76\textpm0.39 & \textbf{34.41}\textpm0.43 & 17.94\textpm0.38 & 32.67\textpm1.01 & 15.63\textpm0.77 & 32.14\textpm1.21 & {16.20}\textpm0.72 & \textbf{33.77}\textpm0.48 \\
\cmidrule(r){1-12}
\multicolumn{3}{l}{Mixup} & 30.63\textpm0.65 & 17.09\textpm0.58 & 30.55\textpm0.72 & 17.06\textpm0.54 & 30.23\textpm1.02 & 13.00\textpm0.81 & 29.97\textpm1.13 & 13.36\textpm0.66 & 30.35\textpm0.26 \\
\multicolumn{3}{l}{DIR} & {32.89}\textpm0.20 & {16.39}\textpm0.28 & {32.62}\textpm0.04 & {16.61}\textpm0.17 & {30.53}\textpm0.28 & \textbf{16.60}\textpm0.43 & {30.32}\textpm0.21 & \textbf{16.86}\textpm0.26 & {30.50}\textpm0.69 \\ 

\bottomrule[2pt]
\end{tabular}
}
% }
\caption{Performance on GOOD-PCBA with size domain.}
\label{table:res4}
\end{table*}

\begin{table*}[!htbp]
% \setlength{\abovecaptionskip}{0.1cm}
% \footnotesize
% \scalebox{0.622}{
% % \setlength{\tabcolsep}{0.3mm}{
% \centering
\centering
\resizebox{\textwidth}{!}
{
\begin{tabular}{lllccccccccc}
\toprule[2pt]
\multicolumn{3}{c}{\multirow{2}{*}{GOOD-ZINC}} & \multicolumn{9}{c}{scaffold}\\ 
\cmidrule(r){4-12}
& & & \multicolumn{4}{c|}{covariate} & \multicolumn{4}{c|}{concept} & \multicolumn{1}{c}{no shift} \\
\cmidrule(r){1-12}
\multicolumn{3}{c}{\multirow{2}{*}{MAE}} & \multicolumn{2}{c|}{ID validation} & \multicolumn{2}{c|}{OOD validation} & \multicolumn{2}{c|}{ID validation} & \multicolumn{2}{c|}{OOD validation} & \multicolumn{1}{c}{ID validation} \\
\cmidrule(r){4-12}
& & & \multicolumn{1}{c|}{ID test} & \multicolumn{1}{c|}{OOD test} & \multicolumn{1}{c|}{ID test} & \multicolumn{1}{c|}{OOD test} & \multicolumn{1}{c|}{ID test} & \multicolumn{1}{c|}{OOD test} & \multicolumn{1}{c|}{ID test} & \multicolumn{1}{c|}{OOD test} & ID test \\
\midrule[1pt]

\multicolumn{3}{l}{ERM} & 0.1224\textpm0.0029 & 0.1825\textpm0.0129 & 0.1384\textpm0.0075 & 0.1995\textpm0.0114 & 0.1222\textpm0.0052 & 0.1328\textpm0.0060 & 0.1225\textpm0.0055 & 0.1306\textpm0.0038 & 0.1233\textpm0.0045 \\
\multicolumn{3}{l}{IRM} & 0.1213\textpm0.0044 & 0.1787\textpm0.0094 & 0.1463\textpm0.0128 & 0.2025\textpm0.0145 & 0.1225\textpm0.0036 & 0.1319\textpm0.0039 & 0.1223\textpm0.0035 & 0.1314\textpm0.0042 & 0.1200\textpm0.0049 \\
\multicolumn{3}{l}{VREx} & 0.1211\textpm0.0025 & 0.1771\textpm0.0099 & 0.1512\textpm0.0130 & 0.2094\textpm0.0118 & 0.1186\textpm0.0035 & 0.1273\textpm0.0044 & 0.1186\textpm0.0036 & 0.1270\textpm0.0040 & 0.1247\textpm0.0021 \\
\multicolumn{3}{l}{GroupDRO} & \textbf{0.1168}\textpm0.0045 & 0.1784\textpm0.0083 & \textbf{0.1373}\textpm0.0079 & \textbf{0.1934}\textpm0.0114 & 0.1207\textpm0.0037 & 0.1284\textpm0.0042 & 0.1210\textpm0.0038 & 0.1281\textpm0.0041 & 0.1222\textpm0.0059 \\
\multicolumn{3}{l}{DANN} & 0.1186\textpm0.0030 & 0.1762\textpm0.0108 & 0.1404\textpm0.0133 & 0.2004\textpm0.0113 & \textbf{0.1172}\textpm0.0044 & \textbf{0.1262}\textpm0.0051 & \textbf{0.1171}\textpm0.0040 & \textbf{0.1256}\textpm0.0048 & 0.1217\textpm0.0057 \\
\multicolumn{3}{l}{Deep Coral} & 0.1185\textpm0.0045 & \textbf{0.1752}\textpm0.0080 & 0.1438\textpm0.0097 & 0.2036\textpm0.0158 & 0.1187\textpm0.0066 & 0.1287\textpm0.0077 & 0.1191\textpm0.0070 & 0.1279\textpm0.0067 & \textbf{0.1156}\textpm0.0055 \\
\cmidrule(r){1-12}
\multicolumn{3}{l}{Mixup} & 0.1279\textpm0.0056 & 0.1951\textpm0.0124 & 0.1575\textpm0.0191 & 0.2240\textpm0.0258 & 0.1353\textpm0.0068 & 0.1479\textpm0.0056 & 0.1357\textpm0.0067 & 0.1475\textpm0.0059 & 0.1418\textpm0.0064 \\
\multicolumn{3}{l}{DIR} & 0.3799\textpm0.0321 & 0.6155\textpm0.0589 & 0.3980\textpm0.0401 & 0.6493\textpm0.0717 & 0.3501\textpm0.1102 & 0.3883\textpm0.1019 & 0.3523\textpm0.1074 & 0.3865\textpm0.1040 & 0.6623\textpm0.3615 \\

\bottomrule[2pt]
\end{tabular}
}
% }
\caption{Performance on GOOD-ZINC with scaffold domain.}
\label{table:res5}
\end{table*}

\begin{table*}[!htbp]
% \setlength{\abovecaptionskip}{0.1cm}
% \footnotesize
% \scalebox{0.622}{
% % \setlength{\tabcolsep}{0.3mm}{
% \centering
\centering
\resizebox{\textwidth}{!}
{
\begin{tabular}{lllccccccccc}
\toprule[2pt]
\multicolumn{3}{c}{\multirow{2}{*}{GOOD-ZINC}} & \multicolumn{9}{c}{size}\\ 
\cmidrule(r){4-12}
& & & \multicolumn{4}{c|}{covariate} & \multicolumn{4}{c|}{concept} & \multicolumn{1}{c}{no shift} \\
\cmidrule(r){1-12}
\multicolumn{3}{c}{\multirow{2}{*}{MAE}} & \multicolumn{2}{c|}{ID validation} & \multicolumn{2}{c|}{OOD validation} & \multicolumn{2}{c|}{ID validation} & \multicolumn{2}{c|}{OOD validation} & \multicolumn{1}{c}{ID validation} \\
\cmidrule(r){4-12}
& & & \multicolumn{1}{c|}{ID test} & \multicolumn{1}{c|}{OOD test} & \multicolumn{1}{c|}{ID test} & \multicolumn{1}{c|}{OOD test} & \multicolumn{1}{c|}{ID test} & \multicolumn{1}{c|}{OOD test} & \multicolumn{1}{c|}{ID test} & \multicolumn{1}{c|}{OOD test} & ID test \\
\midrule[1pt]

\multicolumn{3}{l}{ERM} & 0.1199\textpm0.0060 & 0.2569\textpm0.0138 & 0.1323\textpm0.0092 & 0.2427\textpm0.0068 & 0.1315\textpm0.0073 & 0.1418\textpm0.0057 & 0.1346\textpm0.0079 & 0.1403\textpm0.0065 & 0.1233\textpm0.0045 \\
\multicolumn{3}{l}{IRM} & 0.1222\textpm0.0059 & \textbf{0.2536}\textpm0.0227 & 0.1317\textpm0.0100 & 0.2403\textpm0.0106 & 0.1278\textpm0.0077 & 0.1403\textpm0.0138 & 0.1302\textpm0.0084 & 0.1368\textpm0.0119 & 0.1200\textpm0.0049 \\
\multicolumn{3}{l}{VREx} & 0.1234\textpm0.0054 & 0.2560\textpm0.0212 & 0.1327\textpm0.0089 & \textbf{0.2384}\textpm0.0098 & 0.1309\textpm0.0064 & 0.1462\textpm0.0139 & 0.1352\textpm0.0092 & 0.1419\textpm0.0090 & 0.1247\textpm0.0021 \\
\multicolumn{3}{l}{GroupDRO} & 0.1180\textpm0.0054 & 0.2598\textpm0.0213 & 0.1293\textpm0.0069 & 0.2423\textpm0.0097 & \textbf{0.1251}\textpm0.0066 & 0.1402\textpm0.0091 & \textbf{0.1273}\textpm0.0089 & 0.1369\textpm0.0076 & 0.1222\textpm0.0059 \\
\multicolumn{3}{l}{DANN} & 0.1188\textpm0.0048 & 0.2555\textpm0.0183 & 0.1303\textpm0.0057 & 0.2439\textpm0.0056 & 0.1253\textpm0.0034 & \textbf{0.1371}\textpm0.0084 & 0.1297\textpm0.0055 & \textbf{0.1339}\textpm0.0048 & 0.1217\textpm0.0057 \\
\multicolumn{3}{l}{Deep Coral} & \textbf{0.1134}\textpm0.0071 & 0.2545\textpm0.0159 & \textbf{0.1269}\textpm0.0092 & 0.2505\textpm0.0073 & 0.1287\textpm0.0041 & 0.1415\textpm0.0074 & 0.1310\textpm0.0058 & 0.1370\textpm0.0052 & \textbf{0.1156}\textpm0.0055 \\
\cmidrule(r){1-12}
\multicolumn{3}{l}{Mixup} & 0.1255\textpm0.0071 & 0.2776\textpm0.0215 & 0.1317\textpm0.0145 & 0.2748\textpm0.0167 & 0.1423\textpm0.0062 & 0.1599\textpm0.0115 & 0.1459\textpm0.0073 & 0.1522\textpm0.0064 & 0.1418\textpm0.0064 \\
\multicolumn{3}{l}{DIR} & 0.1541\textpm0.0036 & 0.6011\textpm0.0147 & 0.1718\textpm0.0097 & 0.5482\textpm0.0279 & 0.2348\textpm0.0455 & 0.3130\textpm0.0747 & 0.2485\textpm0.0361 & 0.2871\textpm0.0958 & 0.6623\textpm0.3615 \\

\bottomrule[2pt]
\end{tabular}
}
% }
\caption{Performance on GOOD-ZINC with size domain.}
\label{table:res6}
\end{table*}

\begin{table*}[!htbp]
% \setlength{\abovecaptionskip}{0.1cm}
% \footnotesize
% \scalebox{0.622}{
% % \setlength{\tabcolsep}{0.3mm}{
% \centering
\centering
\resizebox{\textwidth}{!}
{
\begin{tabular}{lllccccccccc}
\toprule[2pt]
\multicolumn{3}{c}{\multirow{2}{*}{GOOD-SST2}} & \multicolumn{9}{c}{length}\\ 
\cmidrule(r){4-12}
& & & \multicolumn{4}{c|}{covariate} & \multicolumn{4}{c|}{concept} & \multicolumn{1}{c}{no shift} \\
\cmidrule(r){1-12}
\multicolumn{3}{c}{\multirow{2}{*}{Accuracy}} & \multicolumn{2}{c|}{ID validation} & \multicolumn{2}{c|}{OOD validation} & \multicolumn{2}{c|}{ID validation} & \multicolumn{2}{c|}{OOD validation} & \multicolumn{1}{c}{ID validation} \\
\cmidrule(r){4-12}
& & & \multicolumn{1}{c|}{ID test} & \multicolumn{1}{c|}{OOD test} & \multicolumn{1}{c|}{ID test} & \multicolumn{1}{c|}{OOD test} & \multicolumn{1}{c|}{ID test} & \multicolumn{1}{c|}{OOD test} & \multicolumn{1}{c|}{ID test} & \multicolumn{1}{c|}{OOD test} & ID test \\
\midrule[1pt]

\multicolumn{3}{l}{ERM} & \textbf{89.82}\textpm0.01 & 77.76\textpm1.14 & 89.26\textpm0.22 & 81.30\textpm0.35 & \textbf{94.43}\textpm0.05 & 67.26\textpm0.05 & \textbf{93.82}\textpm0.09 & 72.43\textpm0.48 & 91.61\textpm0.02 \\
\multicolumn{3}{l}{IRM} & 89.41\textpm0.11 & 78.22\textpm2.20 & 88.83\textpm0.35 & 79.91\textpm1.97 & 94.10\textpm0.06 & 66.64\textpm0.16 & 82.91\textpm3.89 & \textbf{77.47}\textpm0.71 & 91.43\textpm0.05 \\
\multicolumn{3}{l}{VREx} & 89.51\textpm0.03 & 79.60\textpm1.05 & 89.57\textpm0.09 & 80.64\textpm0.35 & 94.26\textpm0.02 & \textbf{69.14}\textpm0.86 & 92.93\textpm0.26 & 73.16\textpm0.47 & 91.61\textpm0.18 \\
\multicolumn{3}{l}{GroupDRO} & 89.59\textpm0.09 & 79.21\textpm1.02 & \textbf{89.66}\textpm0.04 & \textbf{81.35}\textpm0.54 & 94.41\textpm0.07 & 67.30\textpm0.41 & 93.00\textpm0.49 & 71.86\textpm0.23 & 91.66\textpm0.19 \\
\multicolumn{3}{l}{DANN} & 89.60\textpm0.19 & 76.15\textpm1.34 & 89.50\textpm0.13 & 79.71\textpm1.35 & 94.02\textpm0.10 & 66.55\textpm1.08 & 90.47\textpm1.14 & 76.03\textpm1.49 & 91.67\textpm0.04 \\
\multicolumn{3}{l}{Deep Coral} & 89.68\textpm0.06 & 78.99\textpm0.43 & 88.99\textpm0.36 & 79.81\textpm0.22 & 94.25\textpm0.18 & 67.84\textpm0.78 & 93.42\textpm0.38 & 72.34\textpm0.51 & \textbf{91.89}\textpm0.15 \\
\cmidrule(r){1-12}
\multicolumn{3}{l}{Mixup} & 89.78\textpm0.20 & \textbf{80.22}\textpm0.60 & 89.62\textpm0.09 & 80.88\textpm0.60 & 94.12\textpm0.10 & 67.31\textpm0.74 & 93.18\textpm0.10 & 73.34\textpm0.40 & 91.69\textpm0.04 \\
\multicolumn{3}{l}{DIR} & 84.30\textpm0.46 & 74.76\textpm2.31 & 82.73\textpm0.76 & 77.65\textpm1.93 & 93.71\textpm0.18 & 63.61\textpm1.32 & 91.03\textpm1.55 & 68.76\textpm1.04 & 89.11\textpm0.11 \\

\bottomrule[2pt]
\end{tabular}
}
% }
\caption{Performance on GOOD-SST2 with length domain.}
\label{table:res65}
\end{table*}

\begin{table*}[!htbp]
% \setlength{\abovecaptionskip}{0.1cm}
% \footnotesize
% \scalebox{0.622}{
% % \setlength{\tabcolsep}{0.3mm}{
% \centering
\centering
\resizebox{\textwidth}{!}
{
\begin{tabular}{lllccccccccc}
\toprule[2pt]
\multicolumn{3}{c}{\multirow{2}{*}{GOOD-CMNIST}} & \multicolumn{9}{c}{color}\\ 
\cmidrule(r){4-12}
& & & \multicolumn{4}{c|}{covariate} & \multicolumn{4}{c|}{concept} & \multicolumn{1}{c}{no shift} \\
\cmidrule(r){1-12}
\multicolumn{3}{c}{\multirow{2}{*}{Accuracy}} & \multicolumn{2}{c|}{ID validation} & \multicolumn{2}{c|}{OOD validation} & \multicolumn{2}{c|}{ID validation} & \multicolumn{2}{c|}{OOD validation} & \multicolumn{1}{c}{ID validation} \\
\cmidrule(r){4-12}
& & & \multicolumn{1}{c|}{ID test} & \multicolumn{1}{c|}{OOD test} & \multicolumn{1}{c|}{ID test} & \multicolumn{1}{c|}{OOD test} & \multicolumn{1}{c|}{ID test} & \multicolumn{1}{c|}{OOD test} & \multicolumn{1}{c|}{ID test} & \multicolumn{1}{c|}{OOD test} & ID test \\
\midrule[1pt]

\multicolumn{3}{l}{ERM} & 77.96\textpm0.34 & \textbf{26.90}\textpm1.91 & 76.26\textpm0.56 & 28.60\textpm2.01 & 90.00\textpm0.17 & 40.80\textpm1.60 & 89.43\textpm0.33 & 42.87\textpm0.72 & \textbf{77.30}\textpm0.35 \\
\multicolumn{3}{l}{IRM} & 77.92\textpm0.30 & 25.81\textpm2.70 & 75.91\textpm2.89 & 27.83\textpm1.84 & 90.02\textpm0.12 & 41.70\textpm0.54 & 89.44\textpm0.43 & 42.80\textpm0.38 & 77.28\textpm0.21 \\
\multicolumn{3}{l}{VREx} & 77.98\textpm0.32 & 26.75\textpm2.21 & 76.42\textpm0.74 & 28.48\textpm2.08 & 89.99\textpm0.18 & 41.26\textpm1.40 & 89.42\textpm0.24 & 43.31\textpm0.78 & 77.03\textpm0.44 \\
\multicolumn{3}{l}{GroupDRO} & 77.98\textpm0.38 & 26.51\textpm0.95 & \textbf{76.57}\textpm0.84 & 29.07\textpm2.62 & \textbf{90.02}\textpm0.27 & 41.47\textpm0.95 & 89.33\textpm0.32 & \textbf{43.32}\textpm0.75 & 77.01\textpm0.33 \\
\multicolumn{3}{l}{DANN} & 78.00\textpm0.43 & 26.82\textpm1.64 & 76.02\textpm1.77 & \textbf{29.14}\textpm2.93 & 89.94\textpm0.19 & \textbf{41.86}\textpm0.68 & 89.49\textpm0.39 & 43.11\textpm0.64 & 77.15\textpm0.48 \\
\multicolumn{3}{l}{Deep Coral} & \textbf{78.64}\textpm0.48 & 26.16\textpm1.59 & 76.11\textpm1.60 & 29.05\textpm2.19 & 89.94\textpm0.17 & 41.28\textpm0.86 & 89.42\textpm0.28 & 43.16\textpm0.56 & 77.12\textpm0.32 \\
\cmidrule(r){1-12}
\multicolumn{3}{l}{Mixup} & 77.40\textpm0.22 & 26.24\textpm2.43 & 74.86\textpm1.13 & 26.47\textpm1.73 & 89.95\textpm0.25 & 39.59\textpm1.11 & \textbf{89.63}\textpm0.31 & 40.96\textpm0.81 & 76.62\textpm0.37 \\
\multicolumn{3}{l}{DIR} & {31.09}\textpm5.92 & {16.96}\textpm3.30 & {24.76}\textpm7.30 & {20.60}\textpm4.26 & {86.76}\textpm0.30 & {12.39}\textpm3.44 & {77.90}\textpm2.98 & {22.69}\textpm2.85 & {29.55}\textpm2.67 \\

\bottomrule[2pt]
\end{tabular}
}
% }
\caption{Performance on GOOD-CMNIST with color domain.}
\label{table:res7}
\end{table*}

\begin{table*}[!htbp]
% \setlength{\abovecaptionskip}{0.1cm}
% \footnotesize
% \scalebox{0.622}{
% % \setlength{\tabcolsep}{0.3mm}{
% \centering
\centering
\resizebox{\textwidth}{!}
{
\begin{tabular}{lllccccccccc}
\toprule[2pt]
\multicolumn{3}{c}{\multirow{2}{*}{GOOD-Motif}} & \multicolumn{9}{c}{base}\\ 
\cmidrule(r){4-12}
& & & \multicolumn{4}{c|}{covariate} & \multicolumn{4}{c|}{concept} & \multicolumn{1}{c}{no shift} \\
\cmidrule(r){1-12}
\multicolumn{3}{c}{\multirow{2}{*}{Accuracy}} & \multicolumn{2}{c|}{ID validation} & \multicolumn{2}{c|}{OOD validation} & \multicolumn{2}{c|}{ID validation} & \multicolumn{2}{c|}{OOD validation} & \multicolumn{1}{c}{ID validation} \\
\cmidrule(r){4-12}
& & & \multicolumn{1}{c|}{ID test} & \multicolumn{1}{c|}{OOD test} & \multicolumn{1}{c|}{ID test} & \multicolumn{1}{c|}{OOD test} & \multicolumn{1}{c|}{ID test} & \multicolumn{1}{c|}{OOD test} & \multicolumn{1}{c|}{ID test} & \multicolumn{1}{c|}{OOD test} & ID test \\
\midrule[1pt]

\multicolumn{3}{l}{ERM} & 92.60\textpm0.03 & 69.97\textpm1.94 & 92.43\textpm0.20 & 68.66\textpm3.43 & 92.02\textpm0.05 & \textbf{80.87}\textpm0.65 & 92.05\textpm0.04 & 81.44\textpm0.45 & 92.09\textpm0.04 \\
\multicolumn{3}{l}{IRM} & 92.60\textpm0.02 & 70.30\textpm1.23 & 92.51\textpm0.08 & 70.65\textpm3.18 & 92.00\textpm0.02 & 80.41\textpm0.27 & 92.00\textpm0.03 & 80.71\textpm0.46 & 92.04\textpm0.06 \\
\multicolumn{3}{l}{VREx} & 92.60\textpm0.03 & \textbf{72.23}\textpm2.28 & \textbf{92.52}\textpm0.12 & \textbf{71.47}\textpm2.75 & \textbf{92.05}\textpm0.06 & 80.71\textpm0.79 & \textbf{92.06}\textpm0.04 & \textbf{81.56}\textpm0.35 & 92.09\textpm0.07 \\
\multicolumn{3}{l}{GroupDRO} & 92.61\textpm0.03 & 70.29\textpm2.02 & 92.48\textpm0.13 & 68.24\textpm1.94 & 92.01\textpm0.04 & 80.32\textpm0.57 & 92.02\textpm0.05 & 81.43\textpm0.70 & 92.09\textpm0.08 \\
\multicolumn{3}{l}{DANN} & 92.60\textpm0.03 & 69.04\textpm1.90 & 92.38\textpm0.16 & 65.47\textpm5.35 & 92.02\textpm0.04 & 80.57\textpm0.59 & 92.04\textpm0.03 & 81.33\textpm0.52 & \textbf{92.10}\textpm0.06 \\
\multicolumn{3}{l}{Deep Coral} & 92.61\textpm0.03 & 70.43\textpm1.44 & 92.37\textpm0.27 & 68.88\textpm3.61 & 92.01\textpm0.05 & 80.27\textpm0.72 & 92.04\textpm0.03 & 81.37\textpm0.42 & 92.09\textpm0.07 \\
\cmidrule(r){1-12}
\multicolumn{3}{l}{Mixup} & \textbf{92.68}\textpm0.05 & 69.30\textpm1.00 & 92.48\textpm0.17 & 70.08\textpm2.06 & 91.89\textpm0.03 & 77.57\textpm0.56 & 91.89\textpm0.01 & 77.63\textpm0.57 & 92.04\textpm0.06 \\
\multicolumn{3}{l}{DIR} & {87.73}\textpm2.60 & {59.08}\textpm14.23 & {68.53}\textpm8.43 & {61.50}\textpm15.69 & {91.60}\textpm0.09 & {67.57}\textpm2.71 & {91.16}\textpm0.42 & {72.14}\textpm7.29 & {73.46}\textpm0.85 \\

\bottomrule[2pt]
\end{tabular}
}
% }
\caption{Performance on GOOD-Motif with base domain.}
\label{table:res8}
\end{table*}

\begin{table*}[!htbp]
% \setlength{\abovecaptionskip}{0.1cm}
% \footnotesize
% \scalebox{0.622}{
% % \setlength{\tabcolsep}{0.3mm}{
% \centering
\centering
\resizebox{\textwidth}{!}
{
\begin{tabular}{lllccccccccc}
\toprule[2pt]
\multicolumn{3}{c}{\multirow{2}{*}{GOOD-Motif}} & \multicolumn{9}{c}{size}\\ 
\cmidrule(r){4-12}
& & & \multicolumn{4}{c|}{covariate} & \multicolumn{4}{c|}{concept} & \multicolumn{1}{c}{no shift} \\
\cmidrule(r){1-12}
\multicolumn{3}{c}{\multirow{2}{*}{Accuracy}} & \multicolumn{2}{c|}{ID validation} & \multicolumn{2}{c|}{OOD validation} & \multicolumn{2}{c|}{ID validation} & \multicolumn{2}{c|}{OOD validation} & \multicolumn{1}{c}{ID validation} \\
\cmidrule(r){4-12}
& & & \multicolumn{1}{c|}{ID test} & \multicolumn{1}{c|}{OOD test} & \multicolumn{1}{c|}{ID test} & \multicolumn{1}{c|}{OOD test} & \multicolumn{1}{c|}{ID test} & \multicolumn{1}{c|}{OOD test} & \multicolumn{1}{c|}{ID test} & \multicolumn{1}{c|}{OOD test} & ID test \\
\midrule[1pt]

\multicolumn{3}{l}{ERM} & 92.28\textpm0.10 & 51.28\textpm1.94 & \textbf{92.13}\textpm0.16 & 51.74\textpm2.27 & 91.73\textpm0.10 & 69.41\textpm0.91 & 91.78\textpm0.16 & \textbf{70.75}\textpm0.56 & 92.09\textpm0.04 \\
\multicolumn{3}{l}{IRM} & 92.18\textpm0.09 & 49.65\textpm1.31 & 91.99\textpm0.12 & 51.41\textpm3.30 & 91.68\textpm0.13 & 68.55\textpm1.79 & 91.70\textpm0.12 & 69.77\textpm0.88 & 92.04\textpm0.06 \\
\multicolumn{3}{l}{VREx} & 92.25\textpm0.08 & 48.87\textpm0.99 & 92.09\textpm0.14 & \textbf{52.67}\textpm2.87 & 91.67\textpm0.13 & 68.73\textpm1.23 & 91.76\textpm0.20 & 70.24\textpm0.72 & 92.09\textpm0.07 \\
\multicolumn{3}{l}{GroupDRO} & \textbf{92.29}\textpm0.09 & 49.21\textpm1.50 & 92.12\textpm0.10 & 51.95\textpm2.80 & 91.67\textpm0.14 & 68.28\textpm1.50 & 91.74\textpm0.15 & 69.98\textpm0.86 & 92.09\textpm0.08 \\
\multicolumn{3}{l}{DANN} & 92.23\textpm0.08 & 49.92\textpm2.63 & 92.04\textpm0.25 & 51.46\textpm3.41 & \textbf{91.81}\textpm0.16 & \textbf{69.68}\textpm1.40 & 91.69\textpm0.32 & 70.72\textpm1.16 & \textbf{92.10}\textpm0.06 \\
\multicolumn{3}{l}{Deep Coral} & 92.22\textpm0.13 & \textbf{52.70}\textpm3.04 & 92.05\textpm0.13 & 50.97\textpm1.76 & 91.68\textpm0.10 & 68.76\textpm0.95 & \textbf{91.78}\textpm0.09 & 70.49\textpm0.84 & 92.09\textpm0.07 \\
\cmidrule(r){1-12}
\multicolumn{3}{l}{Mixup} & 92.02\textpm0.10 & 49.98\textpm2.19 & 91.90\textpm0.14 & 51.48\textpm3.35 & 91.45\textpm0.13 & 66.42\textpm1.07 & 91.39\textpm0.22 & 67.81\textpm1.13 & 92.04\textpm0.06 \\
\multicolumn{3}{l}{DIR} & {84.53}\textpm1.99 & {42.61}\textpm1.31 & {77.07}\textpm4.06 & {50.41}\textpm5.66 & {73.10}\textpm5.89 & {53.21}\textpm4.03 & {72.31}\textpm5.49 & {56.28}\textpm5.51 & {73.46}\textpm0.85 \\

\bottomrule[2pt]
\end{tabular}
}
% }
\caption{Performance on GOOD-Motif with size domain.}
\label{table:res9}
\end{table*}

\begin{table*}[!htbp]
% \setlength{\abovecaptionskip}{0.1cm}
% \footnotesize
% \scalebox{0.622}{
% % \setlength{\tabcolsep}{0.3mm}{
% \centering
\centering
\resizebox{\textwidth}{!}
{
\begin{tabular}{lllccccccccc}
\toprule[2pt]
\multicolumn{3}{c}{\multirow{2}{*}{GOOD-Cora}} & \multicolumn{9}{c}{word}\\ 
\cmidrule(r){4-12}
& & & \multicolumn{4}{c|}{covariate} & \multicolumn{4}{c|}{concept} & \multicolumn{1}{c}{no shift} \\
\cmidrule(r){1-12}
\multicolumn{3}{c}{\multirow{2}{*}{Accuracy}} & \multicolumn{2}{c|}{ID validation} & \multicolumn{2}{c|}{OOD validation} & \multicolumn{2}{c|}{ID validation} & \multicolumn{2}{c|}{OOD validation} & \multicolumn{1}{c}{ID validation} \\
\cmidrule(r){4-12}
& & & \multicolumn{1}{c|}{ID test} & \multicolumn{1}{c|}{OOD test} & \multicolumn{1}{c|}{ID test} & \multicolumn{1}{c|}{OOD test} & \multicolumn{1}{c|}{ID test} & \multicolumn{1}{c|}{OOD test} & \multicolumn{1}{c|}{ID test} & \multicolumn{1}{c|}{OOD test} & ID test \\
\midrule[1pt]

\multicolumn{3}{l}{ERM} & 70.43\textpm0.47 & 64.44\textpm0.55 & 70.31\textpm0.39 & 64.86\textpm0.38 & 66.05\textpm0.22 & 64.20\textpm0.56 & 66.16\textpm0.37 & 64.60\textpm0.17 & 69.41\textpm0.30 \\
\multicolumn{3}{l}{IRM} & 70.27\textpm0.33 & \textbf{64.83}\textpm0.25 & 70.07\textpm0.23 & 64.77\textpm0.36 & 66.09\textpm0.32 & 64.16\textpm0.61 & 66.19\textpm0.36 & 64.60\textpm0.16 & 69.42\textpm0.38 \\
\multicolumn{3}{l}{VREx} & 70.47\textpm0.40 & 64.49\textpm0.55 & 70.35\textpm0.42 & 64.80\textpm0.28 & 66.00\textpm0.26 & 64.20\textpm0.54 & 66.37\textpm0.41 & 64.57\textpm0.18 & 69.43\textpm0.29 \\
\multicolumn{3}{l}{GroupDRO} & 70.41\textpm0.46 & 64.49\textpm0.66 & 70.38\textpm0.29 & 64.72\textpm0.34 & 66.17\textpm0.30 & {64.38}\textpm0.34 & 66.36\textpm0.44 & \textbf{64.62}\textpm0.17 & 69.46\textpm0.25 \\
\multicolumn{3}{l}{DANN} & 70.66\textpm0.36 & 64.72\textpm0.22 & 70.51\textpm0.47 & 64.77\textpm0.42 & 66.16\textpm0.31 & 64.29\textpm0.33 & 66.14\textpm0.41 & 64.51\textpm0.19 & 69.25\textpm0.33 \\
\multicolumn{3}{l}{Deep Coral} & 70.47\textpm0.37 & 64.63\textpm0.38 & 70.37\textpm0.32 & 64.72\textpm0.36 & 66.13\textpm0.18 & 64.38\textpm0.36 & 66.34\textpm0.40 & 64.58\textpm0.18 & 69.46\textpm0.27 \\
\cmidrule(r){1-12}
\multicolumn{3}{l}{Mixup} & \textbf{71.54}\textpm0.63 & 63.07\textpm1.52 & \textbf{72.14}\textpm0.70 & \textbf{65.23}\textpm0.56 & \textbf{69.66}\textpm0.45 & 64.22\textpm0.33 & \textbf{69.56}\textpm0.45 & 64.44\textpm0.10 & \textbf{70.56}\textpm0.35 \\
\multicolumn{3}{l}{EERM} & {68.79}\textpm0.34 & {60.80}\textpm0.61 & {69.23}\textpm0.13 & {61.98}\textpm0.10 & {65.75}\textpm0.15 & {63.35}\textpm0.03 & {65.88}\textpm0.21 & {63.09}\textpm0.36 & {70.10}\textpm0.22 \\
\multicolumn{3}{l}{SRGNN} & {70.27}\textpm0.23 & {64.49}\textpm0.19 & {70.15}\textpm0.24 & {64.66}\textpm0.21 & {66.45}\textpm0.09 & \textbf{64.90}\textpm0.03 & {65.77}\textpm0.14 & \textbf{64.62}\textpm0.07 & {69.05}\textpm0.54 \\

\bottomrule[2pt]
\end{tabular}
}
% }
\caption{Performance on GOOD-Cora with word domain.}
\label{table:res10}
\end{table*}

\begin{table*}[!htbp]
% \setlength{\abovecaptionskip}{0.1cm}
% \footnotesize
% \scalebox{0.622}{
% % \setlength{\tabcolsep}{0.3mm}{
% \centering
\centering
\resizebox{\textwidth}{!}
{
\begin{tabular}{lllccccccccc}
\toprule[2pt]
\multicolumn{3}{c}{\multirow{2}{*}{GOOD-Cora}} & \multicolumn{9}{c}{degree}\\ 
\cmidrule(r){4-12}
& & & \multicolumn{4}{c|}{covariate} & \multicolumn{4}{c|}{concept} & \multicolumn{1}{c}{no shift} \\
\cmidrule(r){1-12}
\multicolumn{3}{c}{\multirow{2}{*}{Accuracy}} & \multicolumn{2}{c|}{ID validation} & \multicolumn{2}{c|}{OOD validation} & \multicolumn{2}{c|}{ID validation} & \multicolumn{2}{c|}{OOD validation} & \multicolumn{1}{c}{ID validation} \\
\cmidrule(r){4-12}
& & & \multicolumn{1}{c|}{ID test} & \multicolumn{1}{c|}{OOD test} & \multicolumn{1}{c|}{ID test} & \multicolumn{1}{c|}{OOD test} & \multicolumn{1}{c|}{ID test} & \multicolumn{1}{c|}{OOD test} & \multicolumn{1}{c|}{ID test} & \multicolumn{1}{c|}{OOD test} & ID test \\
\midrule[1pt]

\multicolumn{3}{l}{ERM} & 72.27\textpm0.57 & 55.76\textpm0.82 & 72.51\textpm0.57 & 56.30\textpm0.49 & 68.71\textpm0.56 & 60.38\textpm0.33 & 68.43\textpm0.28 & 60.54\textpm0.44 & 69.42\textpm0.30 \\
\multicolumn{3}{l}{IRM} & 72.64\textpm0.45 & 55.77\textpm0.46 & 72.75\textpm0.36 & 56.28\textpm0.63 & 68.58\textpm0.40 & 61.00\textpm0.34 & 68.53\textpm0.38 & 61.23\textpm0.32 & 69.40\textpm0.38 \\
\multicolumn{3}{l}{VREx} & 72.25\textpm0.65 & 55.46\textpm0.87 & 72.49\textpm0.59 & 56.30\textpm0.50 & 68.45\textpm0.44 & 60.05\textpm0.72 & 68.37\textpm0.33 & 60.58\textpm0.42 & 69.42\textpm0.29 \\
\multicolumn{3}{l}{GroupDRO} & 72.18\textpm0.58 & 55.44\textpm0.91 & 72.66\textpm0.41 & 56.29\textpm0.43 & 68.37\textpm0.79 & 60.03\textpm0.88 & 68.34\textpm0.25 & 60.65\textpm0.31 & 69.40\textpm0.30 \\
\multicolumn{3}{l}{DANN} & 72.47\textpm0.37 & 55.50\textpm0.60 & 72.51\textpm0.42 & 56.10\textpm0.59 & 68.08\textpm1.05 & 59.65\textpm0.94 & 68.51\textpm0.36 & 60.78\textpm0.38 & 69.24\textpm0.34 \\
\multicolumn{3}{l}{Deep Coral} & 72.16\textpm0.53 & 55.52\textpm0.93 & 72.57\textpm0.37 & 56.35\textpm0.38 & 68.38\textpm0.76 & 60.22\textpm0.55 & 68.30\textpm0.30 & 60.58\textpm0.40 & 69.43\textpm0.30 \\
\cmidrule(r){1-12}
\multicolumn{3}{l}{Mixup} & \textbf{74.57}\textpm0.54 & \textbf{57.21}\textpm1.12 & \textbf{74.34}\textpm0.56 & \textbf{58.20}\textpm0.67 & \textbf{70.32}\textpm0.59 & \textbf{63.49}\textpm0.23 & \textbf{70.44}\textpm0.53 & \textbf{63.65}\textpm0.39 & \textbf{70.87}\textpm0.47 \\
\multicolumn{3}{l}{EERM} & {73.32}\textpm0.06 & {55.23}\textpm0.40 & {73.47}\textpm0.02 & {56.88}\textpm0.32 & {66.50}\textpm0.53 & {57.46}\textpm0.87 & {66.84}\textpm0.62 & {58.38}\textpm0.04 & {70.38}\textpm0.24 \\
\multicolumn{3}{l}{SRGNN} & {71.37}\textpm0.04 & {54.67}\textpm0.36 & {71.20}\textpm0.47 & {54.78}\textpm0.10 & {68.34}\textpm0.90 & {59.96}\textpm0.89 & {68.94}\textpm0.29 & {61.08}\textpm0.09 & {69.08}\textpm0.53 \\

\bottomrule[2pt]
\end{tabular}
}
% }
\caption{Performance on GOOD-Cora with degree domain.}
\label{table:res11}
\end{table*}

\begin{table*}[!htbp]
% \setlength{\abovecaptionskip}{0.1cm}
% \footnotesize
% \scalebox{0.622}{
% % \setlength{\tabcolsep}{0.3mm}{
% \centering
\centering
\resizebox{\textwidth}{!}
{
\begin{tabular}{lllccccccccc}
\toprule[2pt]
\multicolumn{3}{c}{\multirow{2}{*}{GOOD-Arxiv}} & \multicolumn{9}{c}{time}\\ 
\cmidrule(r){4-12}
& & & \multicolumn{4}{c|}{covariate} & \multicolumn{4}{c|}{concept} & \multicolumn{1}{c}{no shift} \\
\cmidrule(r){1-12}
\multicolumn{3}{c}{\multirow{2}{*}{Accuracy}} & \multicolumn{2}{c|}{ID validation} & \multicolumn{2}{c|}{OOD validation} & \multicolumn{2}{c|}{ID validation} & \multicolumn{2}{c|}{OOD validation} & \multicolumn{1}{c}{ID validation} \\
\cmidrule(r){4-12}
& & & \multicolumn{1}{c|}{ID test} & \multicolumn{1}{c|}{OOD test} & \multicolumn{1}{c|}{ID test} & \multicolumn{1}{c|}{OOD test} & \multicolumn{1}{c|}{ID test} & \multicolumn{1}{c|}{OOD test} & \multicolumn{1}{c|}{ID test} & \multicolumn{1}{c|}{OOD test} & ID test \\
\midrule[1pt]

\multicolumn{3}{l}{ERM} & 72.69\textpm0.19 & 70.64\textpm0.47 & 72.66\textpm0.17 & 71.08\textpm0.23 & 74.76\textpm0.18 & \textbf{65.70}\textpm0.42 & 73.68\textpm0.49 & 67.32\textpm0.24 & 73.02\textpm0.14 \\
\multicolumn{3}{l}{IRM} & 72.66\textpm0.15 & 70.55\textpm0.33 & 72.58\textpm0.20 & 71.04\textpm0.16 & 74.67\textpm0.15 & 65.69\textpm0.55 & 73.53\textpm0.46 & 67.41\textpm0.16 & 72.90\textpm0.14 \\
\multicolumn{3}{l}{VREx} & 72.66\textpm0.18 & 70.54\textpm0.33 & 72.58\textpm0.21 & 71.12\textpm0.24 & 74.80\textpm0.14 & 65.40\textpm0.54 & 73.72\textpm0.43 & 67.37\textpm0.27 & 72.84\textpm0.09 \\
\multicolumn{3}{l}{GroupDRO} & 72.68\textpm0.17 & 70.67\textpm0.31 & 72.46\textpm0.26 & 71.15\textpm0.20 & 74.73\textpm0.18 & 65.57\textpm0.66 & 73.55\textpm0.34 & \textbf{67.45}\textpm0.15 & 72.91\textpm0.12 \\
\multicolumn{3}{l}{DANN} & \textbf{72.74}\textpm0.11 & 70.57\textpm0.40 & \textbf{72.67}\textpm0.20 & 71.05\textpm0.29 & 74.73\textpm0.15 & 65.42\textpm0.53 & 73.99\textpm0.35 & 67.28\textpm0.16 & 73.00\textpm0.12 \\
\multicolumn{3}{l}{Deep Coral} & 72.66\textpm0.18 & 70.59\textpm0.29 & 72.54\textpm0.09 & 71.07\textpm0.21 & 74.77\textpm0.16 & 65.53\textpm0.63 & 73.40\textpm0.32 & 67.42\textpm0.22 & 72.95\textpm0.09 \\
\cmidrule(r){1-12}
\multicolumn{3}{l}{Mixup} & 72.49\textpm0.26 & \textbf{71.05}\textpm0.31 & 72.55\textpm0.23 & \textbf{71.34}\textpm0.14 & \textbf{74.92}\textpm0.32 & 64.01\textpm0.50 & \textbf{74.55}\textpm0.18 & 64.84\textpm0.59 & \textbf{73.19}\textpm0.16 \\
\multicolumn{3}{l}{EERM} & \multicolumn{9}{c}{out of memory} \\
\multicolumn{3}{l}{SRGNN} & {72.50}\textpm0.09 & {70.70}\textpm0.42 & {72.34}\textpm0.08 & {70.83}\textpm0.10 & {74.64}\textpm0.10 & {65.37}\textpm0.22 & {73.88}\textpm0.14 & {67.17}\textpm0.23 & {72.99}\textpm0.04 \\

\bottomrule[2pt]
\end{tabular}
}
% }
\caption{Performance on GOOD-Arxiv with time domain.}
\label{table:res12}
\end{table*}

\begin{table*}[!htbp]
% \setlength{\abovecaptionskip}{0.1cm}
% \footnotesize
% \scalebox{0.622}{
% % \setlength{\tabcolsep}{0.3mm}{
% \centering
\centering
\resizebox{\textwidth}{!}
{
\begin{tabular}{lllccccccccc}
\toprule[2pt]
\multicolumn{3}{c}{\multirow{2}{*}{GOOD-Arxiv}} & \multicolumn{9}{c}{degree}\\ 
\cmidrule(r){4-12}
& & & \multicolumn{4}{c|}{covariate} & \multicolumn{4}{c|}{concept} & \multicolumn{1}{c}{no shift} \\
\cmidrule(r){1-12}
\multicolumn{3}{c}{\multirow{2}{*}{Accuracy}} & \multicolumn{2}{c|}{ID validation} & \multicolumn{2}{c|}{OOD validation} & \multicolumn{2}{c|}{ID validation} & \multicolumn{2}{c|}{OOD validation} & \multicolumn{1}{c}{ID validation} \\
\cmidrule(r){4-12}
& & & \multicolumn{1}{c|}{ID test} & \multicolumn{1}{c|}{OOD test} & \multicolumn{1}{c|}{ID test} & \multicolumn{1}{c|}{OOD test} & \multicolumn{1}{c|}{ID test} & \multicolumn{1}{c|}{OOD test} & \multicolumn{1}{c|}{ID test} & \multicolumn{1}{c|}{OOD test} & ID test \\
\midrule[1pt]

\multicolumn{3}{l}{ERM} & 77.47\textpm0.12 & 58.53\textpm0.16 & 77.18\textpm0.23 & 58.91\textpm0.23 & \textbf{75.27}\textpm0.16 & \textbf{61.77}\textpm0.29 & 74.74\textpm0.19 & 62.99\textpm0.20 & 72.99\textpm0.12 \\
\multicolumn{3}{l}{IRM} & 77.50\textpm0.11 & \textbf{58.70}\textpm0.12 & 77.15\textpm0.29 & 58.98\textpm0.28 & 75.23\textpm0.11 & 61.49\textpm0.36 & 74.64\textpm0.42 & 62.97\textpm0.27 & 72.92\textpm0.07 \\
\multicolumn{3}{l}{VREx} & 77.49\textpm0.11 & 58.59\textpm0.21 & 77.33\textpm0.17 & 58.99\textpm0.16 & 75.19\textpm0.14 & 61.61\textpm0.32 & 74.64\textpm0.22 & \textbf{63.00}\textpm0.33 & 72.88\textpm0.09 \\
\multicolumn{3}{l}{GroupDRO} & 77.46\textpm0.18 & 58.46\textpm0.21 & 77.16\textpm0.20 & \textbf{59.08}\textpm0.16 & 75.19\textpm0.14 & 61.59\textpm0.56 & \textbf{74.92}\textpm0.20 & 62.88\textpm0.24 & 72.98\textpm0.10 \\
\multicolumn{3}{l}{DANN} & 77.51\textpm0.08 & 58.56\textpm0.16 & 77.19\textpm0.29 & 59.00\textpm0.18 & 75.25\textpm0.08 & 61.43\textpm0.40 & 74.76\textpm0.25 & 62.91\textpm0.22 & 72.97\textpm0.10 \\
\multicolumn{3}{l}{Deep Coral} & 77.48\textpm0.13 & 58.63\textpm0.21 & 77.16\textpm0.26 & 58.97\textpm0.20 & 75.16\textpm0.15 & 61.77\textpm0.37 & 74.89\textpm0.12 & 62.85\textpm0.29 & 72.91\textpm0.12 \\
\cmidrule(r){1-12}
\multicolumn{3}{l}{Mixup} & \textbf{77.61}\textpm0.15 & 57.43\textpm0.27 & \textbf{77.47}\textpm0.29 & 57.60\textpm0.31 & 72.75\textpm0.38 & 60.60\textpm1.01 & 72.31\textpm0.84 & 61.28\textpm0.87 & \textbf{73.03}\textpm0.14 \\
\multicolumn{3}{l}{EERM} & \multicolumn{9}{c}{out of memory} \\
\multicolumn{3}{l}{SRGNN} & {75.96}\textpm0.08 & {57.48}\textpm0.07 & {75.98}\textpm0.13 & {57.52}\textpm0.10 & {74.83}\textpm0.20 & {61.74}\textpm0.10 & {74.30}\textpm0.07 & {62.09}\textpm0.58 & {72.99}\textpm0.02 \\

\bottomrule[2pt]
\end{tabular}
}
% }
\caption{Performance on GOOD-Arxiv with degree domain.}
\label{table:res13}
\end{table*}

\begin{table*}[!htbp]
% \setlength{\abovecaptionskip}{0.1cm}
% \footnotesize
% \scalebox{0.622}{
% % \setlength{\tabcolsep}{0.3mm}{
% \centering
\centering
\resizebox{\textwidth}{!}
{
\begin{tabular}{lllccccccccc}
\toprule[2pt]
\multicolumn{3}{c}{\multirow{2}{*}{GOOD-Twitch}} & \multicolumn{9}{c}{language}\\ 
\cmidrule(r){4-12}
& & & \multicolumn{4}{c|}{covariate} & \multicolumn{4}{c|}{concept} & \multicolumn{1}{c}{no shift} \\
\cmidrule(r){1-12}
\multicolumn{3}{c}{\multirow{2}{*}{Accuracy}} & \multicolumn{2}{c|}{ID validation} & \multicolumn{2}{c|}{OOD validation} & \multicolumn{2}{c|}{ID validation} & \multicolumn{2}{c|}{OOD validation} & \multicolumn{1}{c}{ID validation} \\
\cmidrule(r){4-12}
& & & \multicolumn{1}{c|}{ID test} & \multicolumn{1}{c|}{OOD test} & \multicolumn{1}{c|}{ID test} & \multicolumn{1}{c|}{OOD test} & \multicolumn{1}{c|}{ID test} & \multicolumn{1}{c|}{OOD test} & \multicolumn{1}{c|}{ID test} & \multicolumn{1}{c|}{OOD test} & ID test \\
\midrule[1pt]

\multicolumn{3}{l}{ERM} & 70.66\textpm0.17 & 47.73\textpm0.72 & 69.40\textpm0.49 & 48.95\textpm3.19 & 80.29\textpm1.01 & 48.57\textpm0.17 & 71.14\textpm1.49 & 57.32\textpm0.18 & 68.05\textpm0.52 \\
\multicolumn{3}{l}{IRM} & 69.75\textpm0.80 & 48.05\textpm0.16 & 67.92\textpm0.48 & 47.21\textpm0.98 & 77.05\textpm0.60 & 49.77\textpm0.82 & 67.35\textpm0.84 & 59.17\textpm0.85 & 68.30\textpm0.29 \\
\multicolumn{3}{l}{VREx} & 70.66\textpm0.18 & 47.70\textpm0.70 & 69.42\textpm0.48 & 48.99\textpm3.20 & 80.29\textpm1.01 & 48.56\textpm0.18 & 71.17\textpm1.35 & 57.37\textpm0.14 & 68.07\textpm0.52 \\
\multicolumn{3}{l}{GroupDRO} & 70.84\textpm0.51 & 47.23\textpm0.26 & 67.66\textpm1.64 & 47.20\textpm0.44 & {81.95}\textpm0.88 & 47.44\textpm1.08 & 69.74\textpm0.33 & \textbf{60.27}\textpm0.62 & {69.19}\textpm0.28 \\
\multicolumn{3}{l}{DANN} & 70.67\textpm0.18 & 47.72\textpm0.73 & 69.42\textpm0.48 & 48.98\textpm3.22 & 80.28\textpm0.99 & 48.57\textpm0.18 & 70.94\textpm1.43 & 57.46\textpm0.14 & 68.07\textpm0.52 \\
\multicolumn{3}{l}{Deep Coral} & 70.67\textpm0.28 & 46.64\textpm0.70 & 68.72\textpm0.71 & 49.64\textpm2.44 & 80.14\textpm0.49 & 47.46\textpm0.32 & 69.70\textpm0.68 & 56.97\textpm0.23 & 68.29\textpm0.65 \\
\cmidrule(r){1-12}
\multicolumn{3}{l}{Mixup} & 71.30\textpm0.14 & {51.33}\textpm1.50 & {70.39}\textpm0.62 & \textbf{52.27}\textpm0.78 & 78.89\textpm0.60 & \textbf{51.87}\textpm0.37 & 69.08\textpm0.59 & 55.28\textpm0.12 & 67.09\textpm0.34 \\
\multicolumn{3}{l}{EERM} & \textbf{73.87}\textpm0.07 & \textbf{52.48}\textpm0.76 & \textbf{72.52}\textpm0.08 & 51.34\textpm1.41 & \textbf{83.91}\textpm0.15 & 44.22\textpm0.81 & \textbf{76.28}\textpm5.81 & 51.94\textpm4.52 & \textbf{70.80}\textpm0.08 \\
\multicolumn{3}{l}{SRGNN} & 70.58\textpm0.53 & 46.17\textpm0.98 & 70.02\textpm0.35 & 47.30\textpm1.43 & 80.21\textpm0.59 & 48.27\textpm1.10 & {71.73}\textpm1.13 & 56.05\textpm0.22 & 67.69\textpm0.13 \\

\bottomrule[2pt]
\end{tabular}
}
% }
\caption{Performance on GOOD-Twitch with language domain.}
\label{table:res14}
\end{table*}

\begin{table*}[!htbp]
% \setlength{\abovecaptionskip}{0.1cm}
% \footnotesize
% \scalebox{0.622}{
% % \setlength{\tabcolsep}{0.3mm}{
% \centering
\centering
\resizebox{\textwidth}{!}
{
\begin{tabular}{lllccccccccc}
\toprule[2pt]
\multicolumn{3}{c}{\multirow{2}{*}{GOOD-WebKB}} & \multicolumn{9}{c}{university}\\ 
\cmidrule(r){4-12}
& & & \multicolumn{4}{c|}{covariate} & \multicolumn{4}{c|}{concept} & \multicolumn{1}{c}{no shift} \\
\cmidrule(r){1-12}
\multicolumn{3}{c}{\multirow{2}{*}{Accuracy}} & \multicolumn{2}{c|}{ID validation} & \multicolumn{2}{c|}{OOD validation} & \multicolumn{2}{c|}{ID validation} & \multicolumn{2}{c|}{OOD validation} & \multicolumn{1}{c}{ID validation} \\
\cmidrule(r){4-12}
& & & \multicolumn{1}{c|}{ID test} & \multicolumn{1}{c|}{OOD test} & \multicolumn{1}{c|}{ID test} & \multicolumn{1}{c|}{OOD test} & \multicolumn{1}{c|}{ID test} & \multicolumn{1}{c|}{OOD test} & \multicolumn{1}{c|}{ID test} & \multicolumn{1}{c|}{OOD test} & ID test \\
\midrule[1pt]

\multicolumn{3}{l}{ERM} & 38.25\textpm0.68 & 11.64\textpm0.90 & 40.98\textpm3.54 & 14.29\textpm3.24 & 65.00\textpm2.72 & 24.77\textpm0.43 & 62.22\textpm0.95 & 27.83\textpm0.76 & 47.85\textpm0.89 \\
\multicolumn{3}{l}{IRM} & 39.34\textpm2.04 & 11.91\textpm2.62 & 45.90\textpm0.77 & 13.49\textpm0.75 & 65.56\textpm3.40 & 24.16\textpm0.80 & 60.00\textpm0.79 & 27.52\textpm0.43 & 47.31\textpm1.21 \\
\multicolumn{3}{l}{VREx} & 39.34\textpm1.34 & 10.58\textpm1.02 & 40.44\textpm2.97 & 14.29\textpm3.24 & 65.00\textpm2.72 & 24.77\textpm0.43 & 59.45\textpm2.77 & 27.83\textpm0.38 & 47.85\textpm0.89 \\
\multicolumn{3}{l}{GroupDRO} & 39.89\textpm1.57 & 12.96\textpm1.95 & 39.34\textpm2.04 & 17.20\textpm0.76 & 65.00\textpm2.72 & 24.77\textpm0.43 & 57.78\textpm4.12 & 28.14\textpm1.12 & 47.85\textpm0.89 \\
\multicolumn{3}{l}{DANN} & 39.89\textpm1.03 & {15.34}\textpm1.02 & 41.53\textpm2.87 & 15.08\textpm0.37 & 65.00\textpm2.72 & 24.77\textpm0.43 & 59.45\textpm0.95 & 26.91\textpm0.63 & 47.85\textpm0.89 \\
\multicolumn{3}{l}{Deep Coral} & 38.25\textpm1.43 & 14.29\textpm2.92 & 46.45\textpm3.35 & 13.76\textpm1.30 & 65.00\textpm2.72 & 24.77\textpm0.43 & 62.78\textpm0.26 & 28.75\textpm1.13 & 48.12\textpm0.89 \\
\cmidrule(r){1-12}
\multicolumn{3}{l}{Mixup} & \textbf{54.65}\textpm3.41 & 10.85\textpm0.66 & \textbf{57.38}\textpm0.77 & {17.46}\textpm1.94 & \textbf{67.22}\textpm1.14 & \textbf{27.83}\textpm1.53 & \textbf{71.67}\textpm0.00 & \textbf{31.19}\textpm0.43 & {51.88}\textpm1.34 \\
\multicolumn{3}{l}{EERM} & {46.99}\textpm1.69 & {11.90}\textpm0.37 & {33.88}\textpm4.92 & \textbf{24.61}\textpm4.86 & {61.67}\textpm2.08 & {24.77}\textpm0.43 & {61.11}\textpm1.46 & {27.83}\textpm4.12 & {50.54}\textpm0.46 \\
\multicolumn{3}{l}{SRGNN} & {39.89}\textpm1.36 & \textbf{16.14}\textpm3.35 & {38.25}\textpm1.57 & {13.23}\textpm2.93 & {61.67}\textpm0.00 & {25.08}\textpm1.13 & {61.11}\textpm0.26 & {27.52}\textpm0.43 & \textbf{52.96}\textpm1.04 \\

\bottomrule[2pt]
\end{tabular}
}
% }
\caption{Performance on GOOD-WebKB with university domain.}
\label{table:res15}
\end{table*}

\begin{table*}[!htbp]
% \setlength{\abovecaptionskip}{0.1cm}
% \footnotesize
% \scalebox{0.622}{
% % \setlength{\tabcolsep}{0.3mm}{
% \centering
\centering
\resizebox{\textwidth}{!}
{
\begin{tabular}{lllccccccccc}
\toprule[2pt]
\multicolumn{3}{c}{\multirow{2}{*}{GOOD-CBAS}} & \multicolumn{9}{c}{color}\\ 
\cmidrule(r){4-12}
& & & \multicolumn{4}{c|}{covariate} & \multicolumn{4}{c|}{concept} & \multicolumn{1}{c}{no shift} \\
\cmidrule(r){1-12}
\multicolumn{3}{c}{\multirow{2}{*}{Accuracy}} & \multicolumn{2}{c|}{ID validation} & \multicolumn{2}{c|}{OOD validation} & \multicolumn{2}{c|}{ID validation} & \multicolumn{2}{c|}{OOD validation} & \multicolumn{1}{c}{ID validation} \\
\cmidrule(r){4-12}
& & & \multicolumn{1}{c|}{ID test} & \multicolumn{1}{c|}{OOD test} & \multicolumn{1}{c|}{ID test} & \multicolumn{1}{c|}{OOD test} & \multicolumn{1}{c|}{ID test} & \multicolumn{1}{c|}{OOD test} & \multicolumn{1}{c|}{ID test} & \multicolumn{1}{c|}{OOD test} & ID test \\
\midrule[1pt]

\multicolumn{3}{l}{ERM} & 89.29\textpm3.16 & 77.57\textpm2.96 & \textbf{89.72}\textpm3.20 & 76.00\textpm3.00 & 89.79\textpm1.18 & 82.22\textpm1.81 & 90.14\textpm1.10 & 82.36\textpm0.97 & 99.43\textpm0.45 \\
\multicolumn{3}{l}{IRM} & 91.00\textpm1.28 & 77.00\textpm2.21 & 87.43\textpm4.05 & 76.00\textpm3.39 & 90.71\textpm0.87 & 81.50\textpm1.46 & 90.21\textpm0.91 & \textbf{83.21}\textpm0.54 & 99.64\textpm0.46 \\
\multicolumn{3}{l}{VREx} & \textbf{91.14}\textpm2.72 & 77.71\textpm2.03 & 88.43\textpm1.81 & 77.14\textpm1.43 & 89.50\textpm1.13 & \textbf{82.50}\textpm1.47 & 90.21\textpm0.96 & 82.86\textpm1.26 & 99.64\textpm0.46 \\
\multicolumn{3}{l}{GroupDRO} & 90.86\textpm2.92 & 77.71\textpm2.00 & 89.71\textpm2.12 & 76.14\textpm1.78 & 90.36\textpm0.91 & 81.22\textpm1.78 & 91.00\textpm1.01 & 82.00\textpm1.46 & 99.72\textpm0.33 \\
\multicolumn{3}{l}{DANN} & 90.14\textpm3.16 & \textbf{79.14}\textpm2.40 & 86.71\textpm4.78 & \textbf{77.57}\textpm2.86 & 89.93\textpm1.25 & 80.50\textpm1.31 & 89.78\textpm1.01 & 82.50\textpm0.72 & 99.65\textpm0.33 \\
\multicolumn{3}{l}{Deep Coral} & \textbf{91.14}\textpm2.02 & 77.86\textpm2.22 & 88.14\textpm2.43 & 75.86\textpm3.06 & 89.36\textpm1.87 & 81.93\textpm1.36 & 90.14\textpm0.98 & 82.64\textpm1.40 & {99.79}\textpm0.28 \\
\cmidrule(r){1-12}
\multicolumn{3}{l}{Mixup} & 73.57\textpm8.72 & 73.72\textpm6.60 & 73.00\textpm9.27 & 70.57\textpm7.41 & \textbf{93.64}\textpm0.57 & 63.57\textpm1.43 & \textbf{92.86}\textpm1.19 & 64.57\textpm1.81 & 98.43\textpm1.72 \\
\multicolumn{3}{l}{EERM} & {67.62}\textpm4.08 & {68.10}\textpm4.12 & {57.62}\textpm7.19 & {52.86}\textpm13.75 & {78.33}\textpm0.11 & {63.10}\textpm0.96 & {80.48}\textpm0.49 & {64.29}\textpm0.00 & {89.05}\textpm0.30 \\
\multicolumn{3}{l}{SRGNN} & {77.62}\textpm1.84 & {73.81}\textpm1.75 & {82.86}\textpm1.78 & {74.29}\textpm4.10 & {88.57}\textpm0.58 & {80.24}\textpm0.49 & {89.76}\textpm0.96 & {81.43}\textpm0.34 & \textbf{100.00}\textpm0.00 \\

\bottomrule[2pt]
\end{tabular}
}
% }
\caption{Performance on GOOD-CBAS with color domain.}
\label{table:res16}
\end{table*}

\subsection{Metric score curves}
\begin{figure}[!htbp]
    \centering
\begin{subfigure}[t]{0.32\textwidth}
    \raisebox{-\height}{\includegraphics[width=\textwidth]{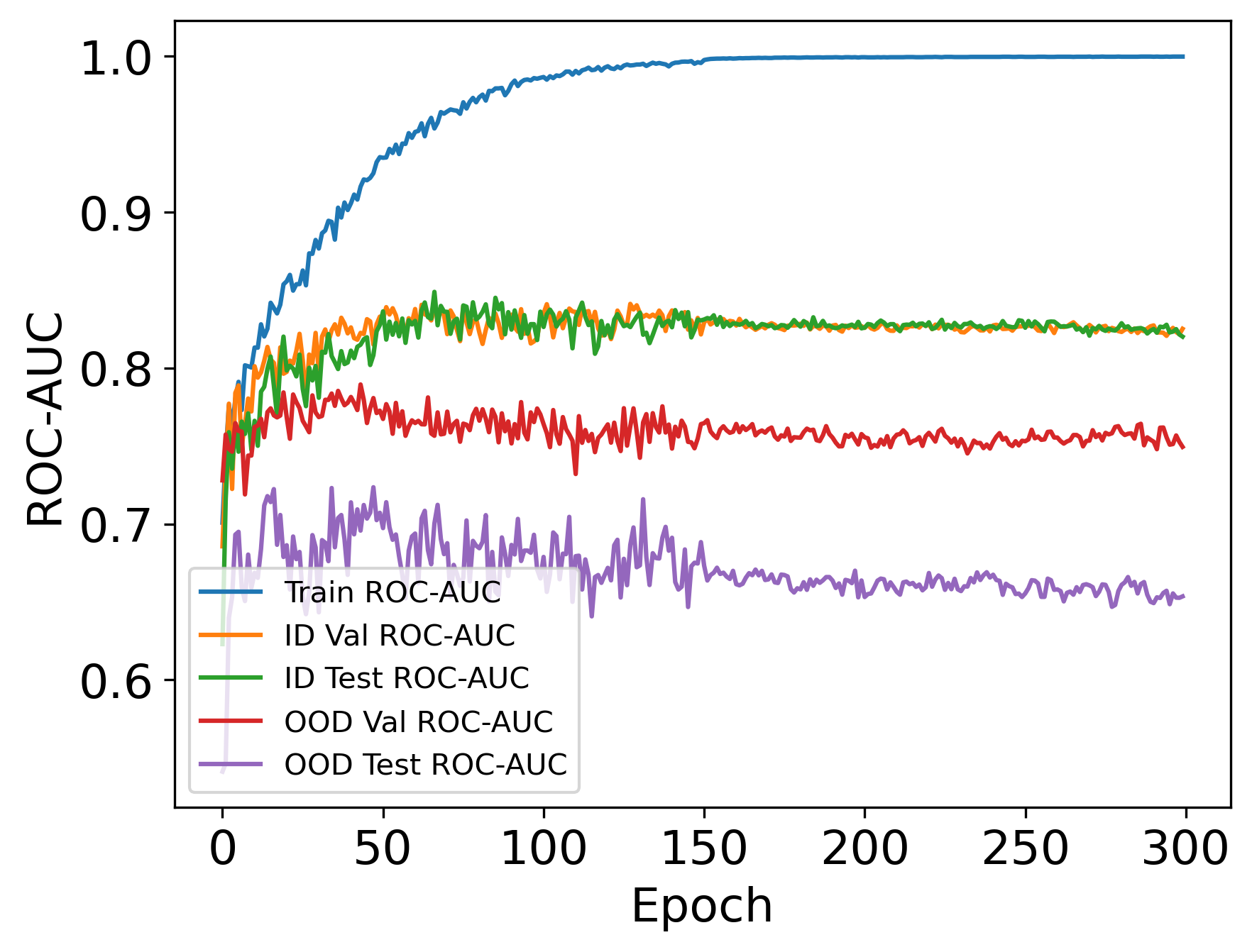}}
    \caption{covariate shift + scaffold domain}
\end{subfigure}
\hfill
\begin{subfigure}[t]{0.32\textwidth}
    \raisebox{-\height}{\includegraphics[width=\textwidth]{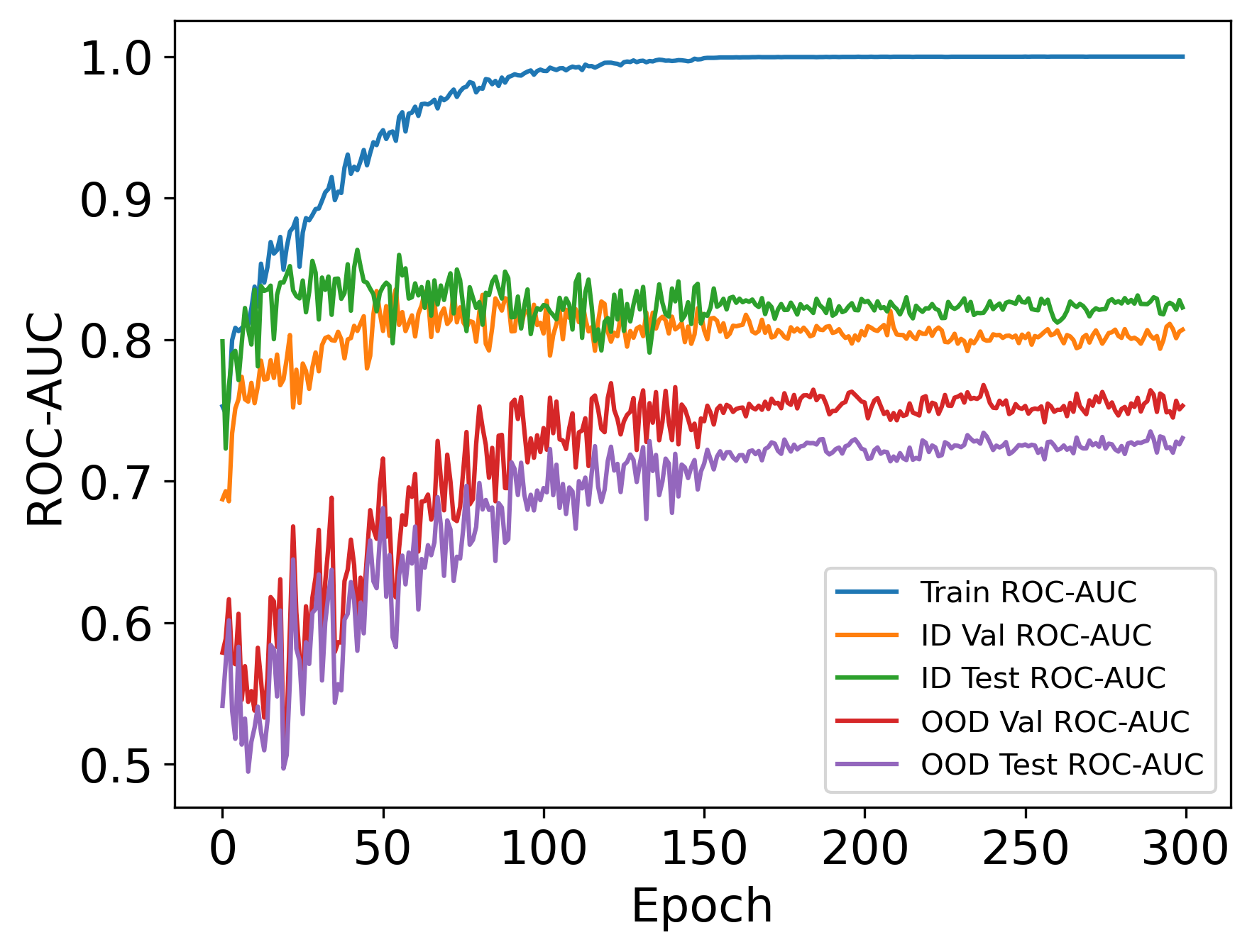}}
    \caption{concept shift + scaffold domain}
\end{subfigure}
\hfill
\begin{subfigure}[t]{0.32\textwidth}
    \raisebox{-\height}{\includegraphics[width=\textwidth]{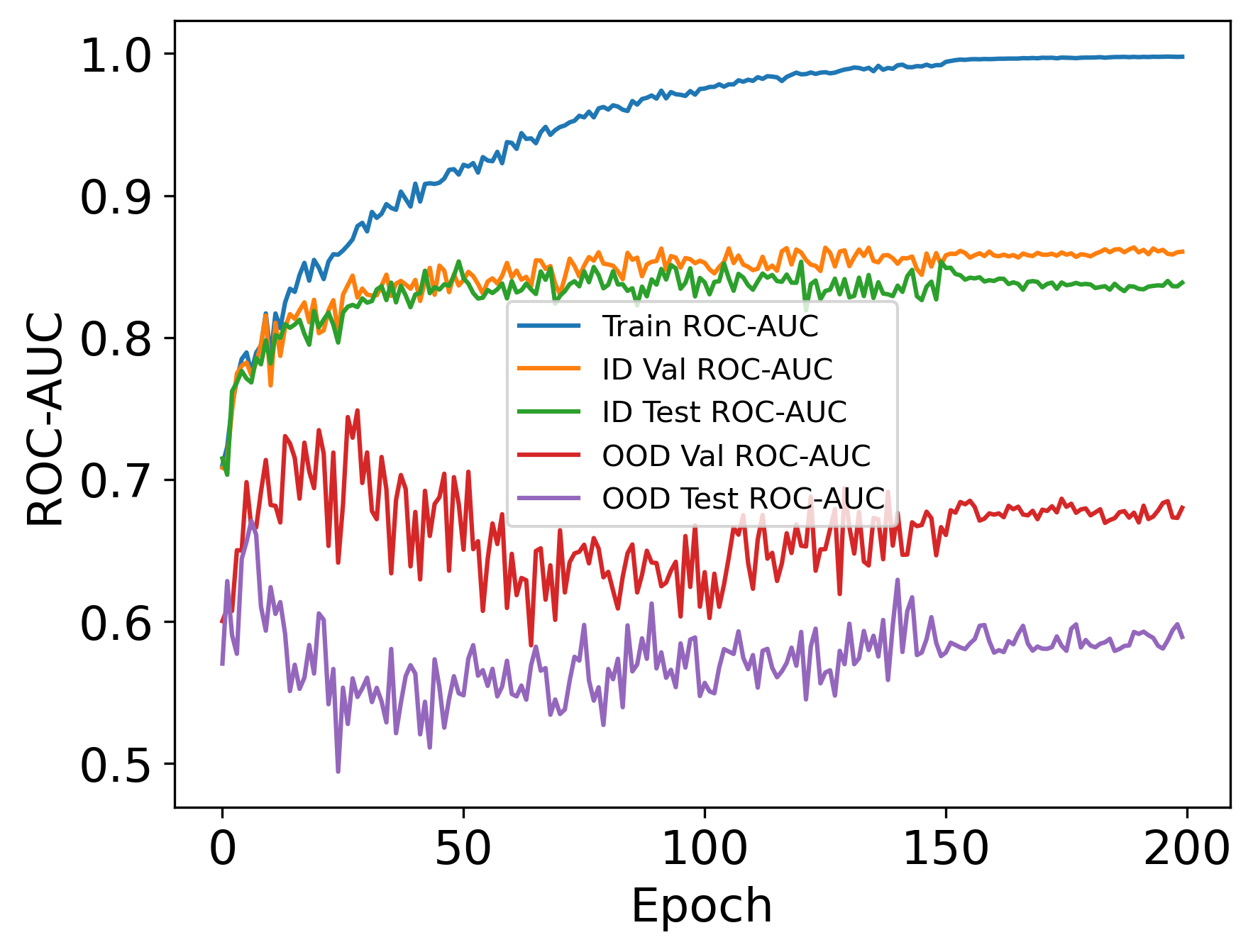}}
    \caption{covariate shift + size domain}
\end{subfigure}
\hspace*{\fill}
\begin{subfigure}[t]{0.32\textwidth}
    \raisebox{-\height}{\includegraphics[width=\textwidth]{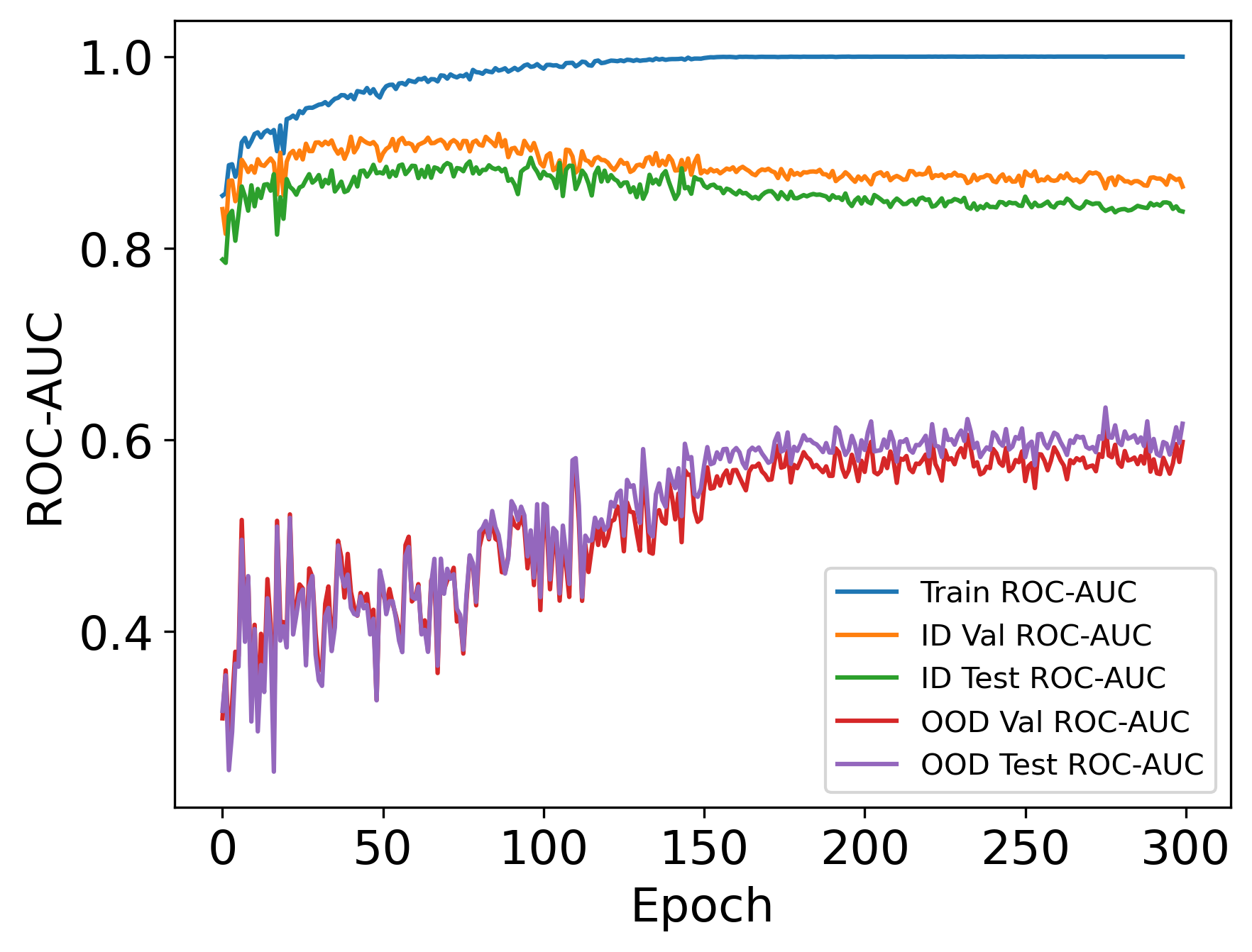}}
    \caption{concept shift + size domain}
\end{subfigure}
\hfill
\begin{subfigure}[t]{0.32\textwidth}
    \raisebox{-\height}{\includegraphics[width=\textwidth]{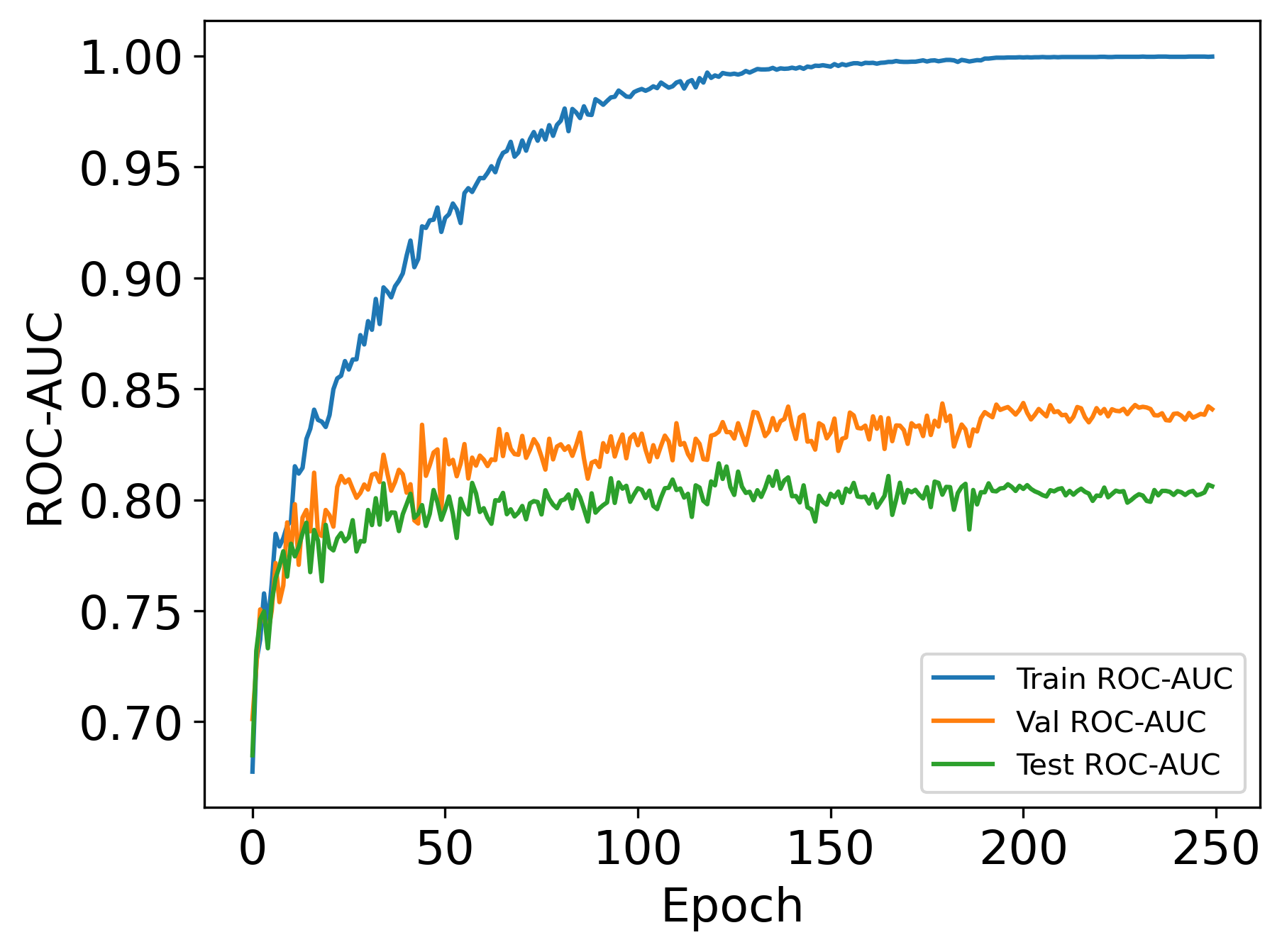}}
    \caption{no shift}
\end{subfigure}
\hspace*{\fill}
\caption{Metric score curves for ERM on GOOD-HIV. Note that we omit the domain selection for no shift since the two cases make no difference in results.}
    \label{fig:curve1}
\end{figure}

\begin{figure}[!htbp]
    \centering
\begin{subfigure}[t]{0.32\textwidth}
    \raisebox{-\height}{\includegraphics[width=\textwidth]{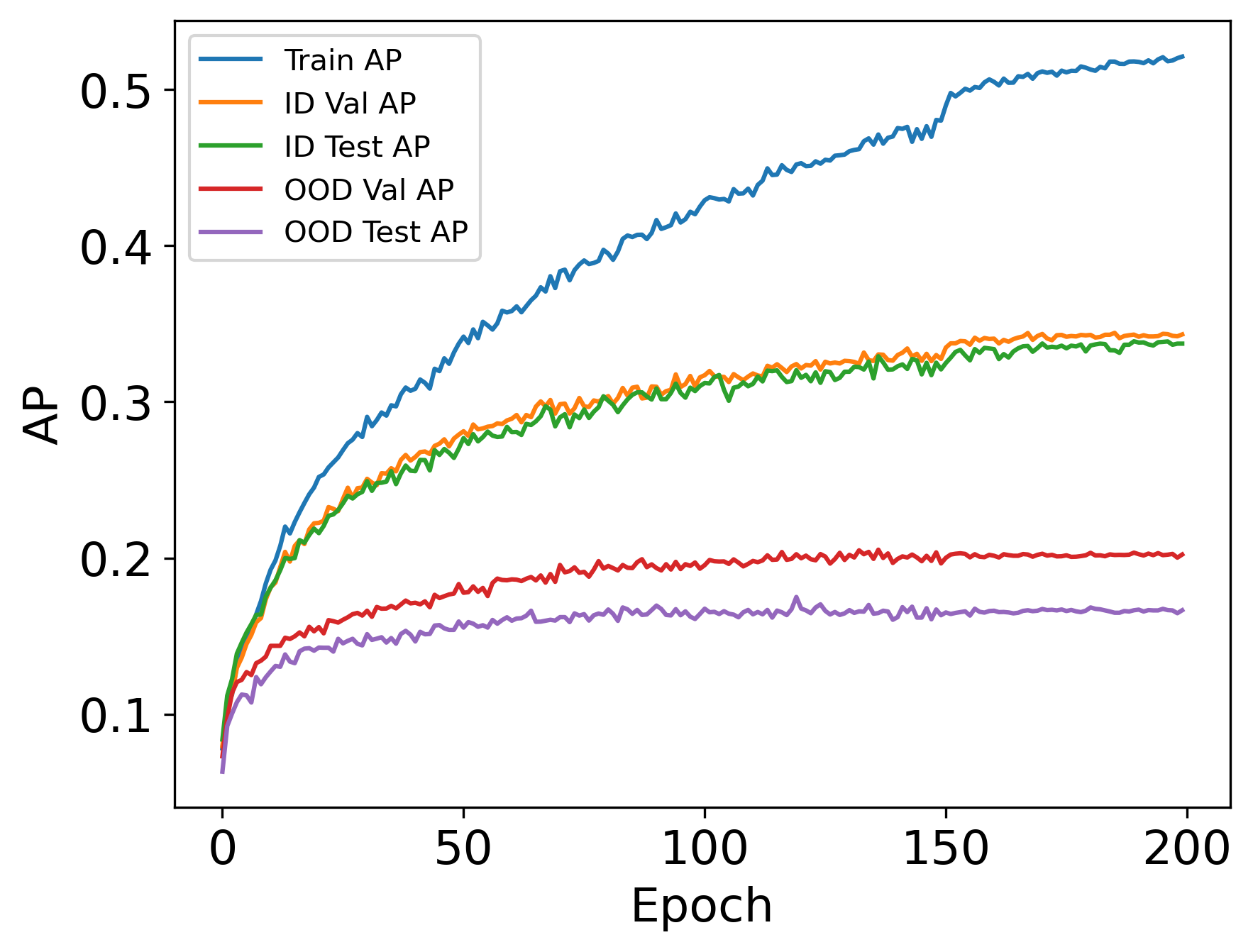}}
    \caption{scaffold domain, covariate shift}
\end{subfigure}
\hfill
\begin{subfigure}[t]{0.32\textwidth}
    \raisebox{-\height}{\includegraphics[width=\textwidth]{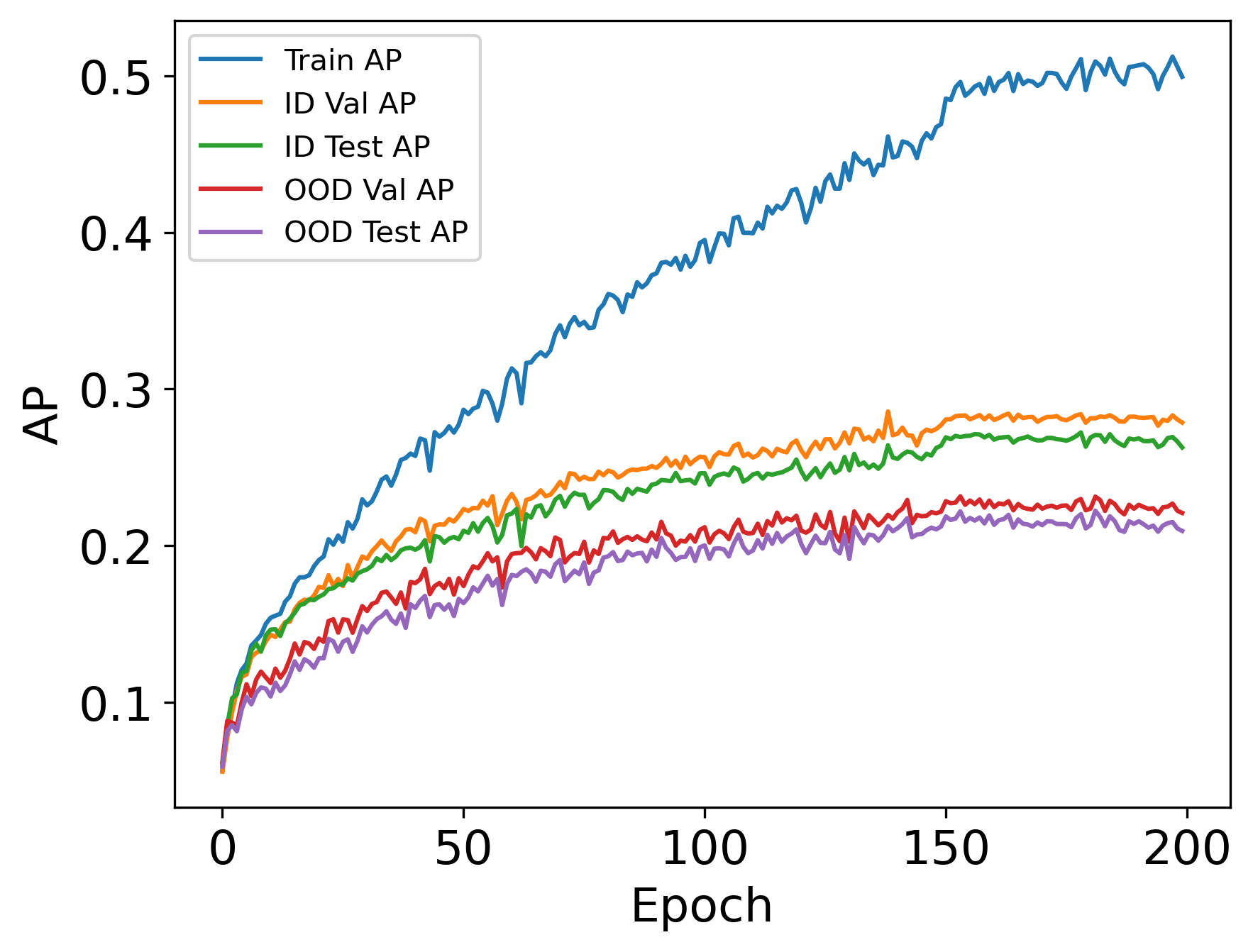}}
    \caption{concept shift + scaffold domain}
\end{subfigure}
\hfill
\begin{subfigure}[t]{0.32\textwidth}
    \raisebox{-\height}{\includegraphics[width=\textwidth]{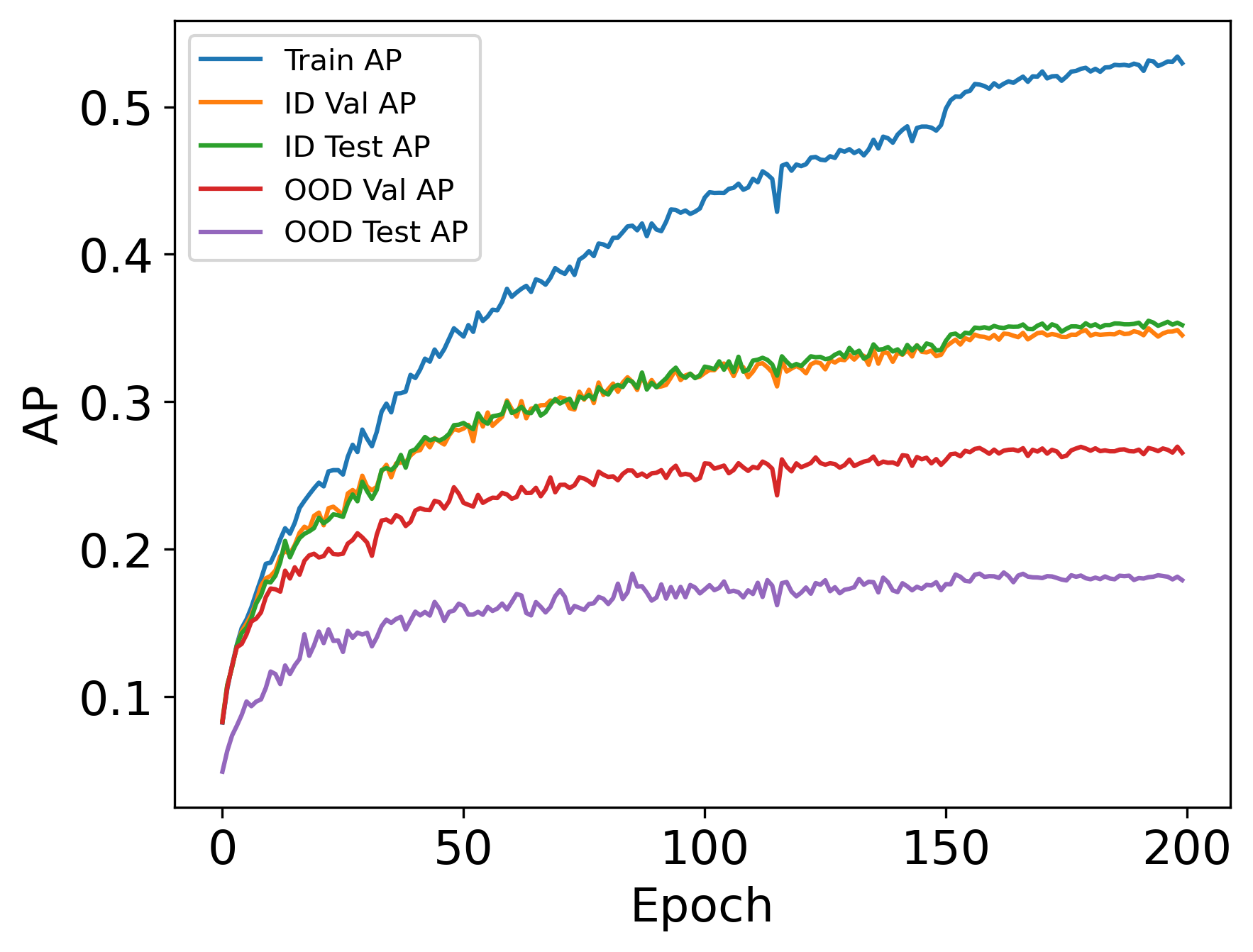}}
    \caption{covariate shift + size domain}
\end{subfigure}
\hspace*{\fill}
\begin{subfigure}[t]{0.32\textwidth}
    \raisebox{-\height}{\includegraphics[width=\textwidth]{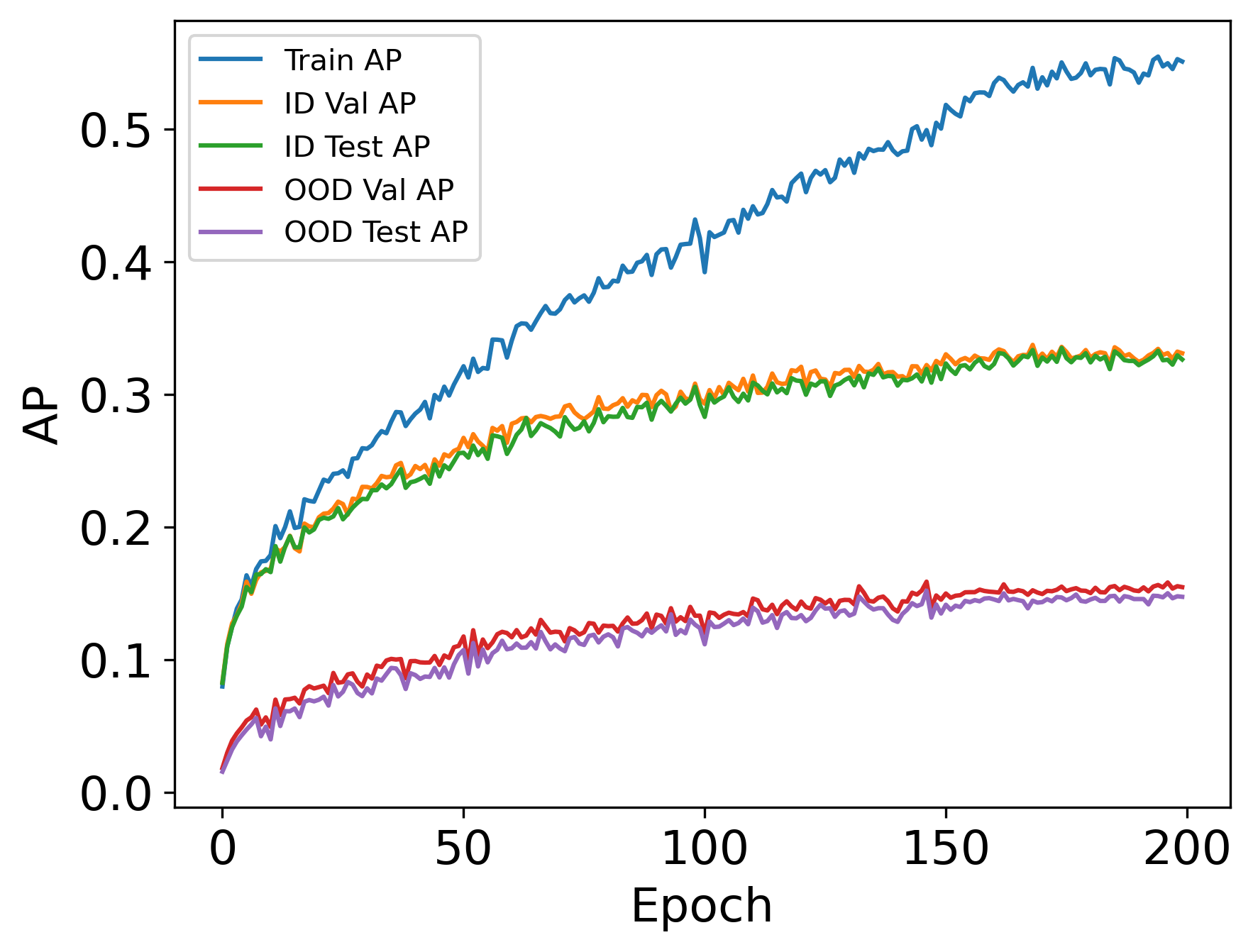}}
    \caption{concept shift + size domain}
\end{subfigure}
\hfill
\begin{subfigure}[t]{0.32\textwidth}
    \raisebox{-\height}{\includegraphics[width=\textwidth]{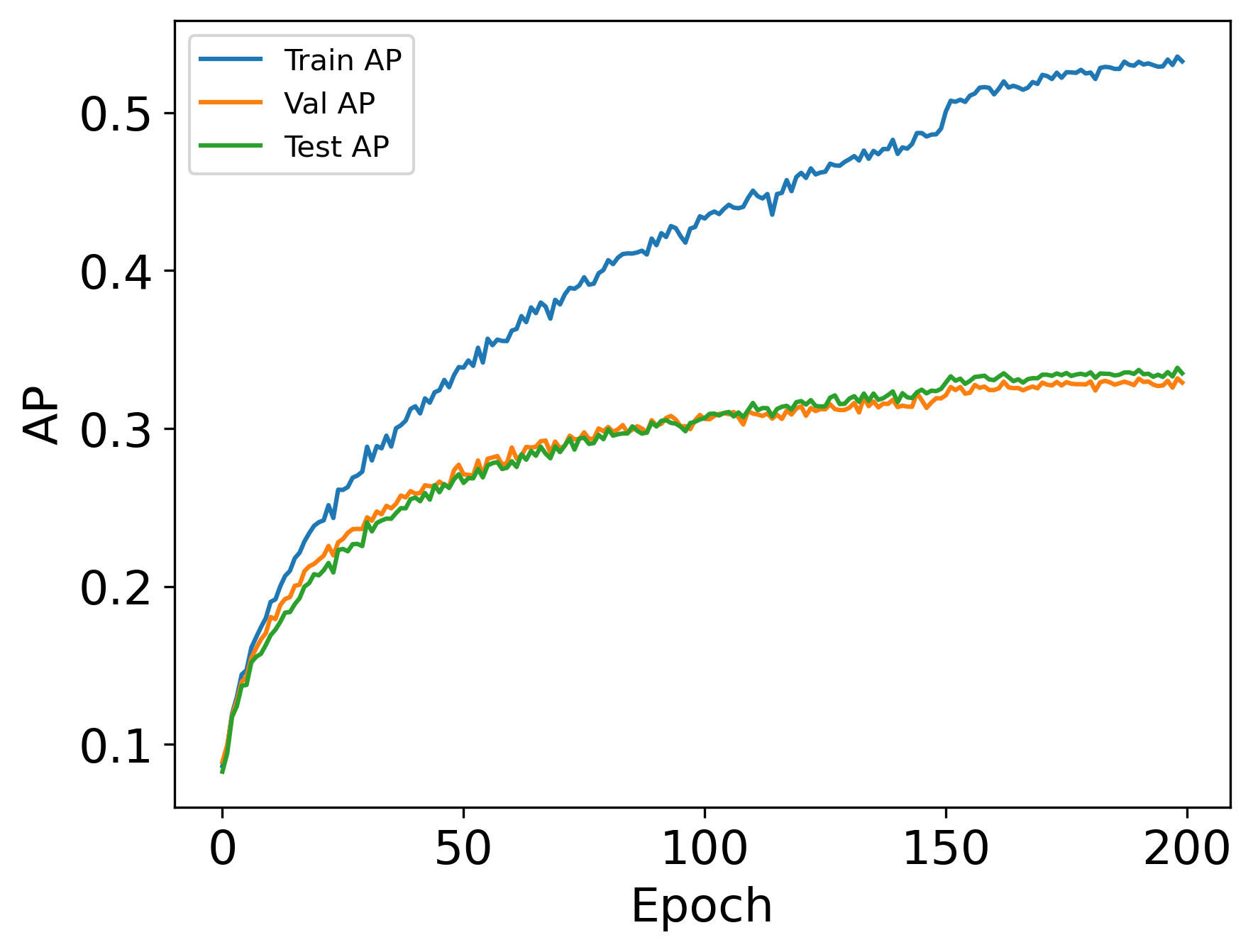}}
    \caption{no shift}
\end{subfigure}
\hspace*{\fill}
\caption{Metric score curves for ERM on GOOD-PCBA.}
    \label{fig:curve2}
\end{figure}

\begin{figure}[!htbp]
    \centering
\begin{subfigure}[t]{0.32\textwidth}
    \raisebox{-\height}{\includegraphics[width=\textwidth]{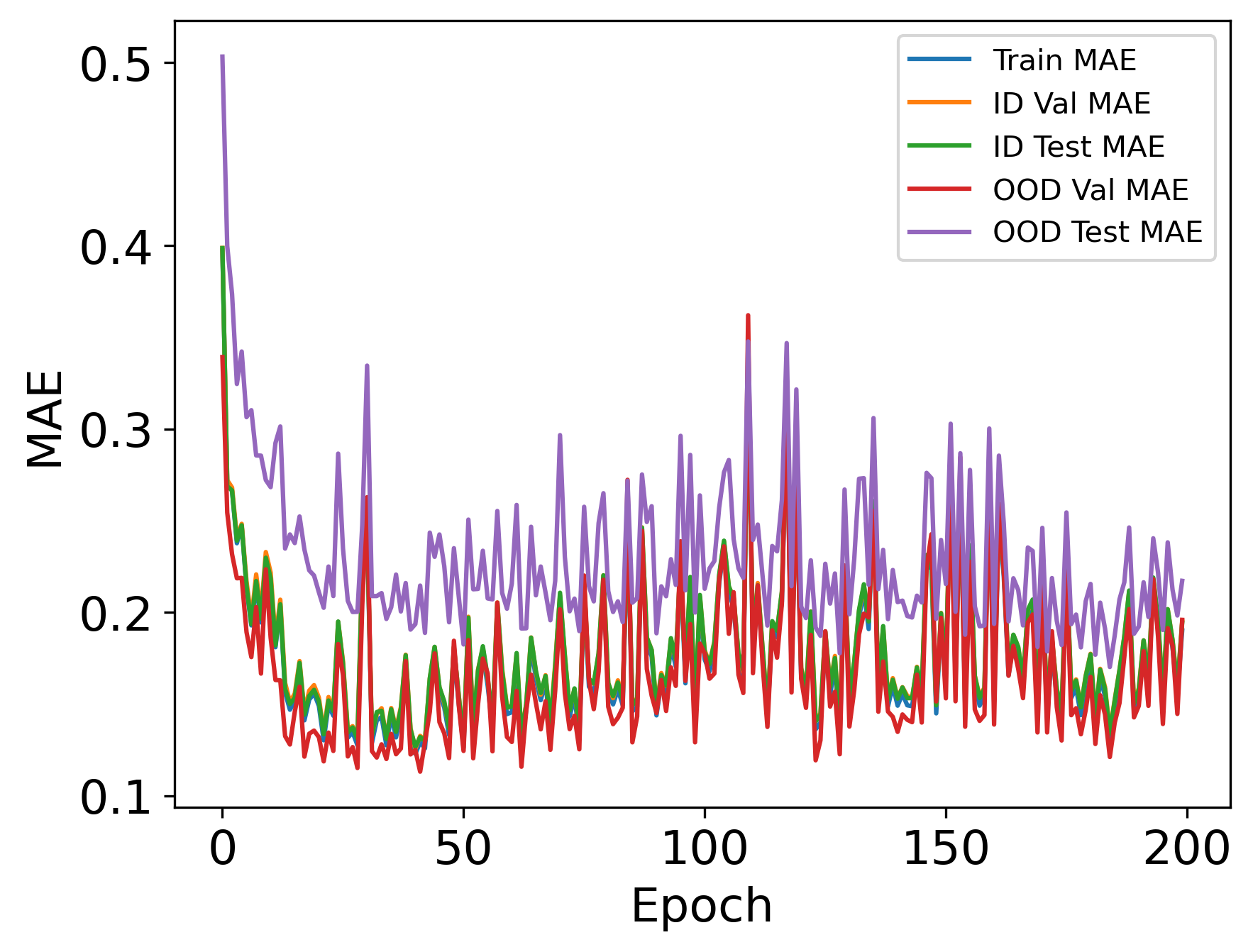}}
    \caption{covariate shift + scaffold domain}
\end{subfigure}
\hfill
\begin{subfigure}[t]{0.32\textwidth}
    \raisebox{-\height}{\includegraphics[width=\textwidth]{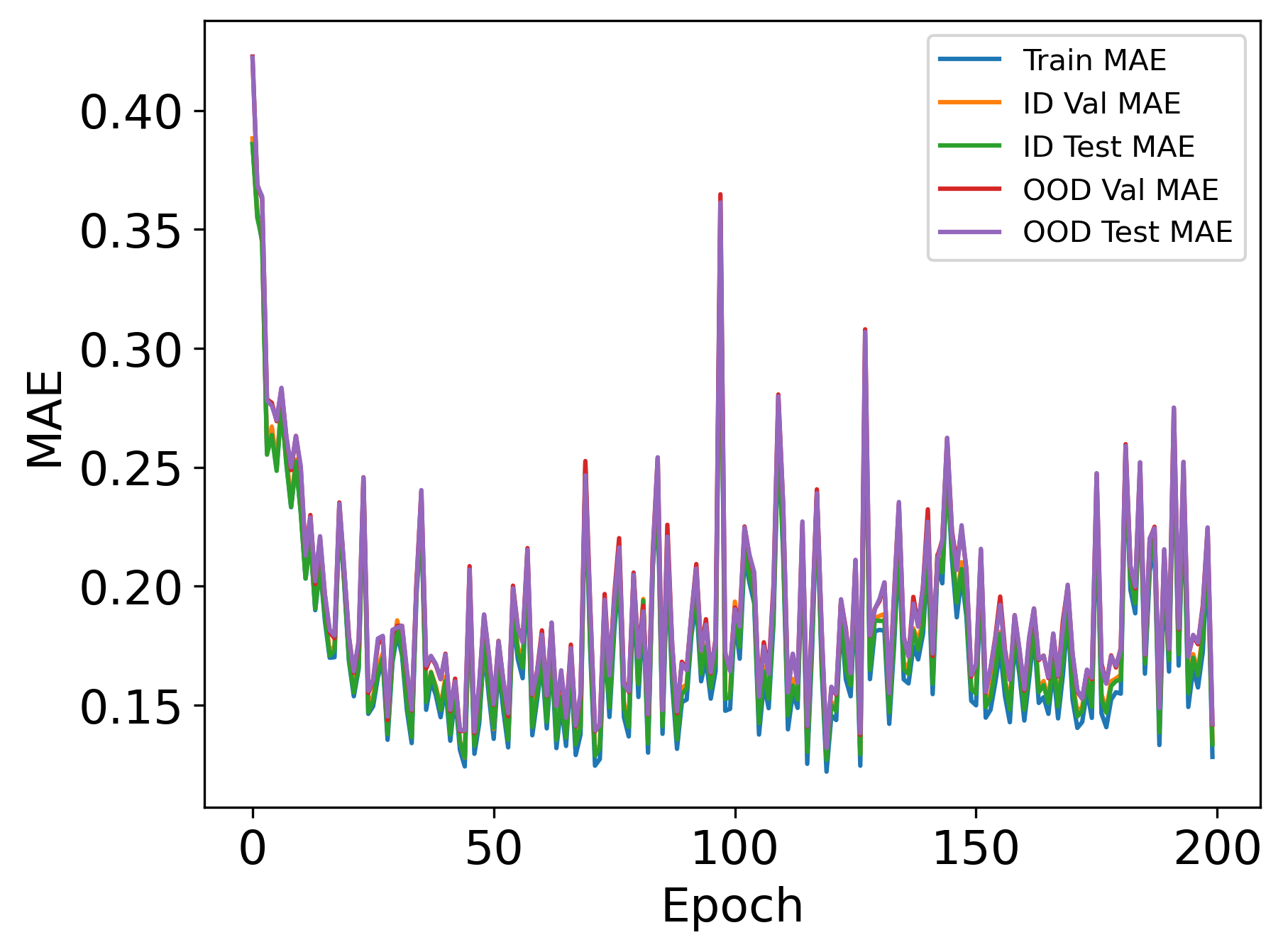}}
    \caption{concept shift + scaffold domain}
\end{subfigure}
\hfill
\begin{subfigure}[t]{0.32\textwidth}
    \raisebox{-\height}{\includegraphics[width=\textwidth]{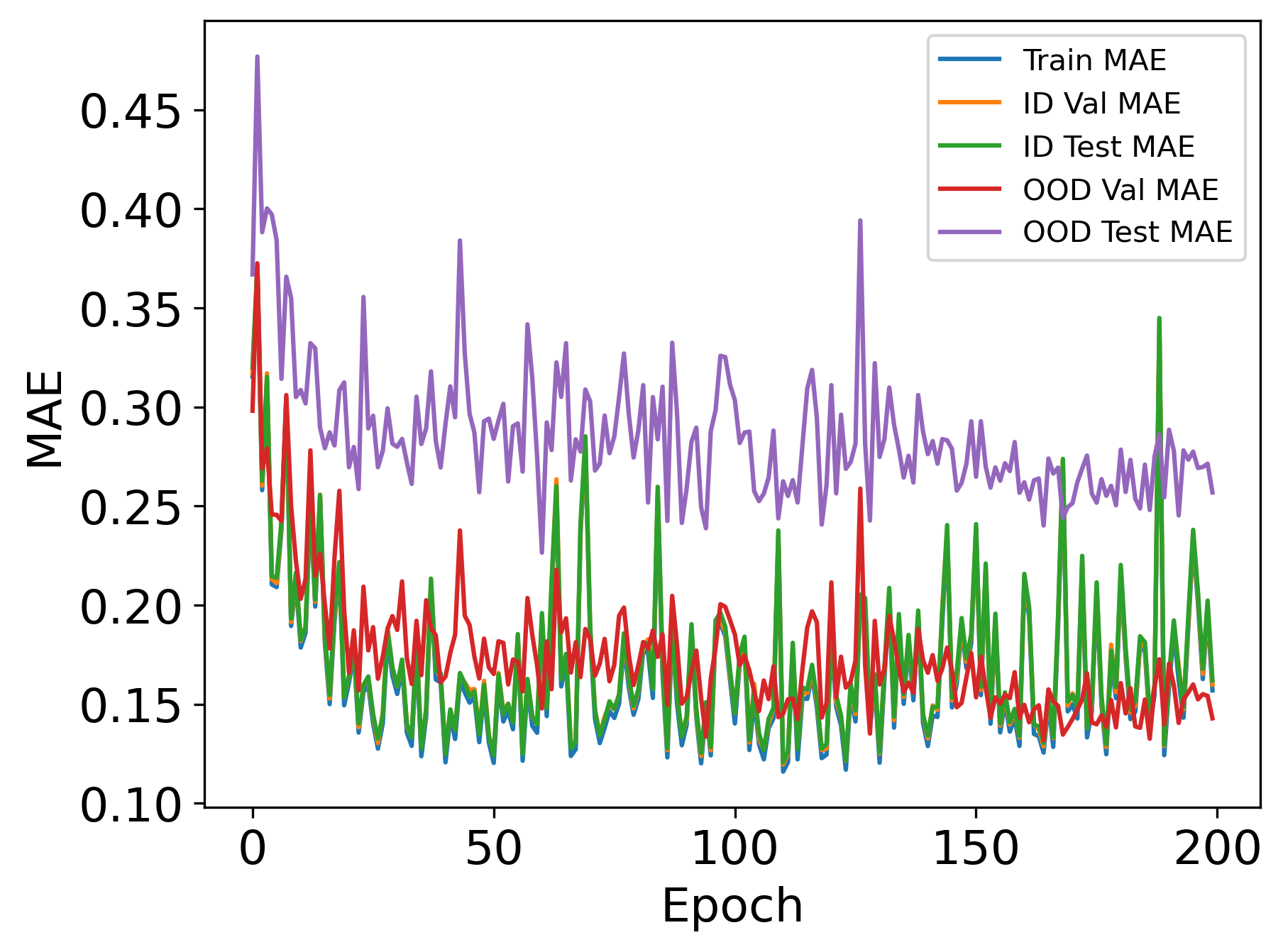}}
    \caption{covariate shift + size domain}
\end{subfigure}
\hspace*{\fill}
\begin{subfigure}[t]{0.32\textwidth}
    \raisebox{-\height}{\includegraphics[width=\textwidth]{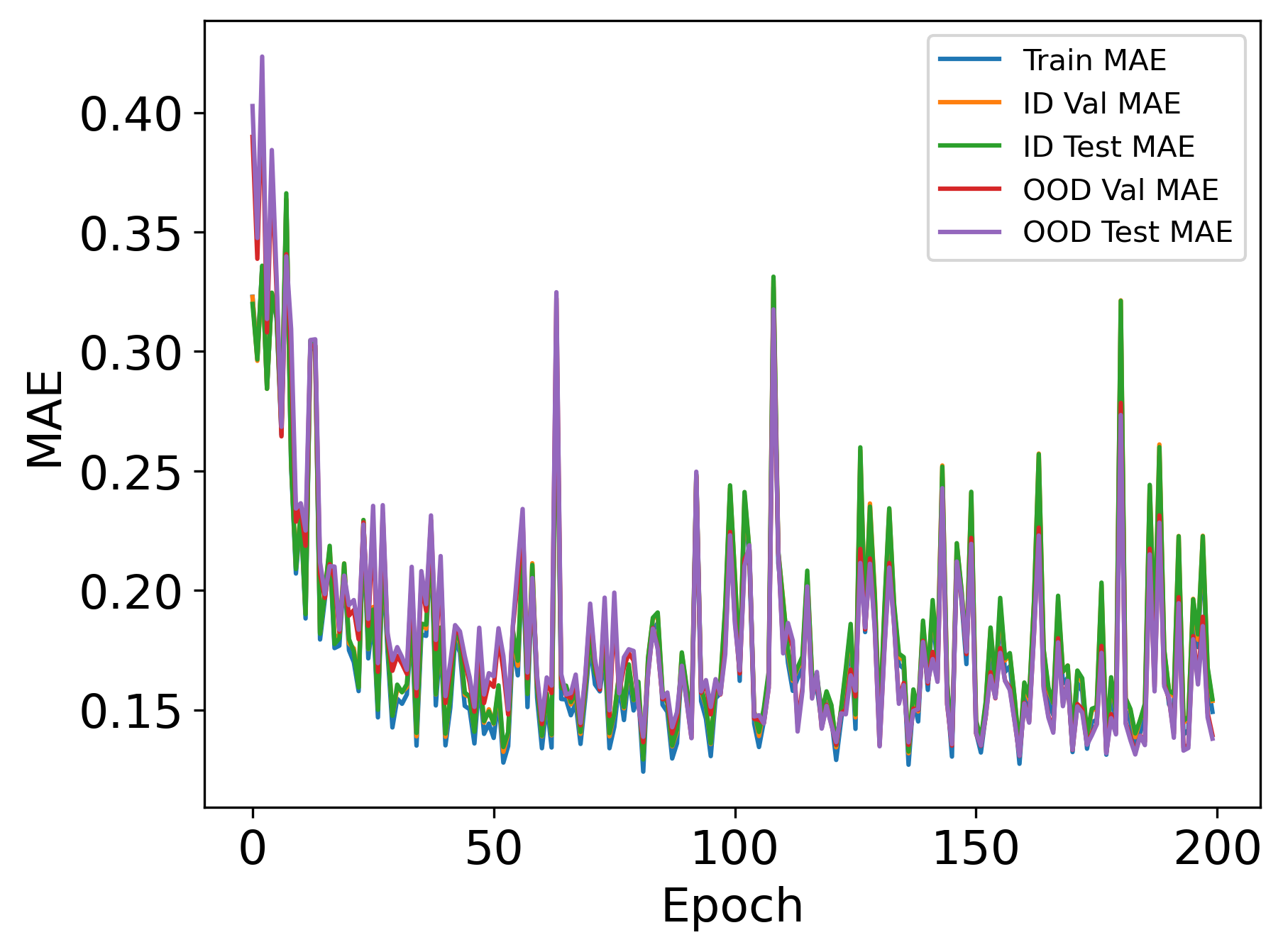}}
    \caption{concept shift + size domain}
\end{subfigure}
\hfill
\begin{subfigure}[t]{0.32\textwidth}
    \raisebox{-\height}{\includegraphics[width=\textwidth]{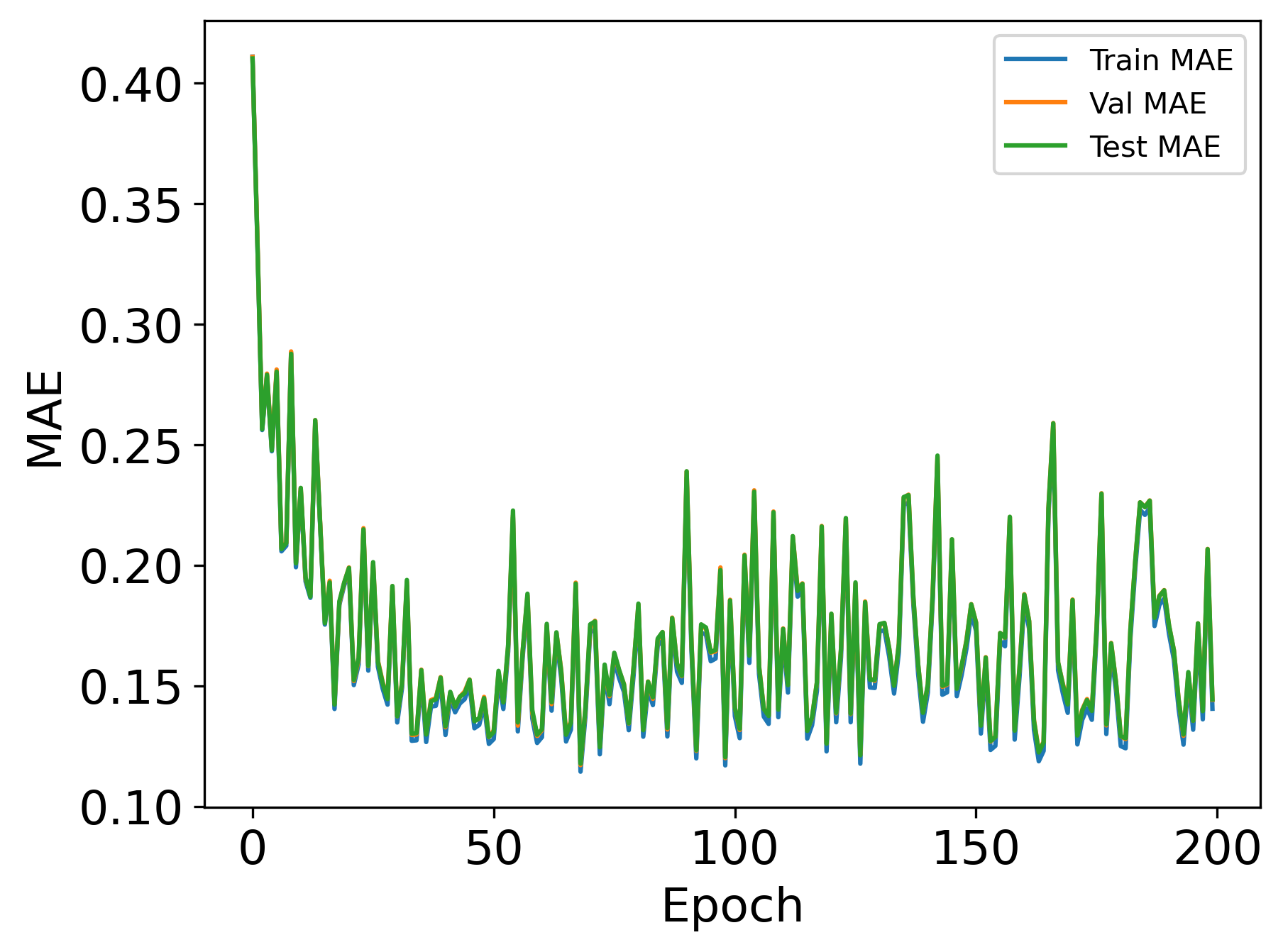}}
    \caption{no shift}
\end{subfigure}
\hspace*{\fill}
\caption{Metric score curves for ERM on GOOD-ZINC.}
    \label{fig:curve3}
\end{figure}

\begin{figure}[!htbp]
    \centering
\begin{subfigure}[t]{0.32\textwidth}
    \raisebox{-\height}{\includegraphics[width=\textwidth]{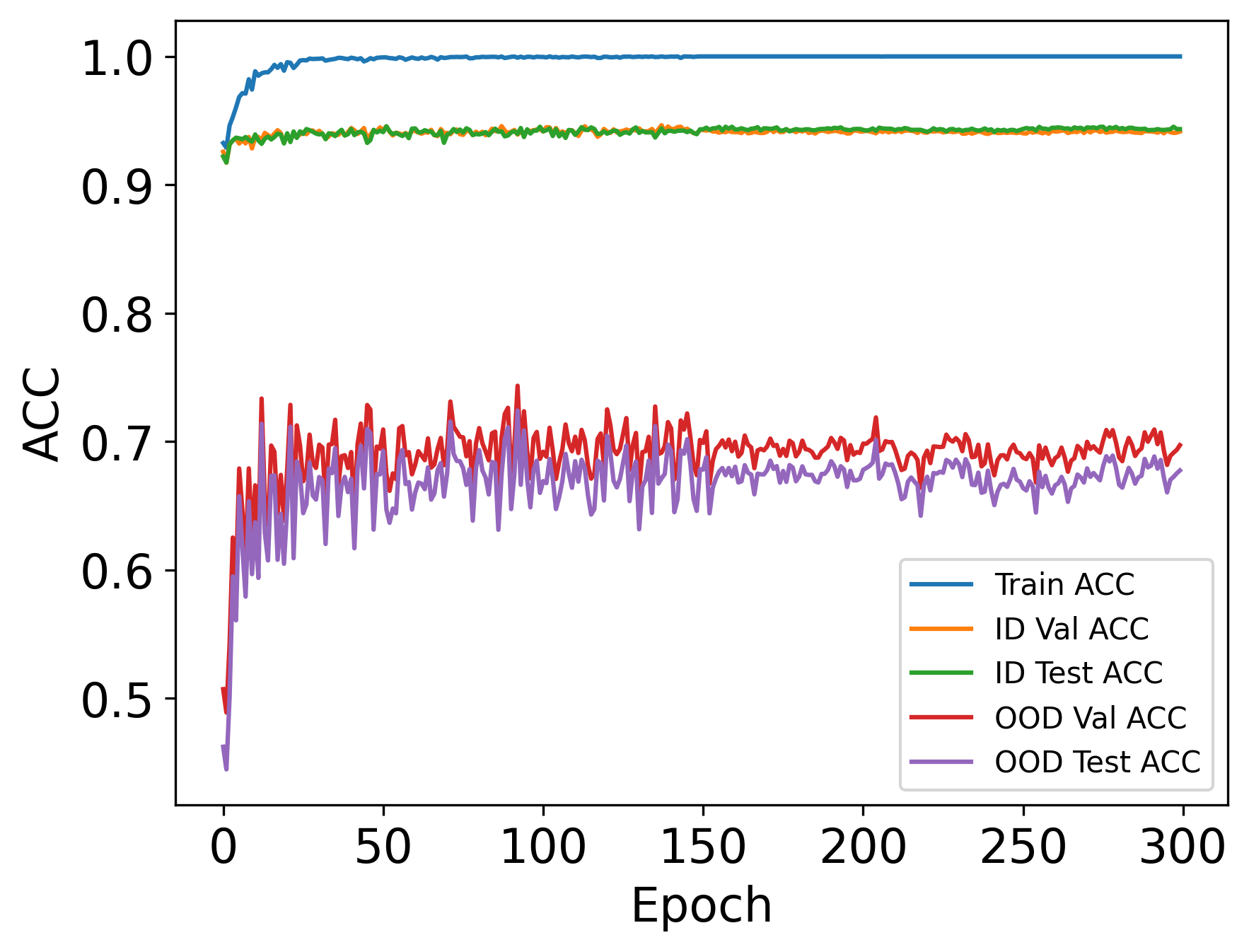}}
    \caption{concept shift + length domain}
\end{subfigure}
\hfill
\begin{subfigure}[t]{0.32\textwidth}
    \raisebox{-\height}{\includegraphics[width=\textwidth]{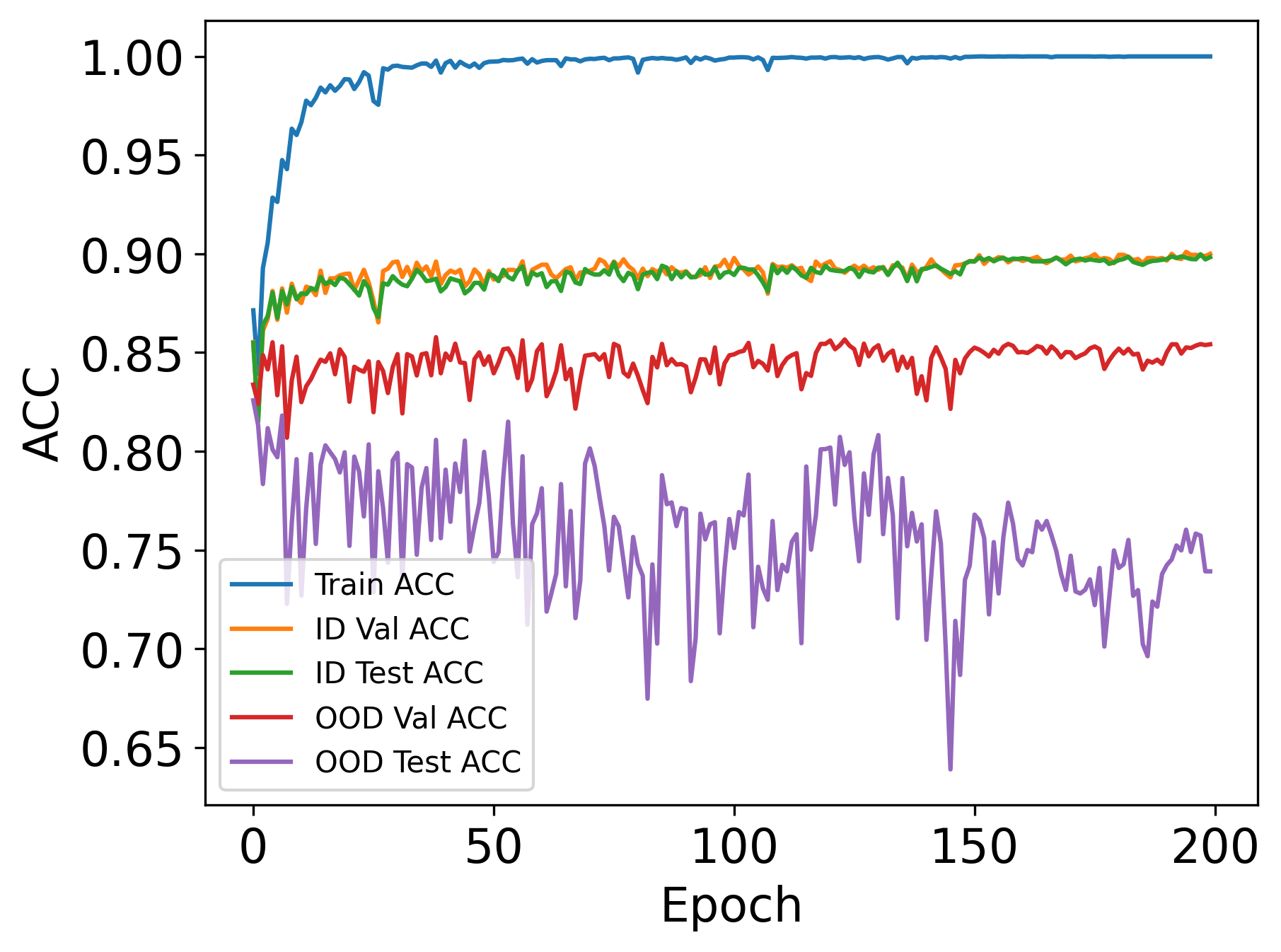}}
    \caption{covariate shift + length domain}
\end{subfigure}
\hfill
\begin{subfigure}[t]{0.32\textwidth}
    \raisebox{-\height}{\includegraphics[width=\textwidth]{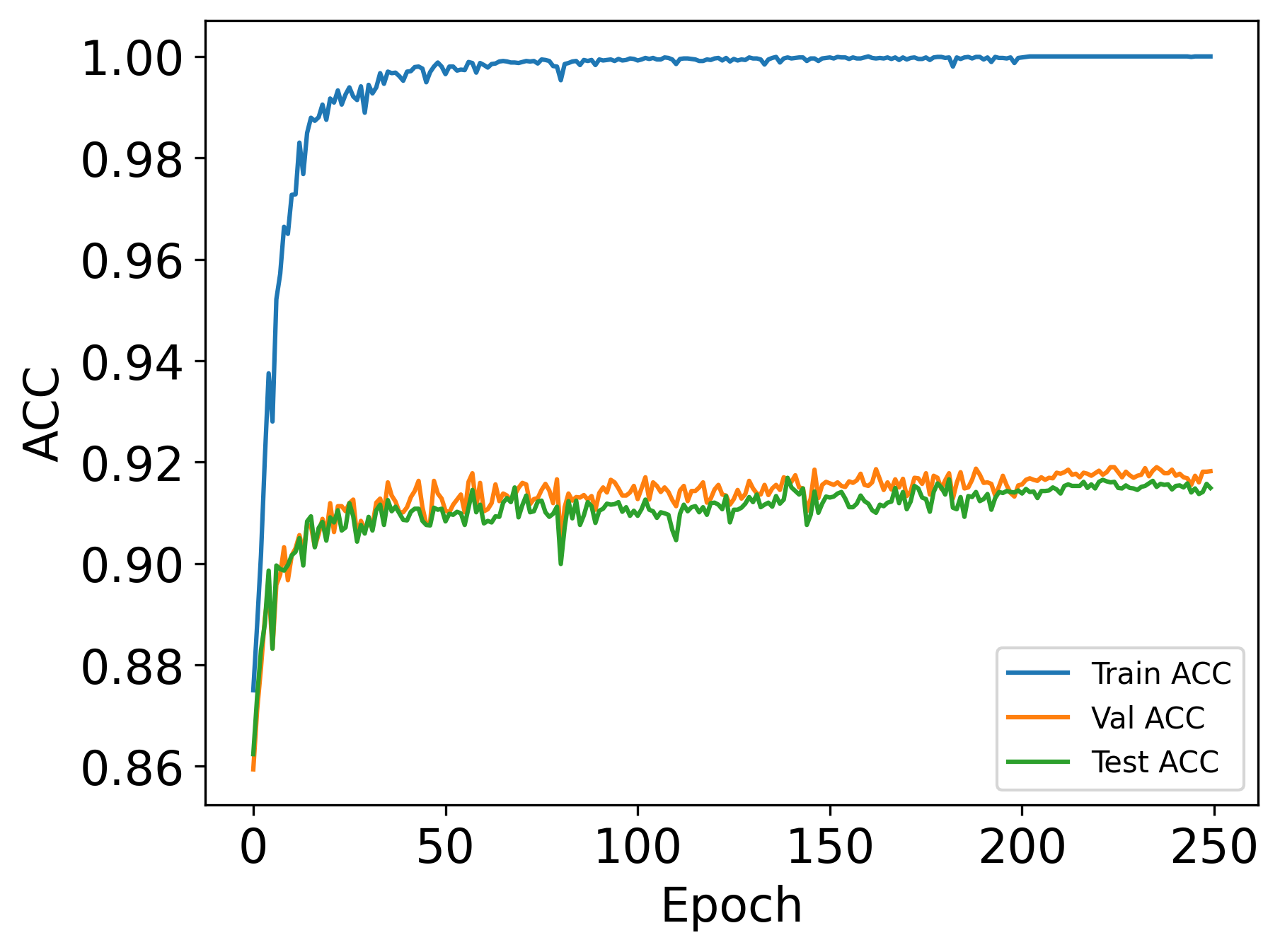}}
    \caption{no shift + length domain}
\end{subfigure}
    \caption{Metric score curves for ERM on GOOD-SST2.}
    \label{fig:curve35}
\end{figure}

\begin{figure}[!htbp]
    \centering
\begin{subfigure}[t]{0.32\textwidth}
    \raisebox{-\height}{\includegraphics[width=\textwidth]{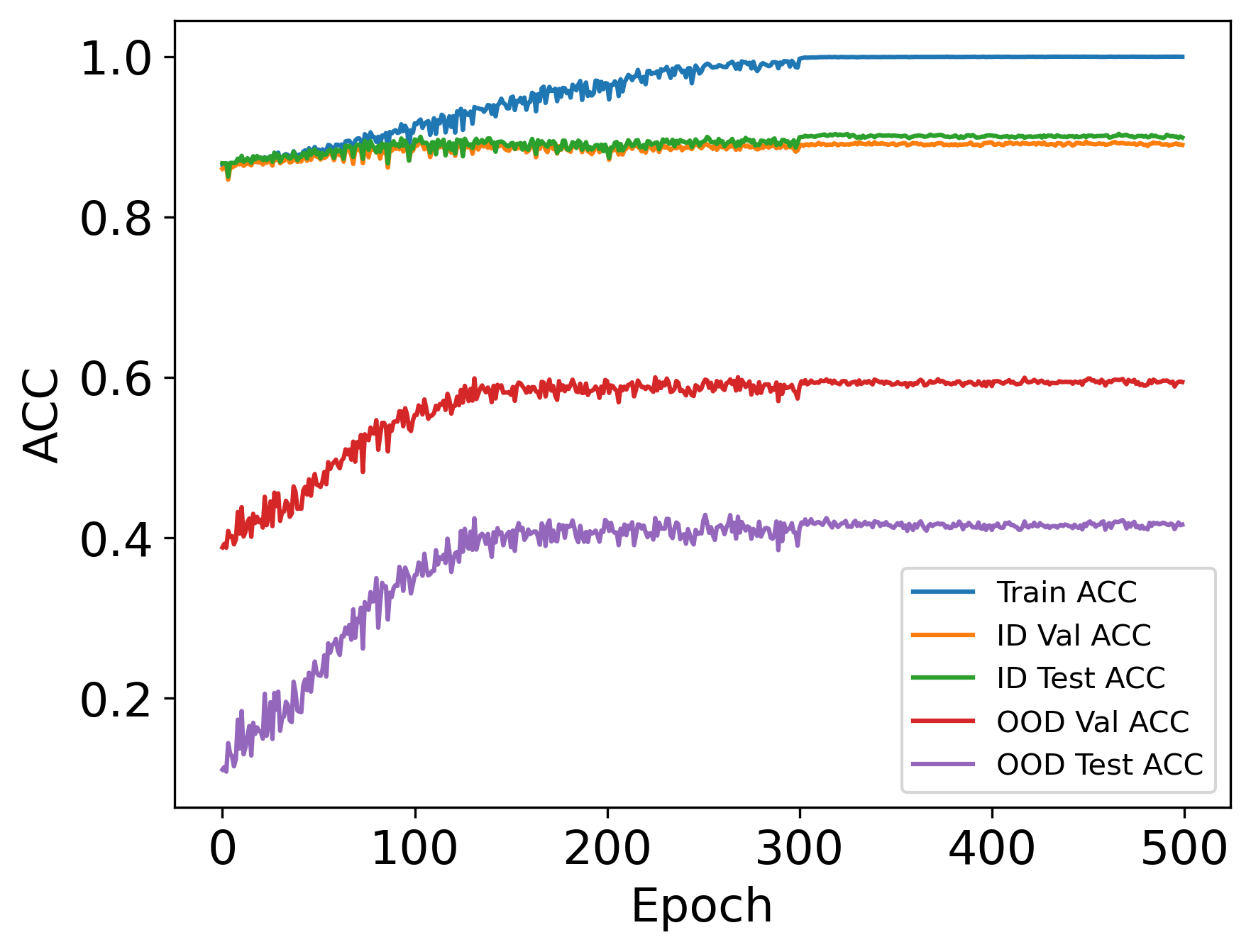}}
    \caption{concept shift + color domain}
\end{subfigure}
\hfill
\begin{subfigure}[t]{0.32\textwidth}
    \raisebox{-\height}{\includegraphics[width=\textwidth]{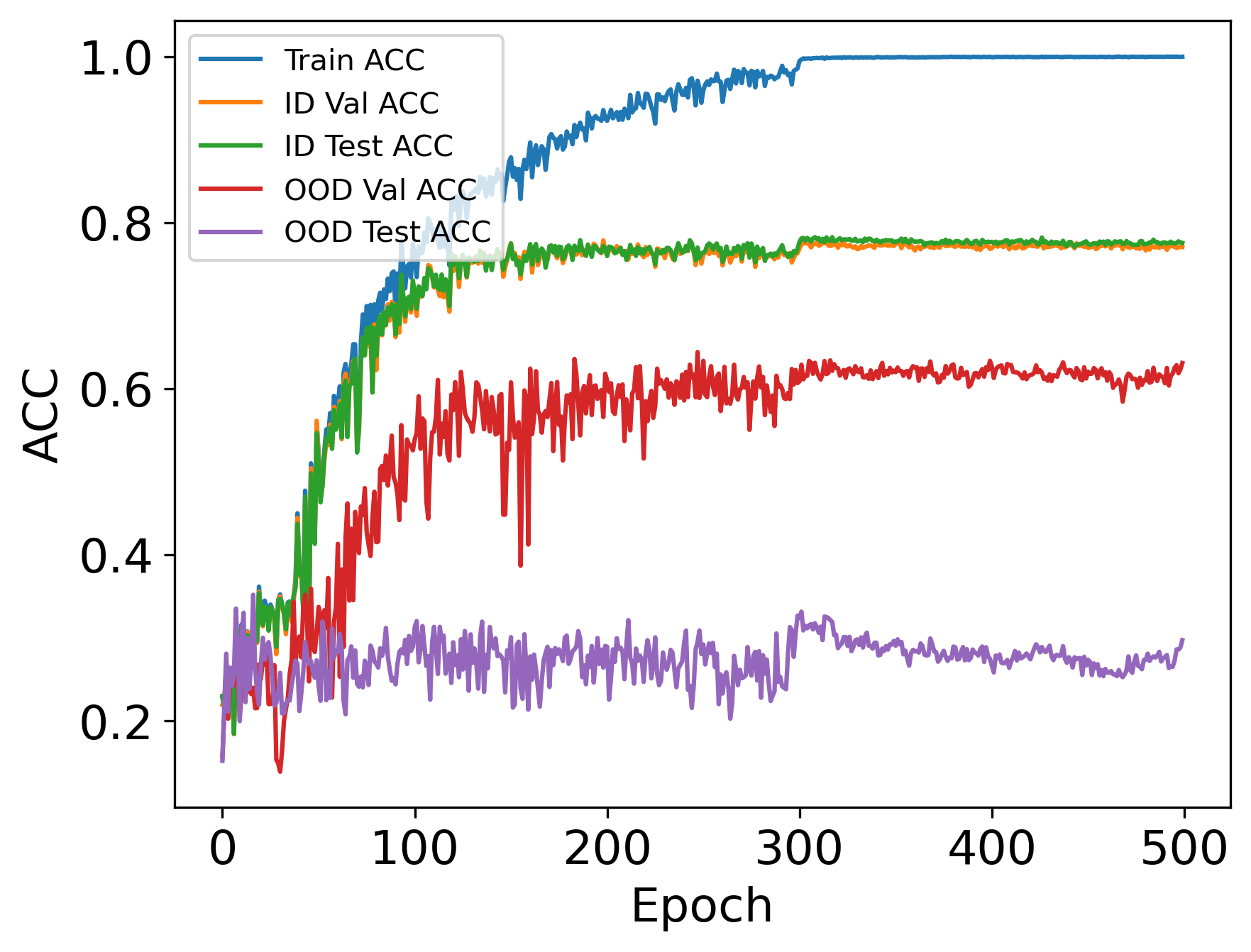}}
    \caption{covariate shift + color domain}
\end{subfigure}
\hfill
\begin{subfigure}[t]{0.32\textwidth}
    \raisebox{-\height}{\includegraphics[width=\textwidth]{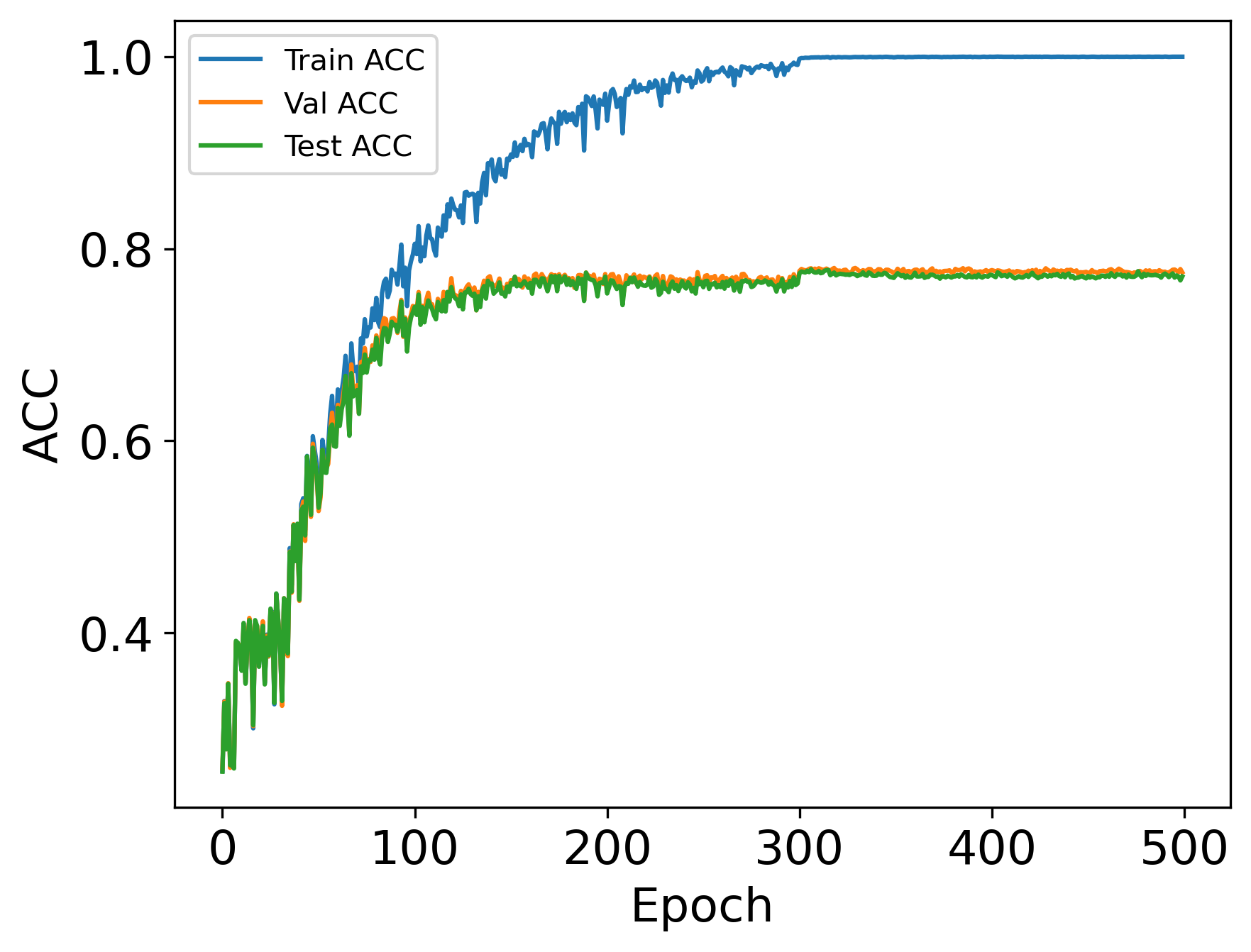}}
    \caption{no shift + color domain}
\end{subfigure}
    \caption{Metric score curves for ERM on GOOD-CMNIST.}
    \label{fig:curve4}
\end{figure}

\begin{figure}[!htbp]
    \centering
\begin{subfigure}[t]{0.32\textwidth}
    \raisebox{-\height}{\includegraphics[width=\textwidth]{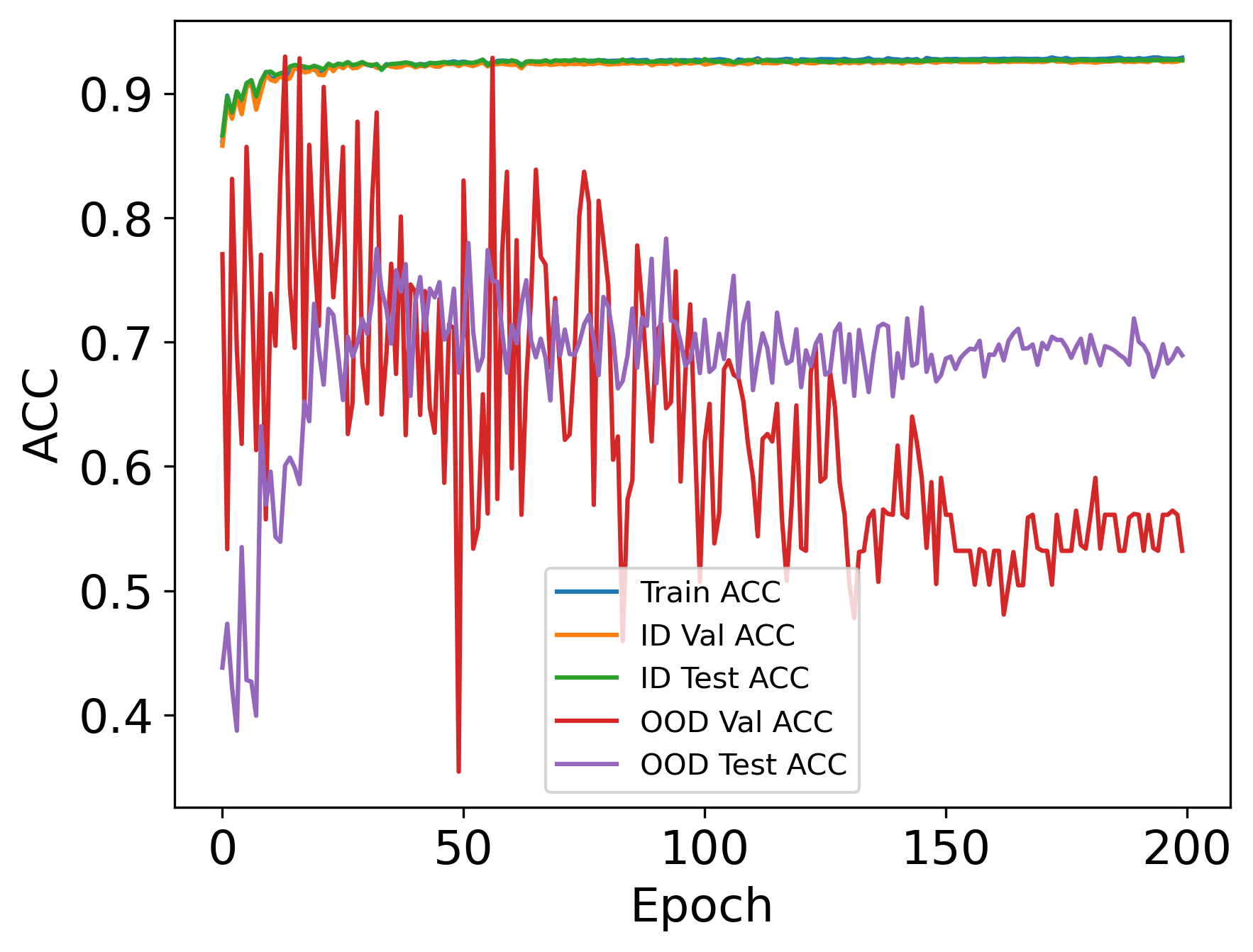}}
    \caption{covariate shift + base domain}
\end{subfigure}
\hfill
\begin{subfigure}[t]{0.32\textwidth}
    \raisebox{-\height}{\includegraphics[width=\textwidth]{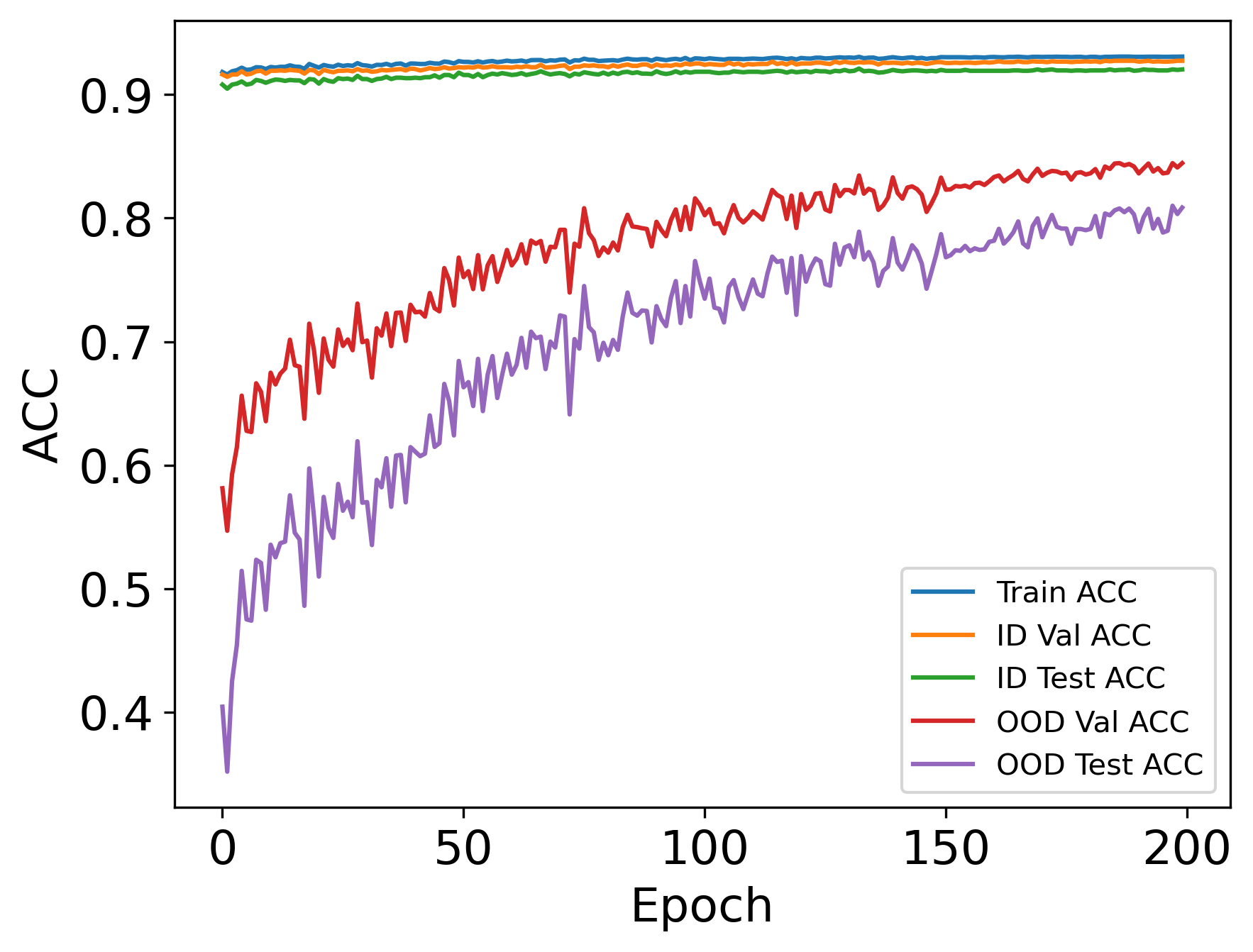}}
    \caption{concept shift + base domain}
\end{subfigure}
\hfill
\begin{subfigure}[t]{0.32\textwidth}
    \raisebox{-\height}{\includegraphics[width=\textwidth]{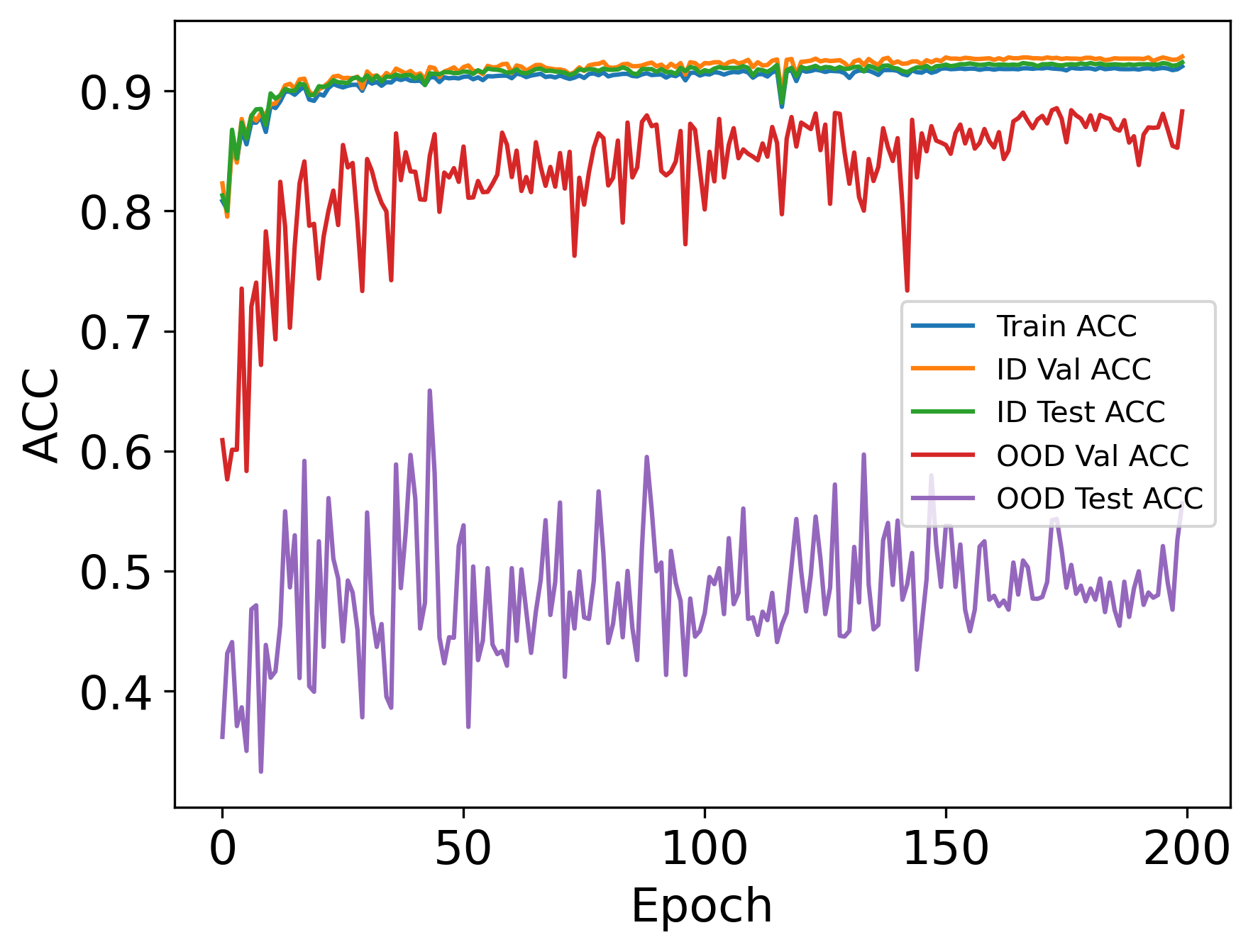}}
    \caption{covariate shift + size domain}
\end{subfigure}
\hspace*{\fill}
\begin{subfigure}[t]{0.32\textwidth}
    \raisebox{-\height}{\includegraphics[width=\textwidth]{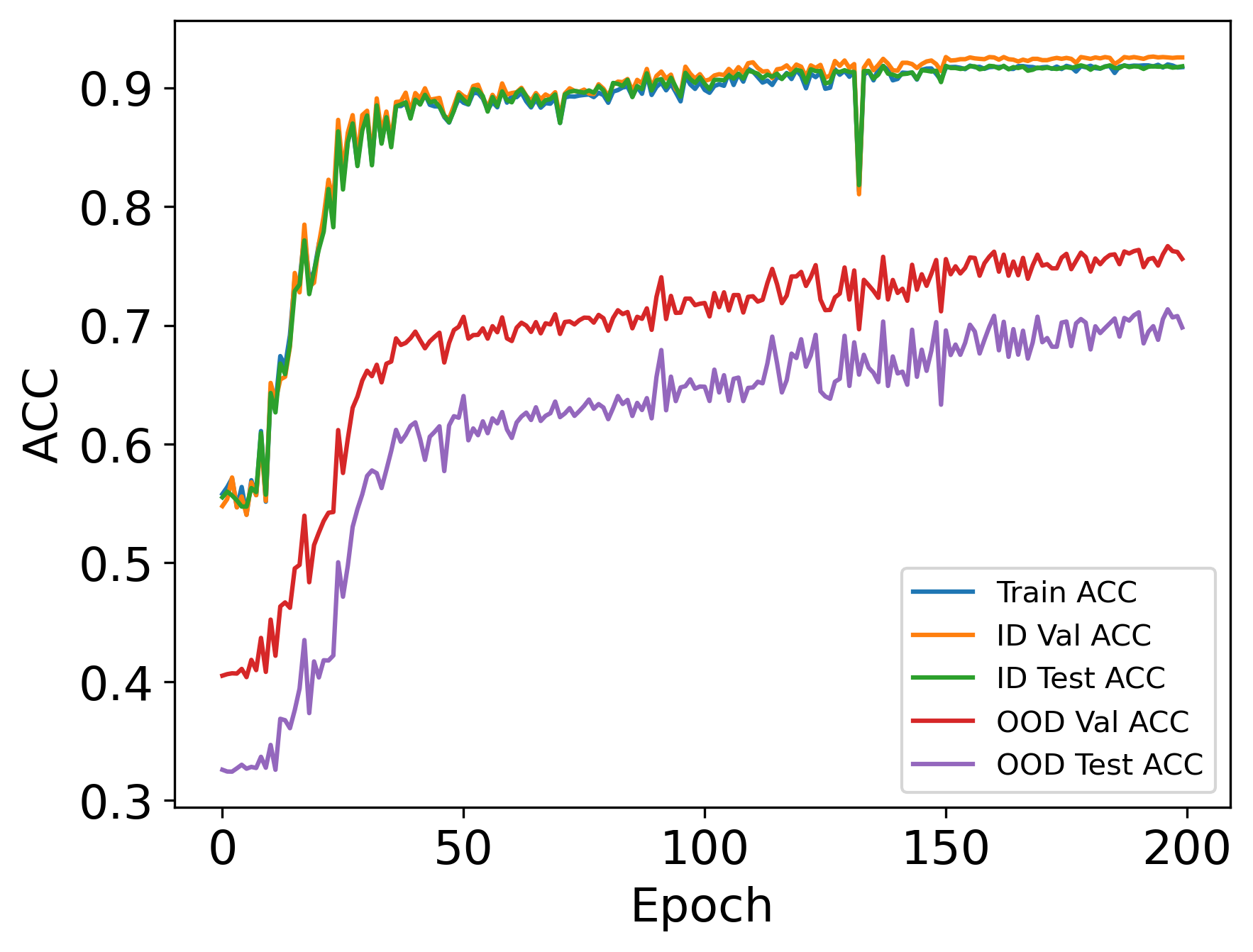}}
    \caption{concept shift + size domain}
\end{subfigure}
\hfill
\begin{subfigure}[t]{0.32\textwidth}
    \raisebox{-\height}{\includegraphics[width=\textwidth]{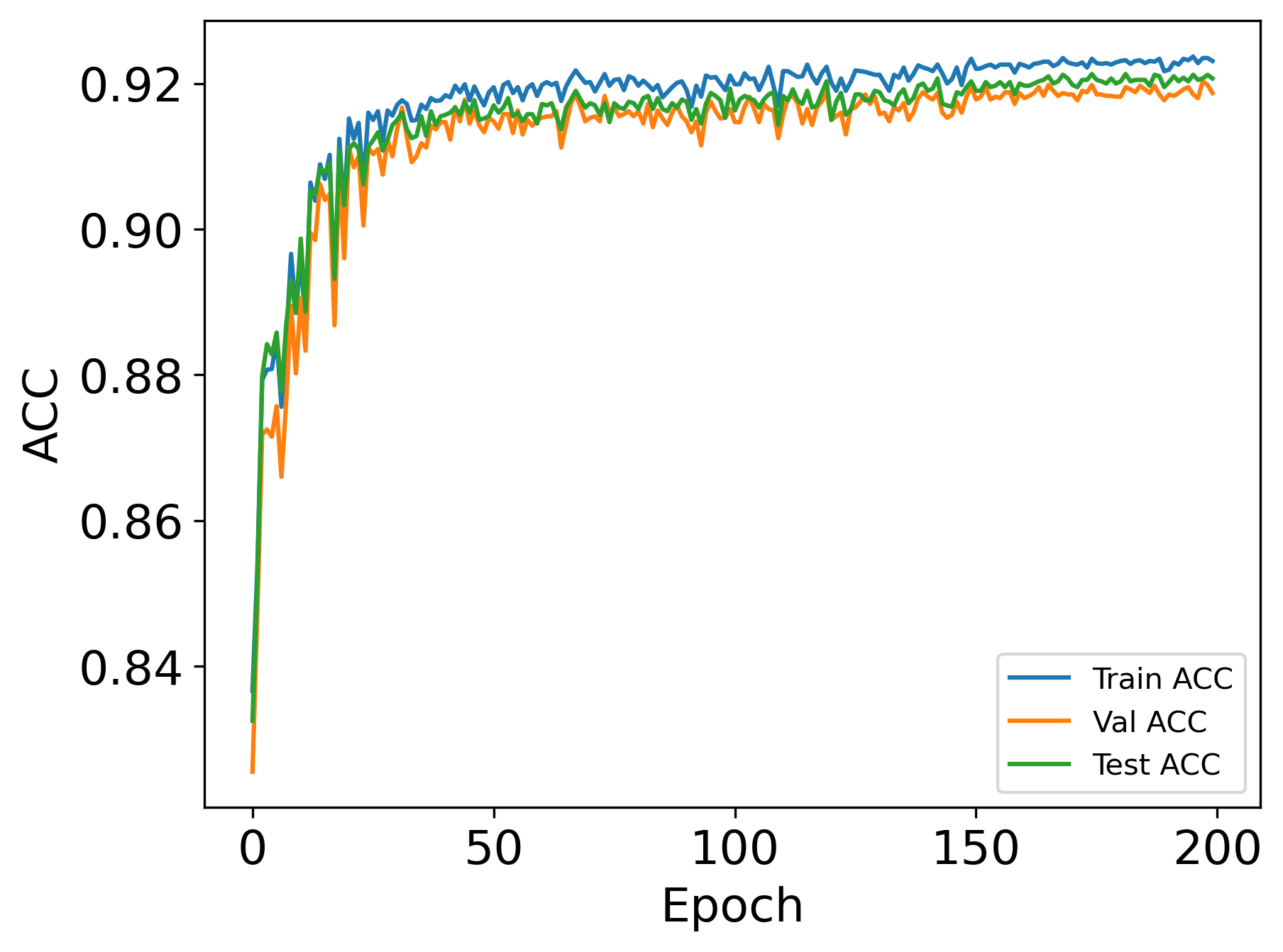}}
    \caption{no shift}
\end{subfigure}
\hspace*{\fill}
\caption{Metric score curves for ERM on GOOD-Motif.}
    \label{fig:curve5}
\end{figure}

\begin{figure}[!htbp]
    \centering
\begin{subfigure}[t]{0.32\textwidth}
    \raisebox{-\height}{\includegraphics[width=\textwidth]{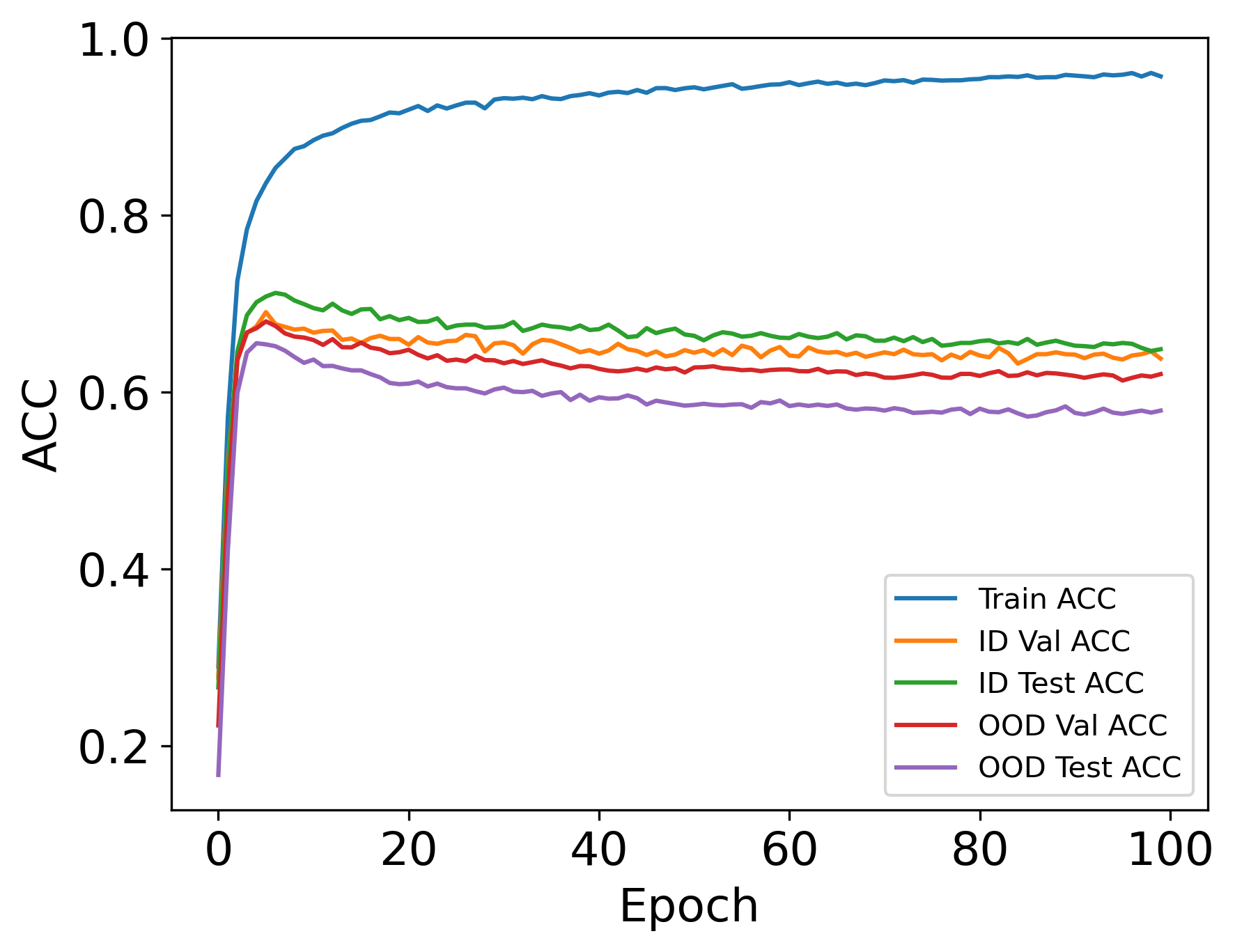}}
    \caption{covariate shift + word domain}
\end{subfigure}
\hfill
\begin{subfigure}[t]{0.32\textwidth}
    \raisebox{-\height}{\includegraphics[width=\textwidth]{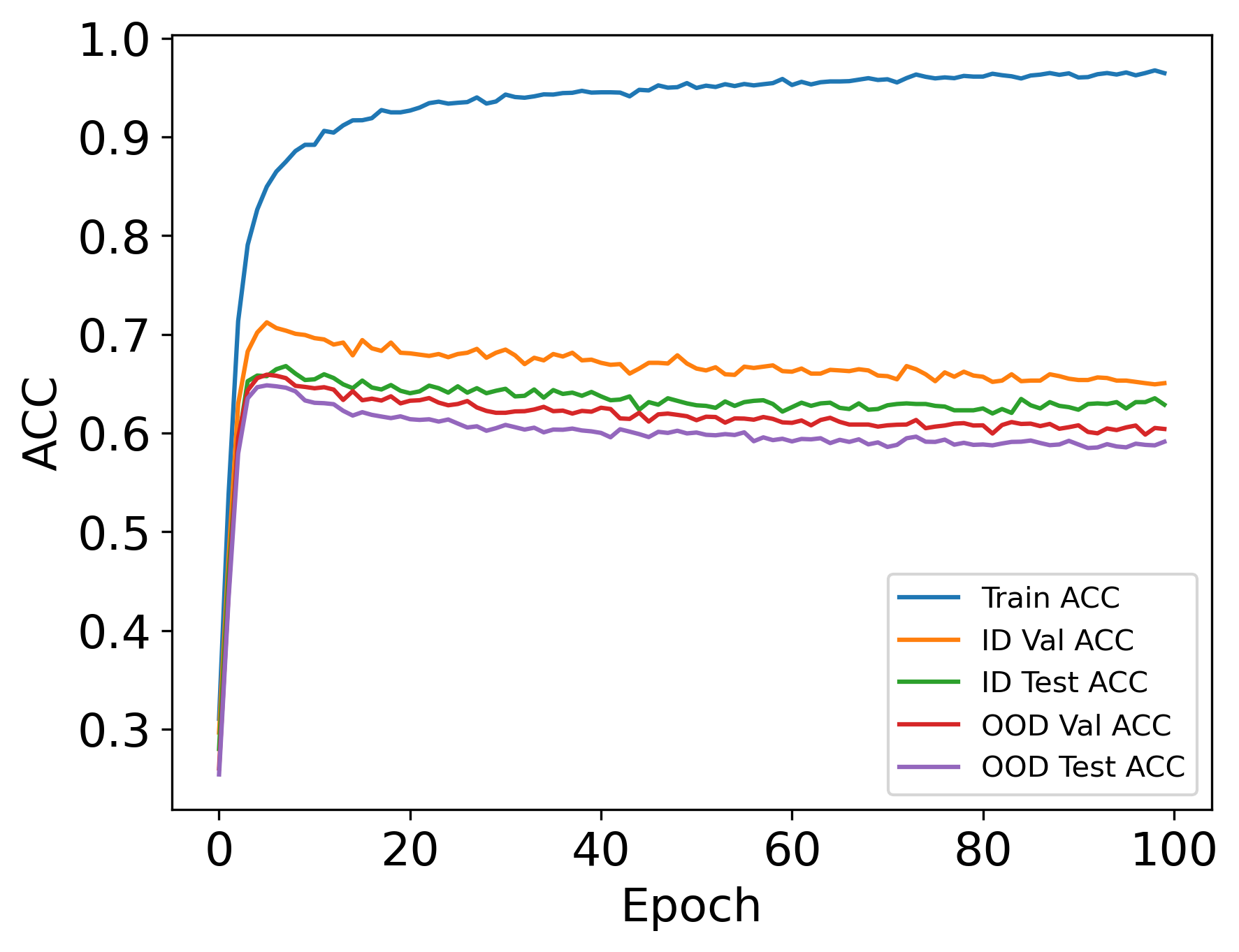}}
    \caption{concept shift + word domain}
\end{subfigure}
\hfill
\begin{subfigure}[t]{0.32\textwidth}
    \raisebox{-\height}{\includegraphics[width=\textwidth]{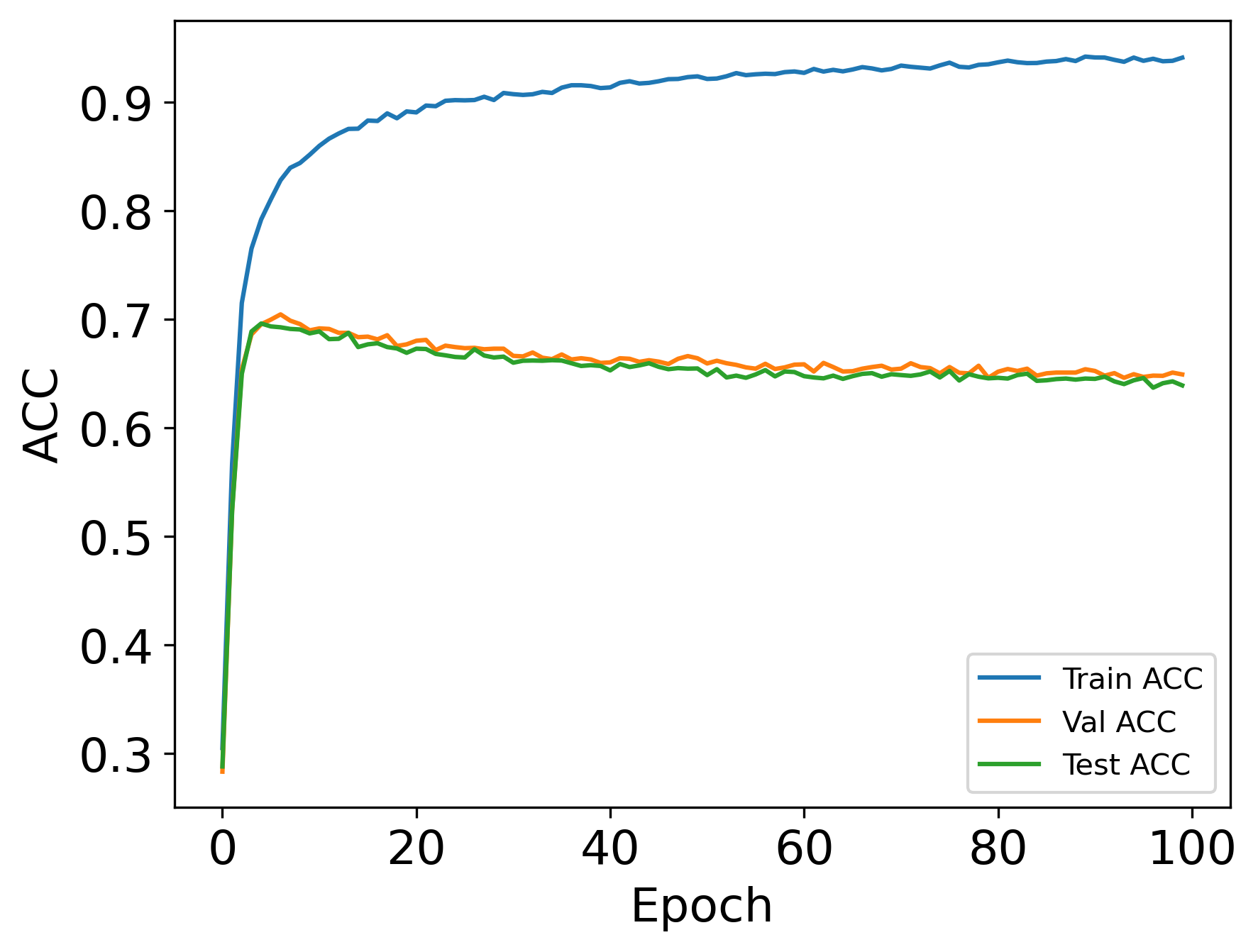}}
    \caption{no shift + word domain}
\end{subfigure}
\hfill
\begin{subfigure}[t]{0.32\textwidth}
    \raisebox{-\height}{\includegraphics[width=\textwidth]{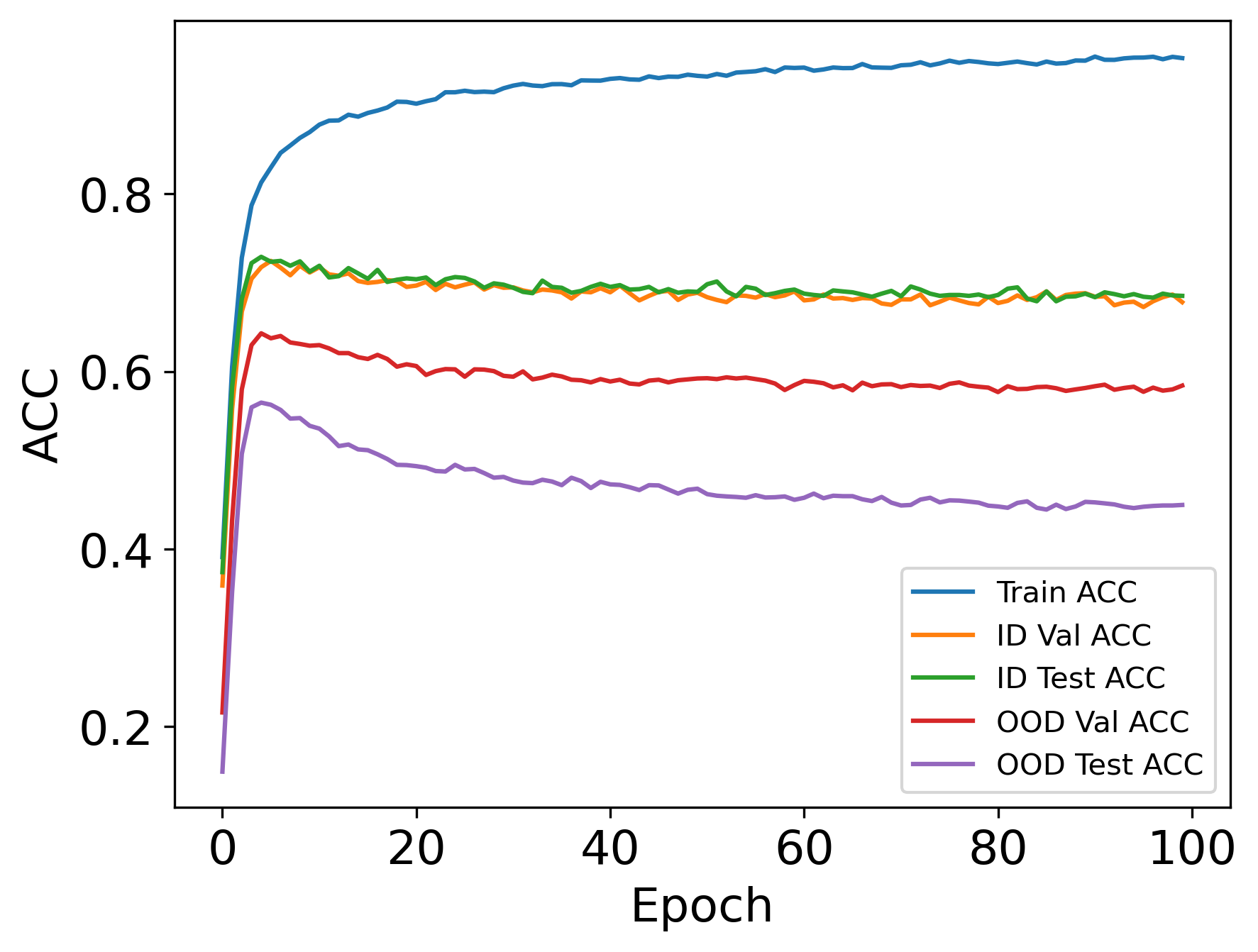}}
    \caption{covariate shift + degree domain}
\end{subfigure}
\hfill
\begin{subfigure}[t]{0.32\textwidth}
    \raisebox{-\height}{\includegraphics[width=\textwidth]{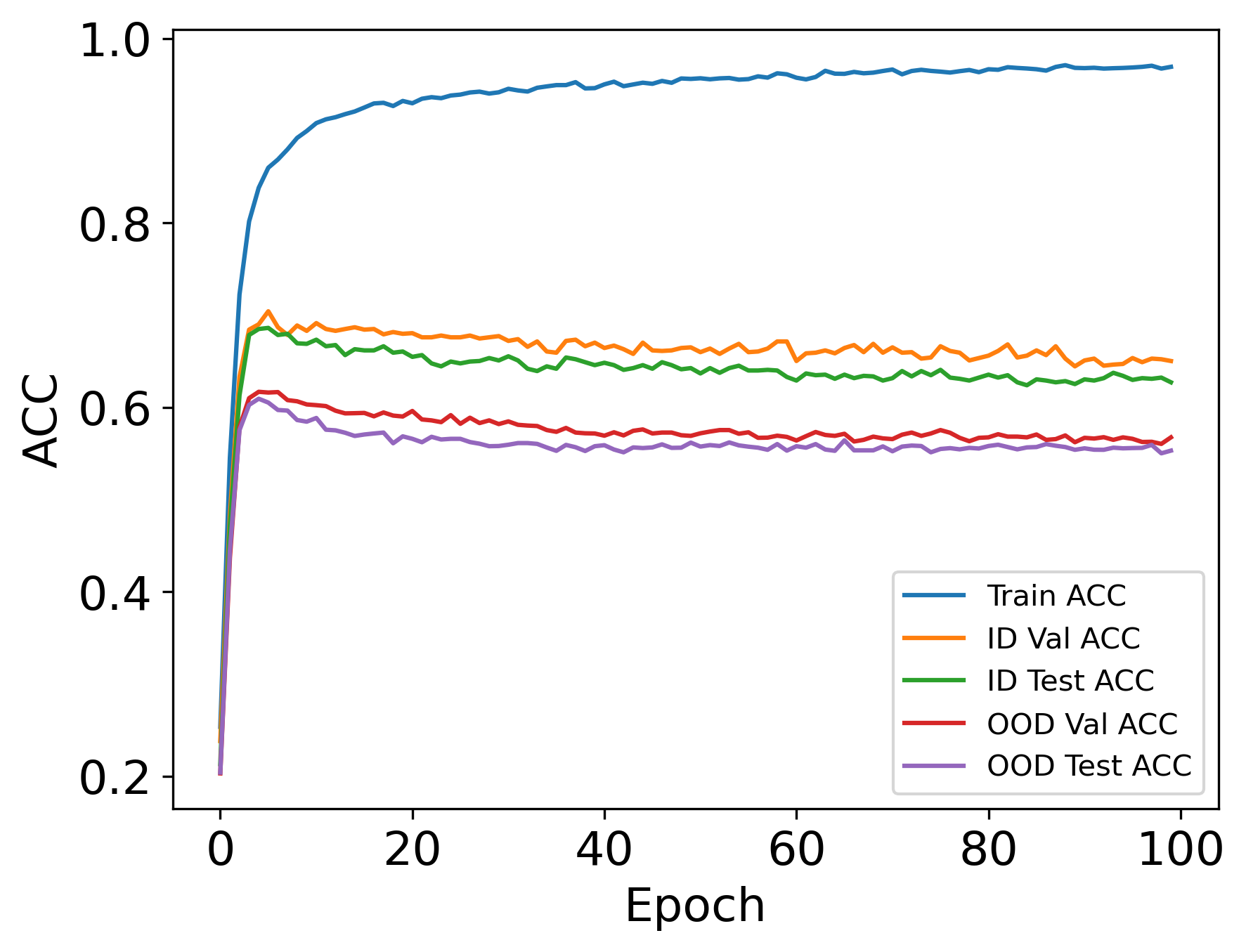}}
    \caption{concept shift + degree domain}
\end{subfigure}
\hfill
\begin{subfigure}[t]{0.32\textwidth}
    \raisebox{-\height}{\includegraphics[width=\textwidth]{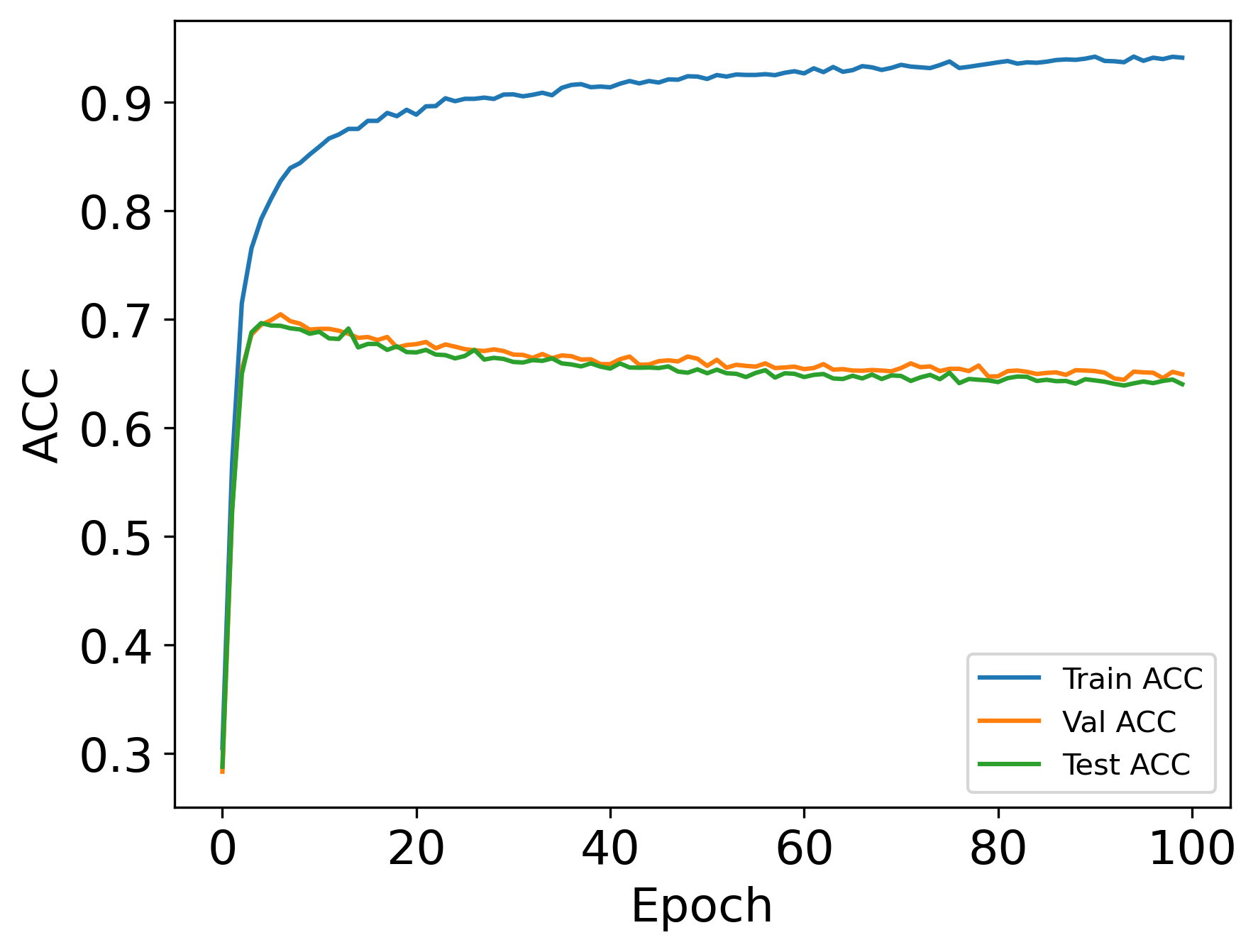}}
    \caption{no shift + degree domain}
\end{subfigure}
\caption{Metric score curves for ERM on GOOD-Cora.}
    \label{fig:curve6}
\end{figure}

\begin{figure}[!htbp]
    \centering
\begin{subfigure}[t]{0.32\textwidth}
    \raisebox{-\height}{\includegraphics[width=\textwidth]{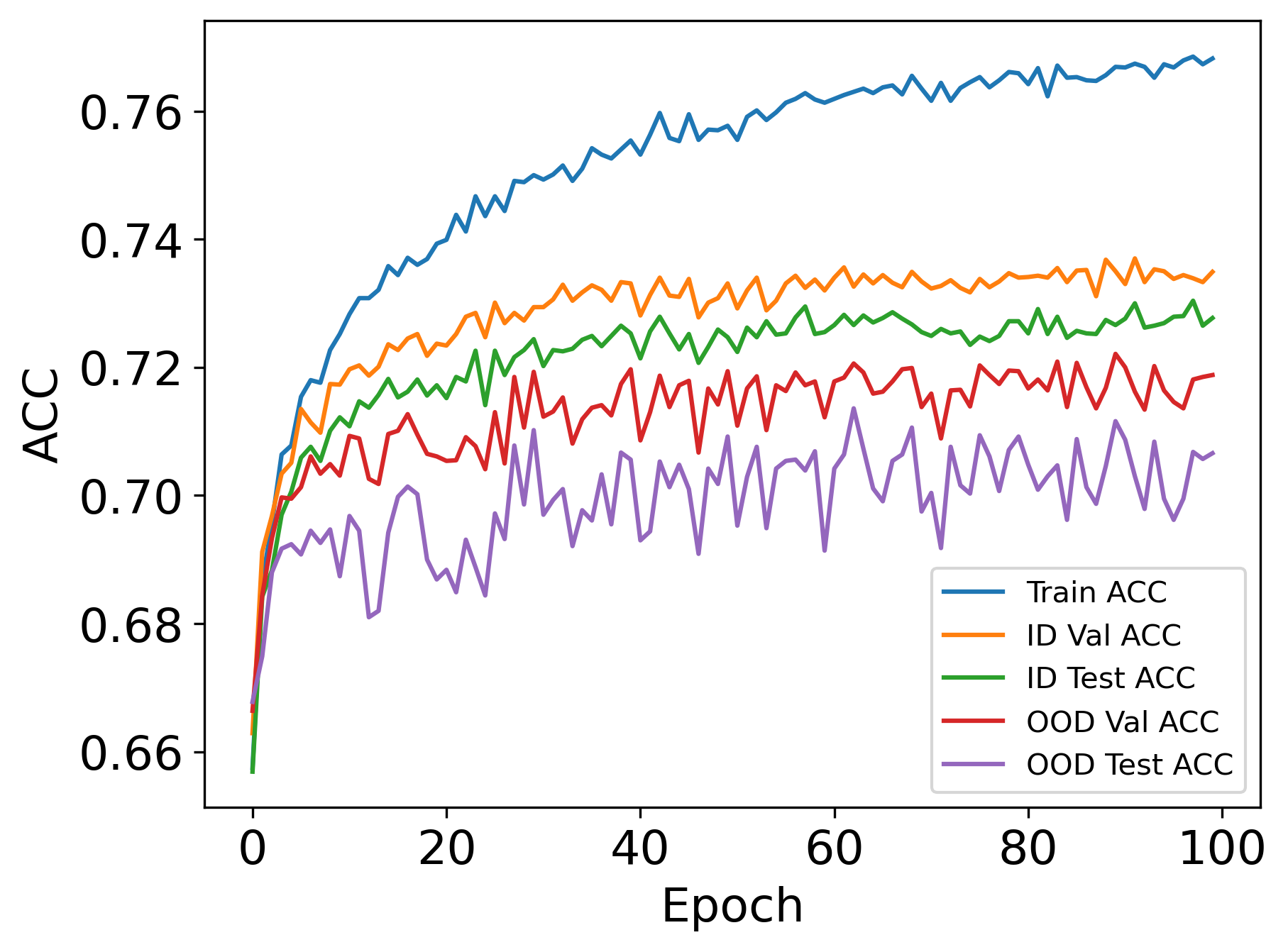}}
    \caption{covariate shift + time domain}
\end{subfigure}
\hfill
\begin{subfigure}[t]{0.32\textwidth}
    \raisebox{-\height}{\includegraphics[width=\textwidth]{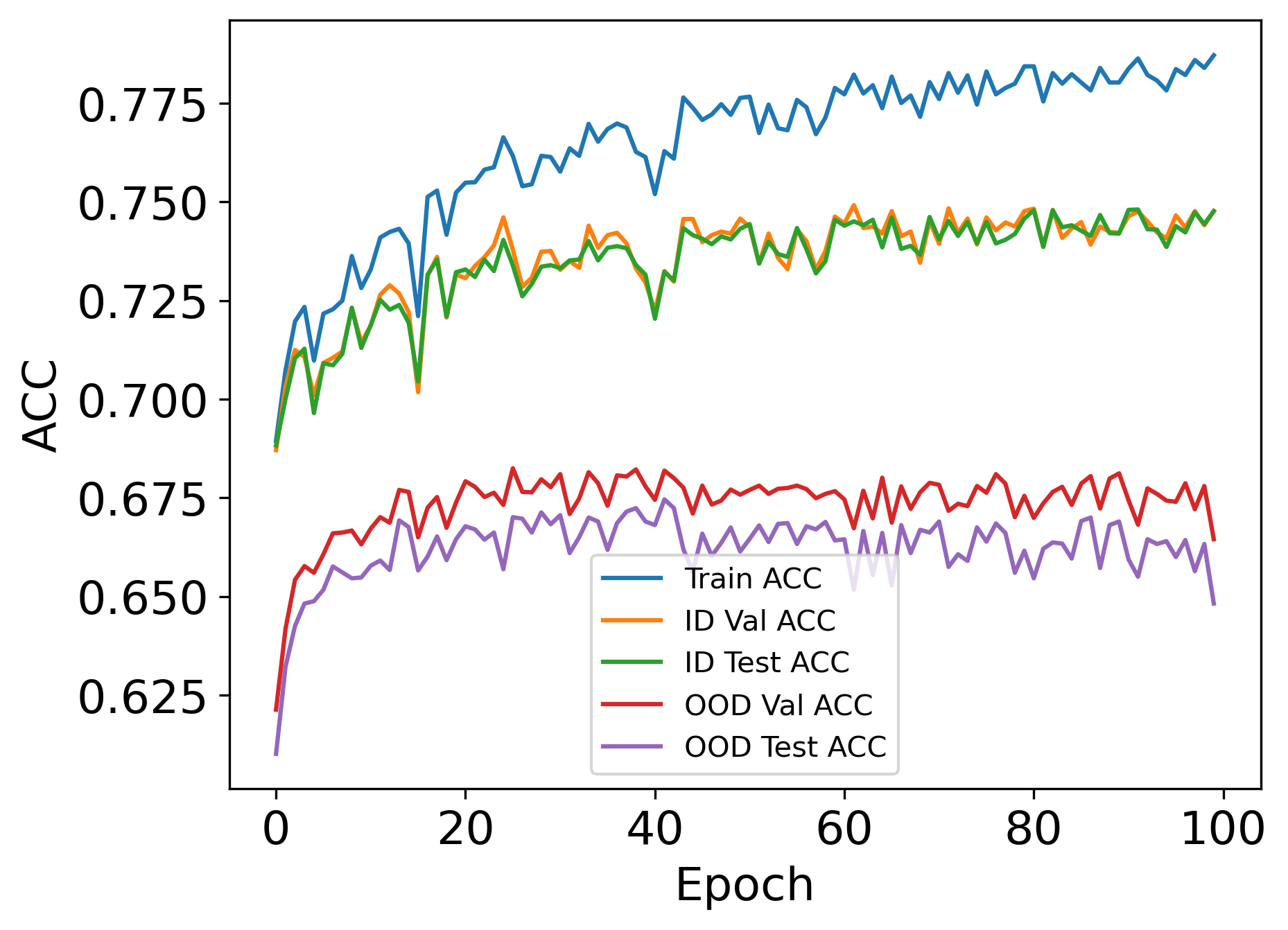}}
    \caption{concept shift + time domain}
\end{subfigure}
\hfill
\begin{subfigure}[t]{0.32\textwidth}
    \raisebox{-\height}{\includegraphics[width=\textwidth]{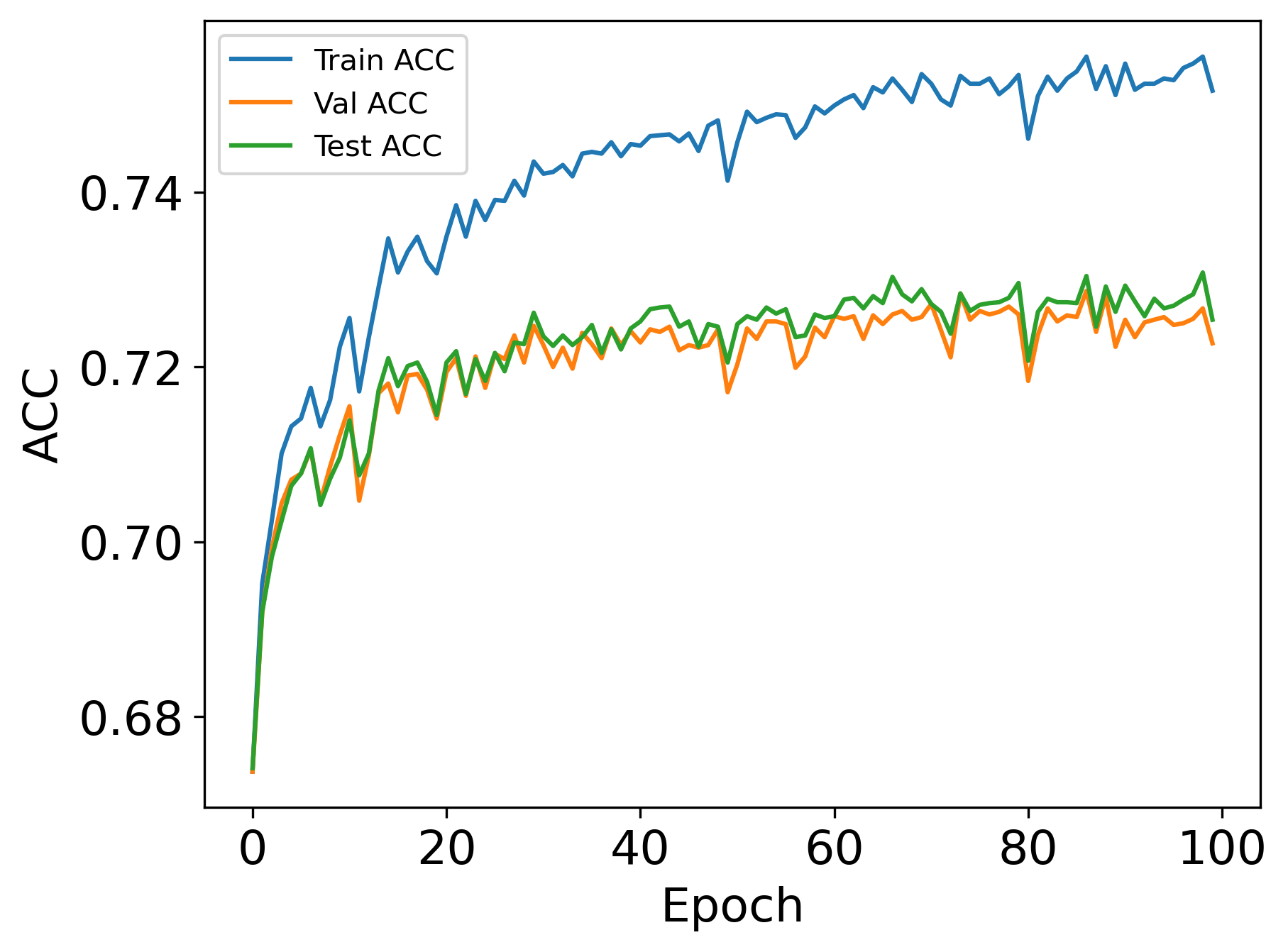}}
    \caption{no shift + time domain}
\end{subfigure}
\hfill
\begin{subfigure}[t]{0.32\textwidth}
    \raisebox{-\height}{\includegraphics[width=\textwidth]{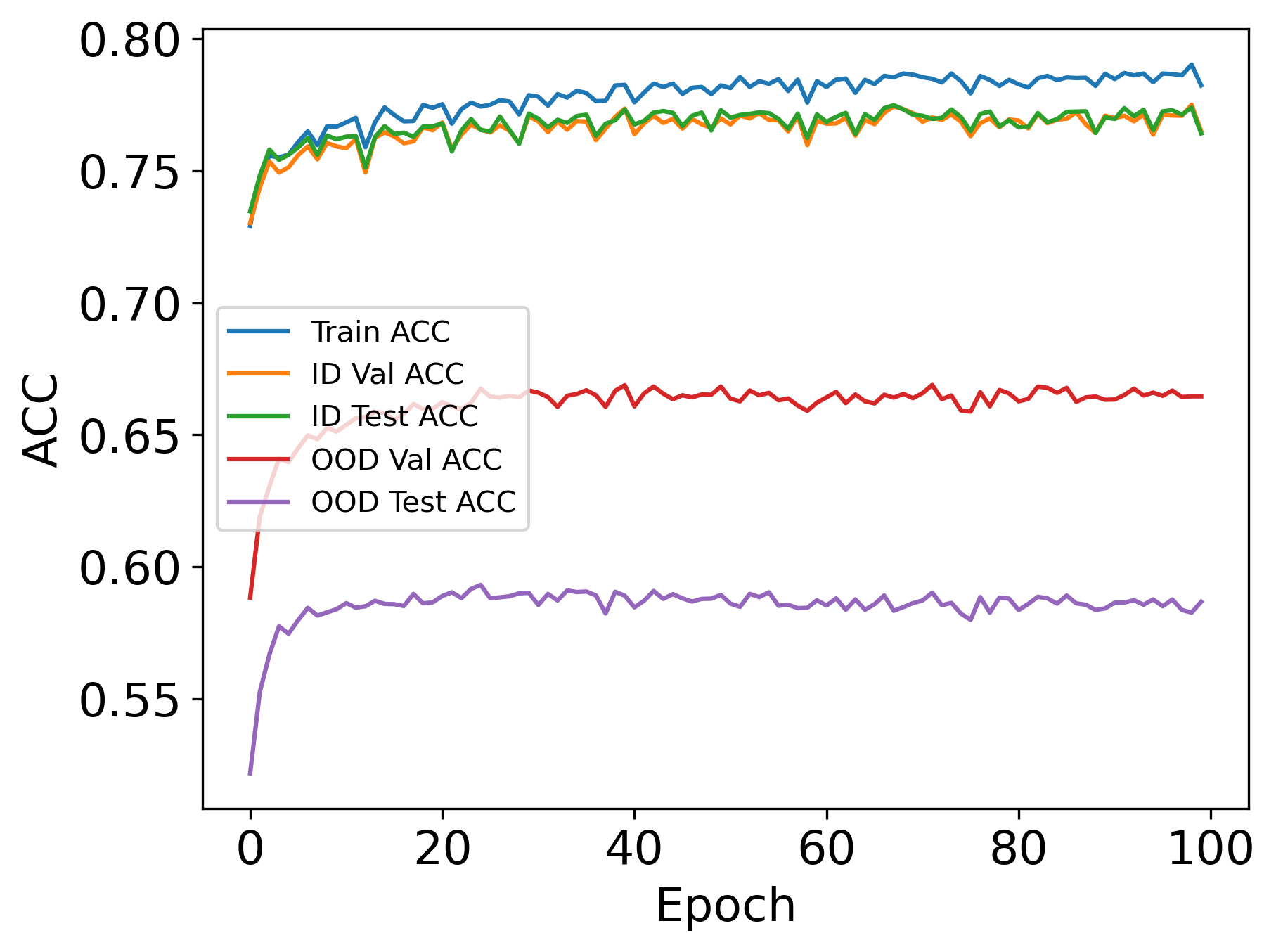}}
    \caption{covariate shift + degree domain}
\end{subfigure}
\hfill
\begin{subfigure}[t]{0.32\textwidth}
    \raisebox{-\height}{\includegraphics[width=\textwidth]{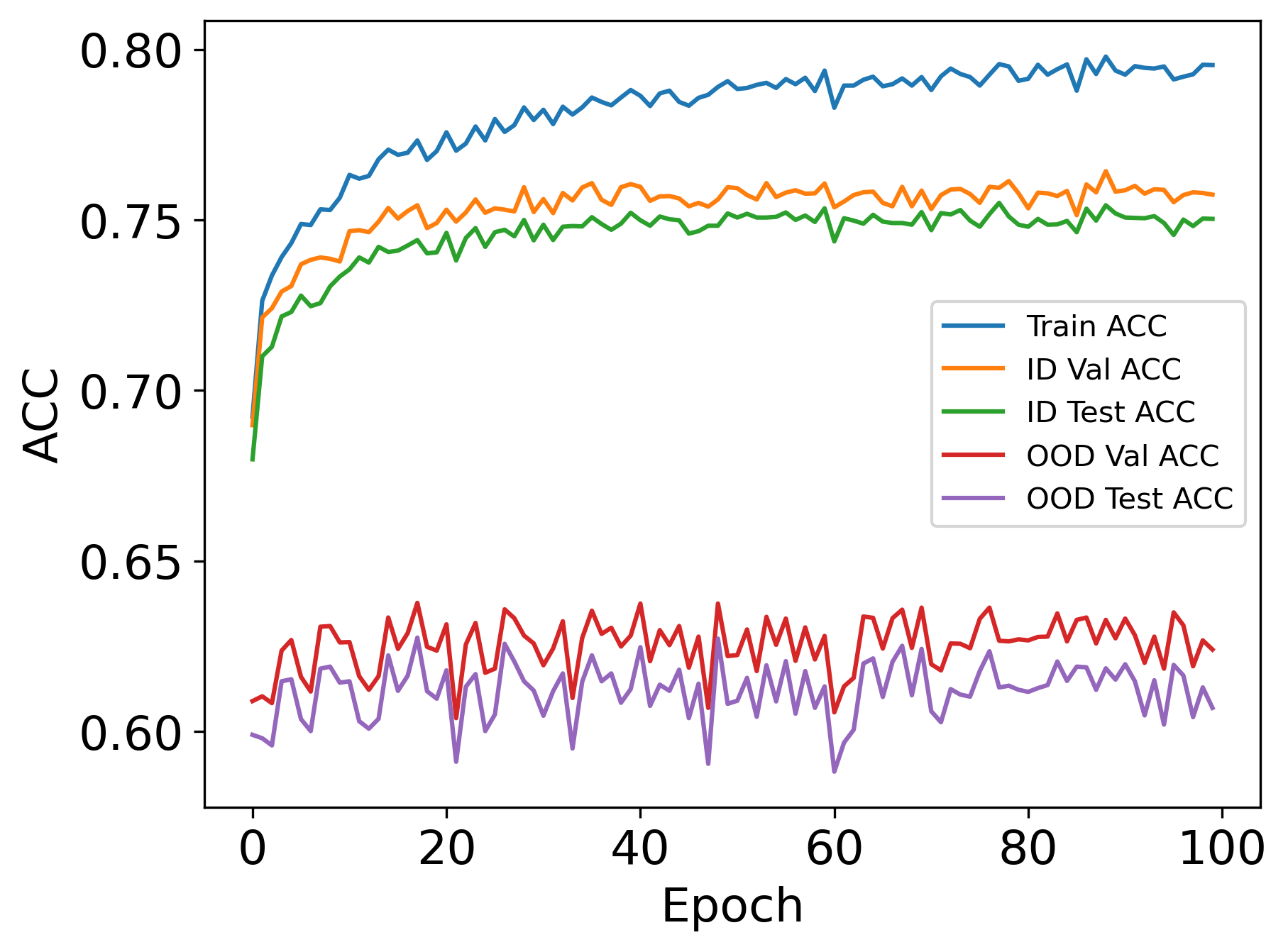}}
    \caption{concept shift + degree domain}
\end{subfigure}
\hfill
\begin{subfigure}[t]{0.32\textwidth}
    \raisebox{-\height}{\includegraphics[width=\textwidth]{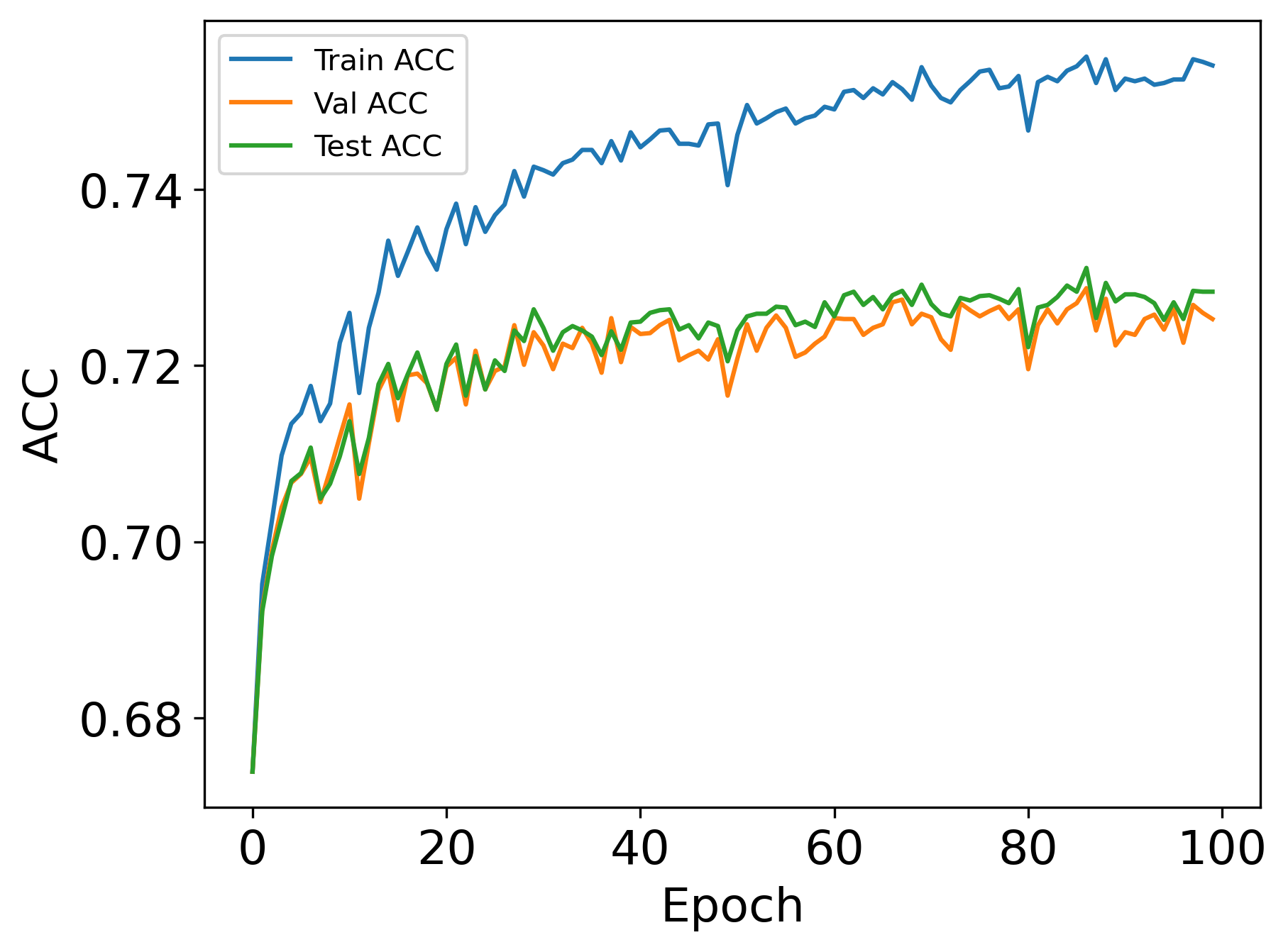}}
    \caption{no shift + degree domain}
\end{subfigure}
\caption{Metric score curves for ERM on GOOD-Arxiv.}
    \label{fig:curve7}
\end{figure}

\begin{figure}[!htbp]
    \centering
\begin{subfigure}[t]{0.32\textwidth}
    \raisebox{-\height}{\includegraphics[width=\textwidth]{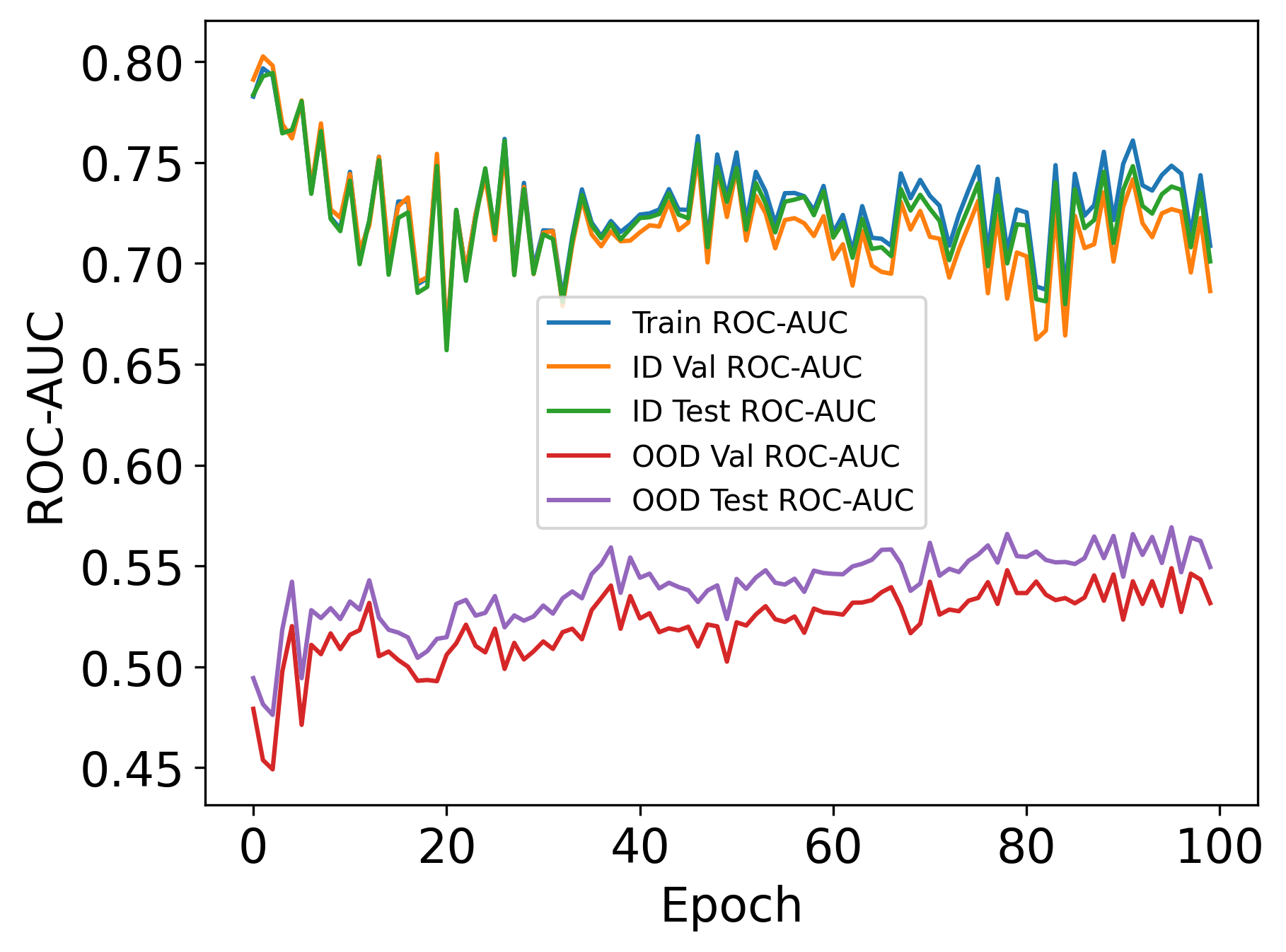}}
    \caption{covariate shift + language domain}
\end{subfigure}
\hfill
\begin{subfigure}[t]{0.32\textwidth}
    \raisebox{-\height}{\includegraphics[width=\textwidth]{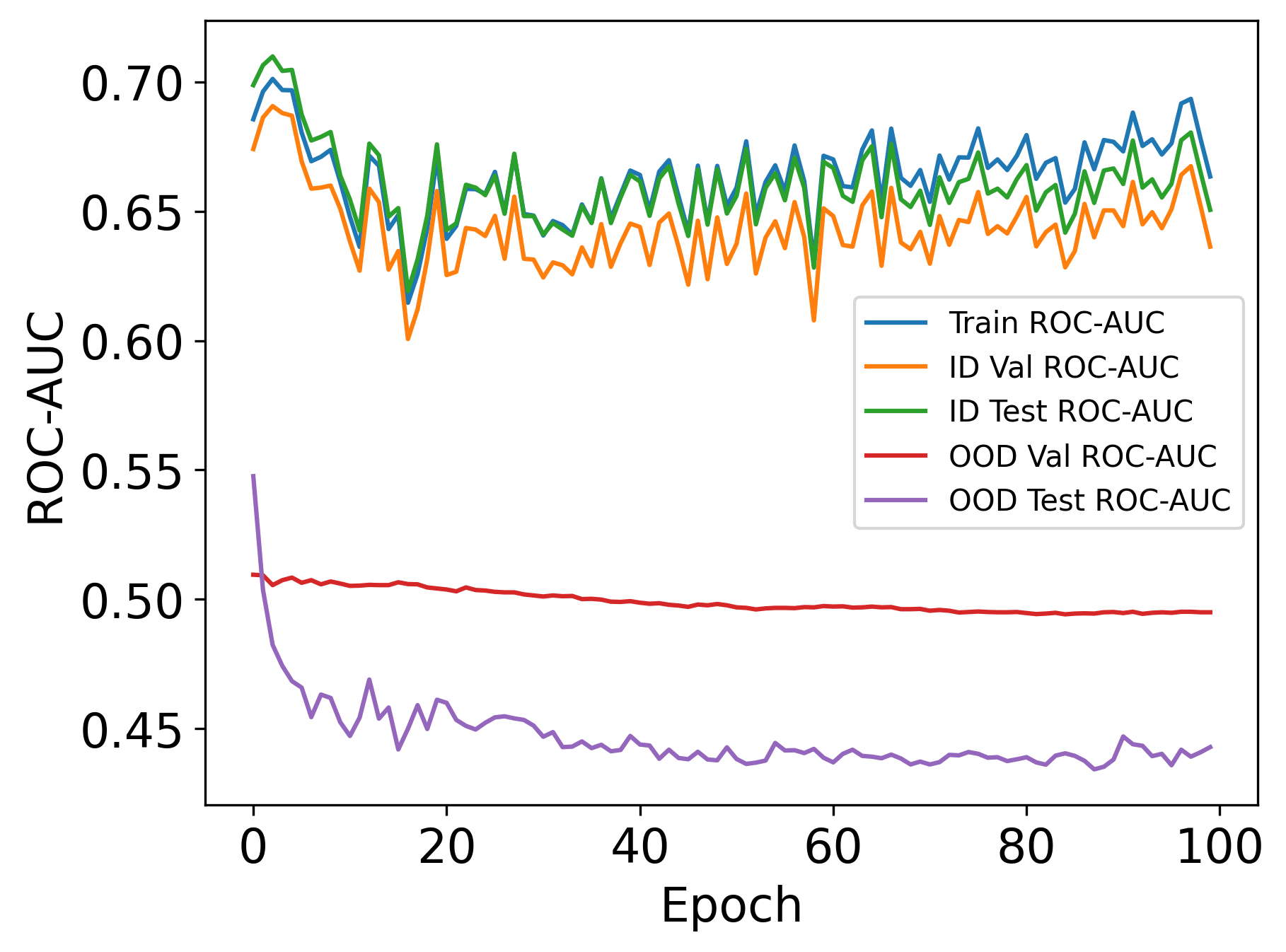}}
    \caption{concept shift + language domain}
\end{subfigure}
\hfill
\begin{subfigure}[t]{0.32\textwidth}
    \raisebox{-\height}{\includegraphics[width=\textwidth]{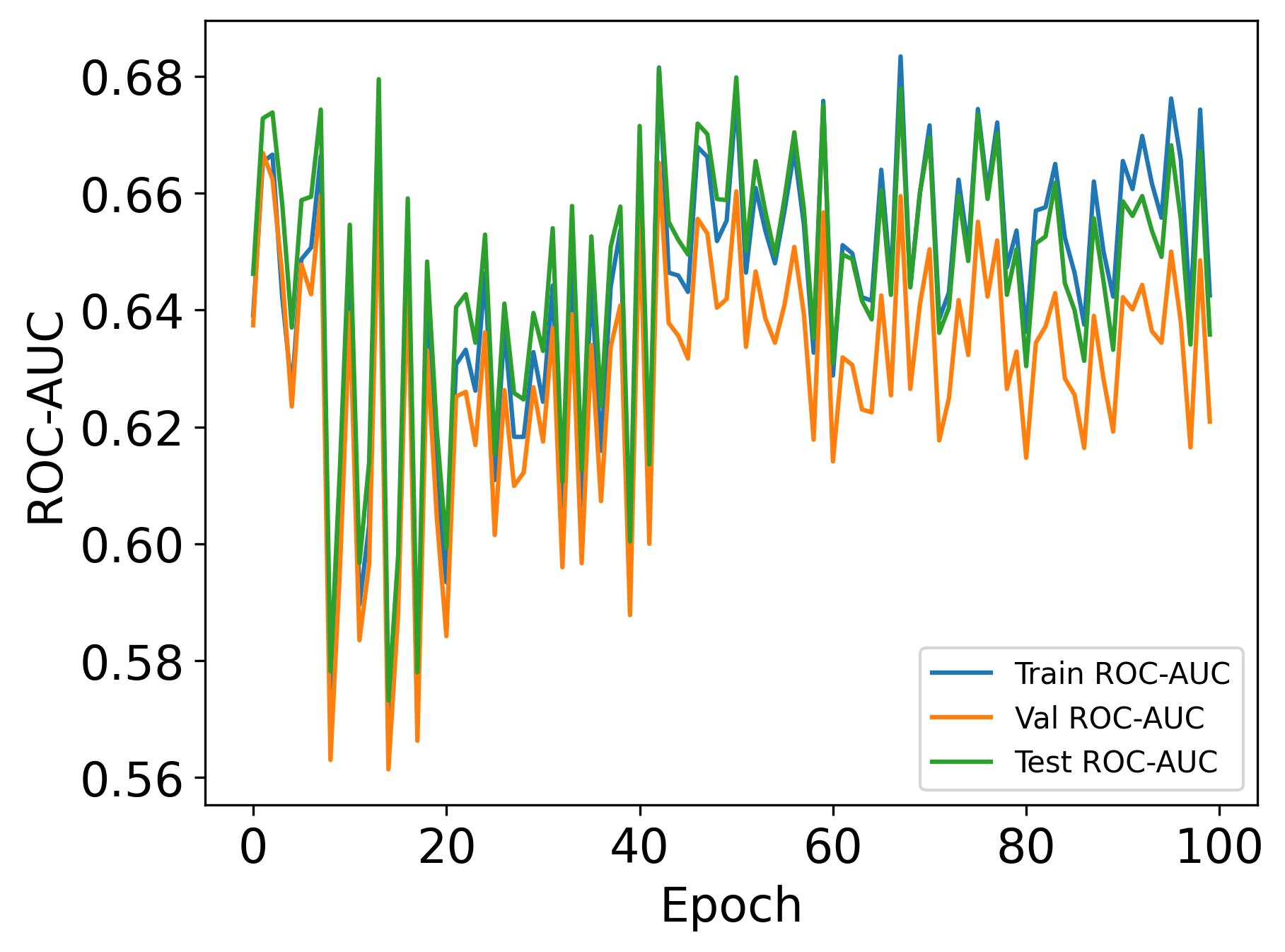}}
    \caption{no shift + language domain}
\end{subfigure}
    \caption{Metric score curves for ERM on GOOD-Twitch.}
    \label{fig:curve8}
\end{figure}

\begin{figure}[!htbp]
    \centering
\begin{subfigure}[t]{0.32\textwidth}
    \raisebox{-\height}{\includegraphics[width=\textwidth]{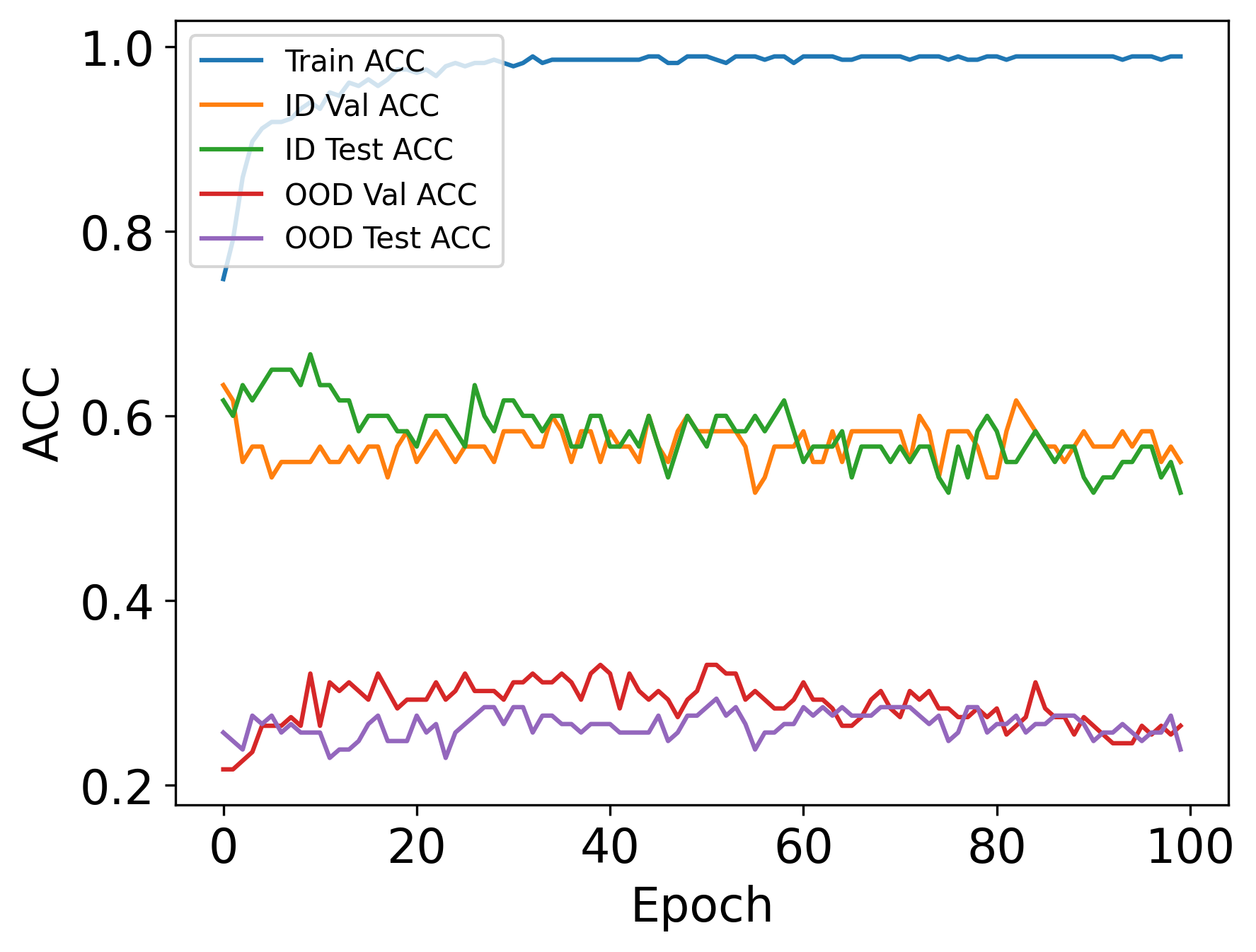}}
    \caption{covariate shift + university domain}
\end{subfigure}
\hfill
\begin{subfigure}[t]{0.32\textwidth}
    \raisebox{-\height}{\includegraphics[width=\textwidth]{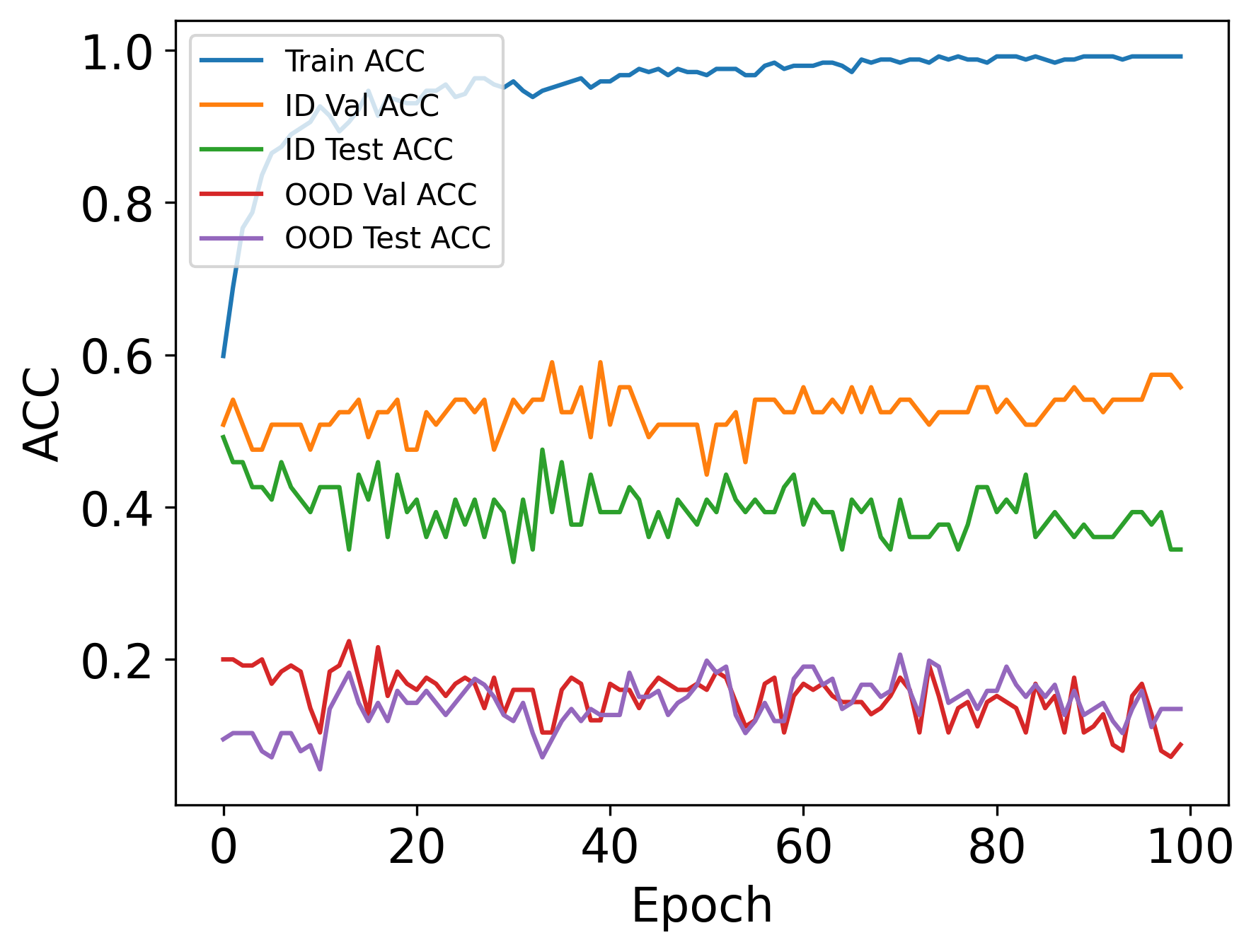}}
    \caption{concept shift + university domain}
\end{subfigure}
\hfill
\begin{subfigure}[t]{0.32\textwidth}
    \raisebox{-\height}{\includegraphics[width=\textwidth]{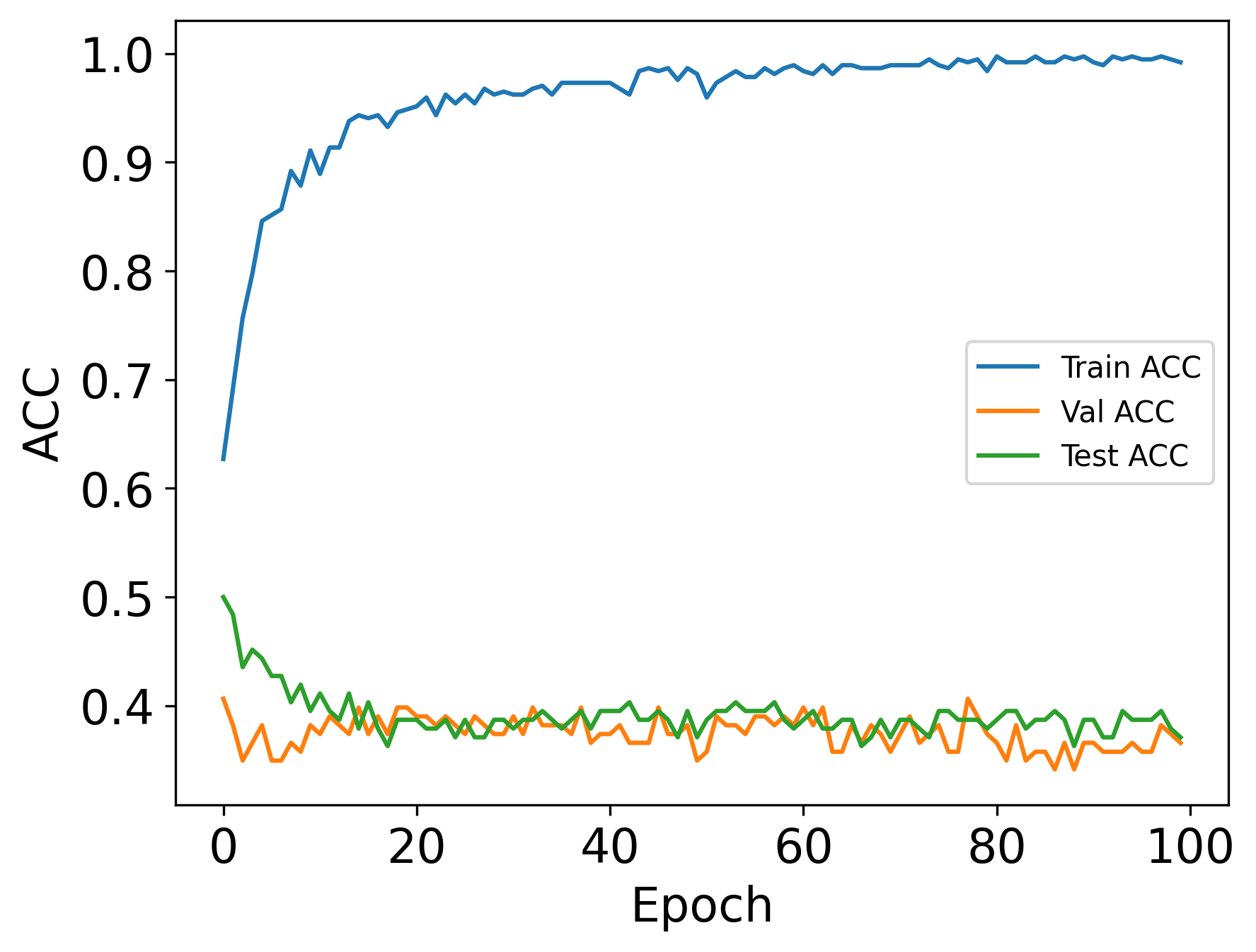}}
    \caption{no shift + university domain}
\end{subfigure}
    \caption{Metric score curves for ERM on GOOD-WebKB.}
    \label{fig:curve9}
\end{figure}

\begin{figure}[!htbp]
    \centering
\begin{subfigure}[t]{0.32\textwidth}
    \raisebox{-\height}{\includegraphics[width=\textwidth]{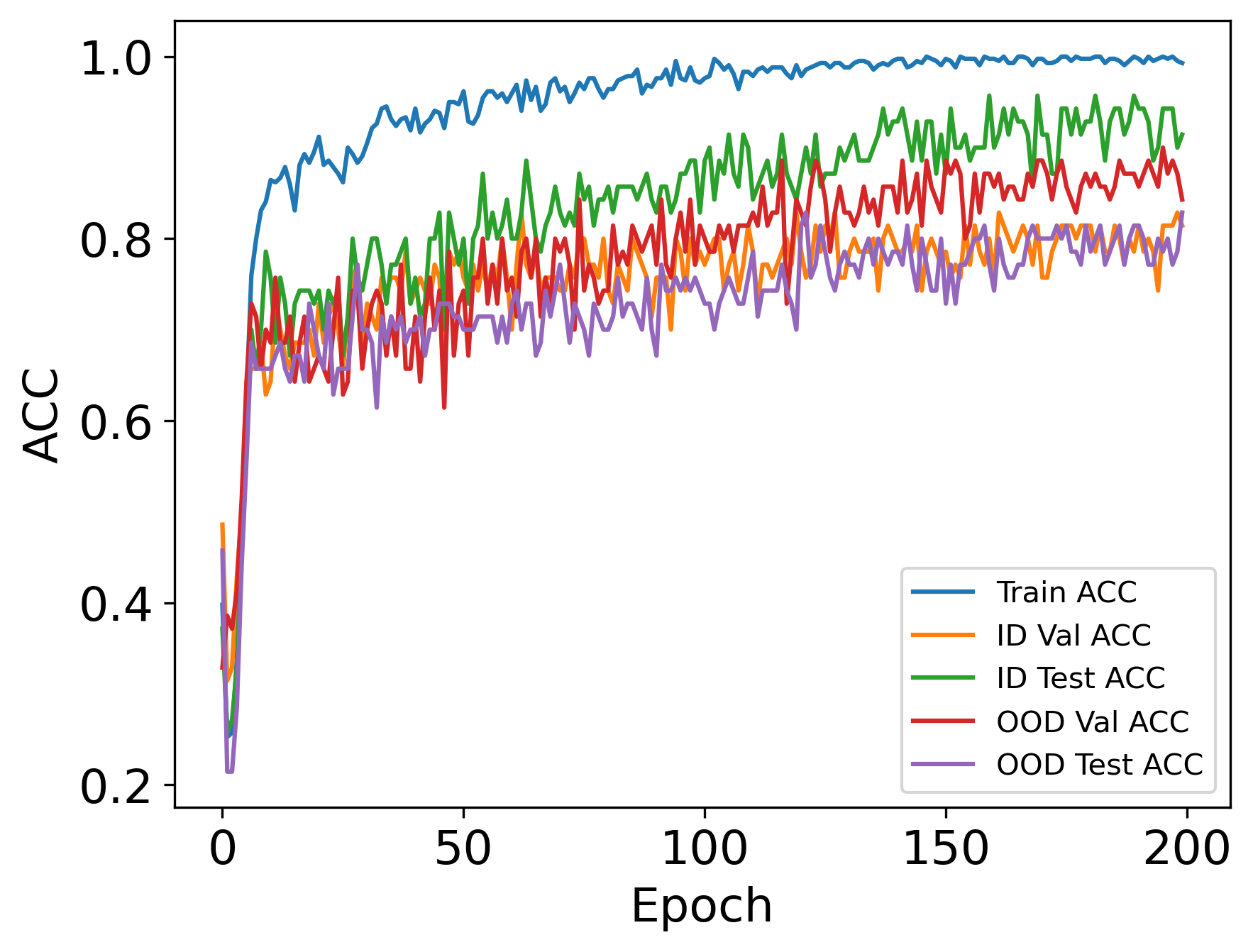}}
    \caption{covariate shift + color domain}
\end{subfigure}
\hfill
\begin{subfigure}[t]{0.32\textwidth}
    \raisebox{-\height}{\includegraphics[width=\textwidth]{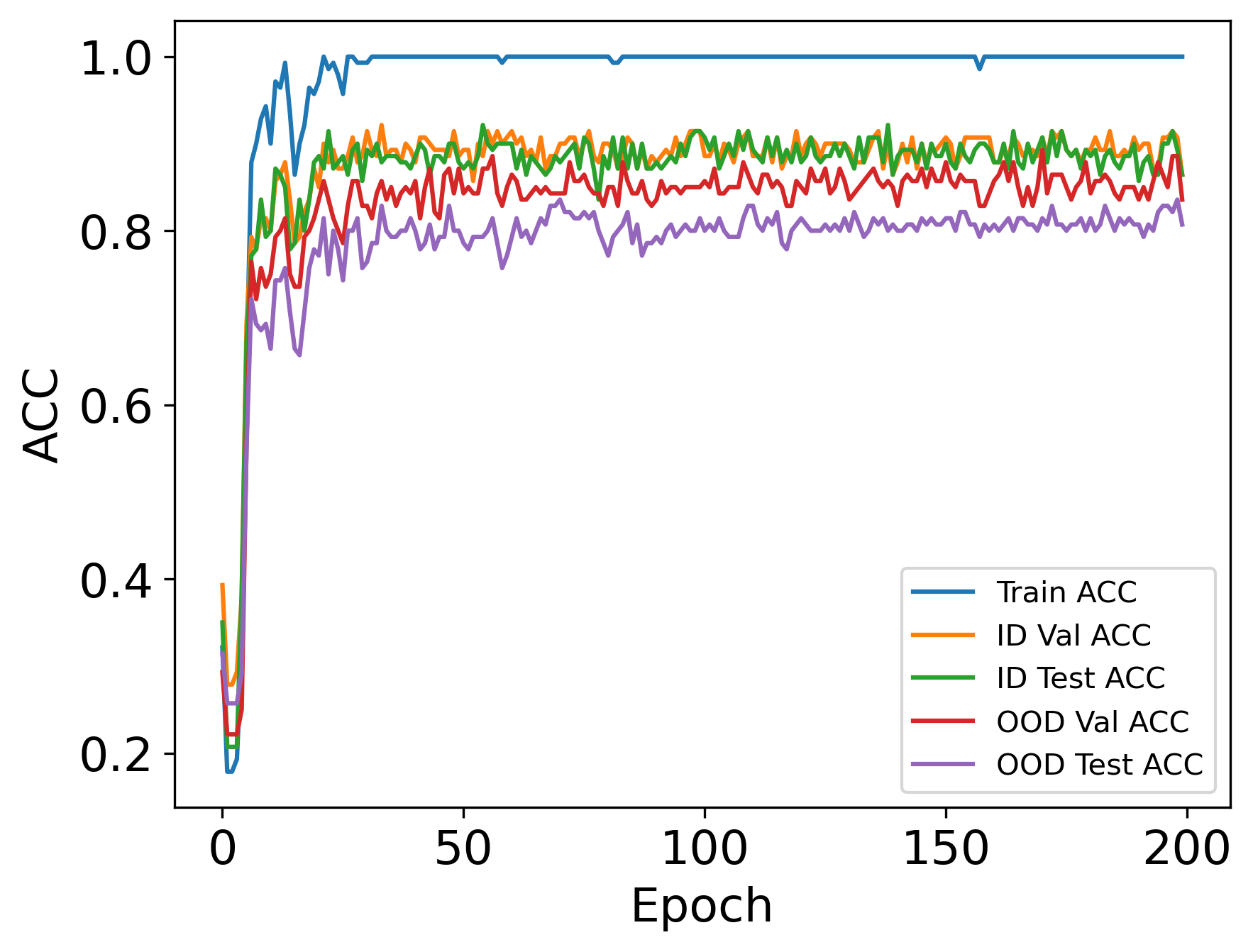}}
    \caption{concept shift + color domain}
\end{subfigure}
\hfill
\begin{subfigure}[t]{0.32\textwidth}
    \raisebox{-\height}{\includegraphics[width=\textwidth]{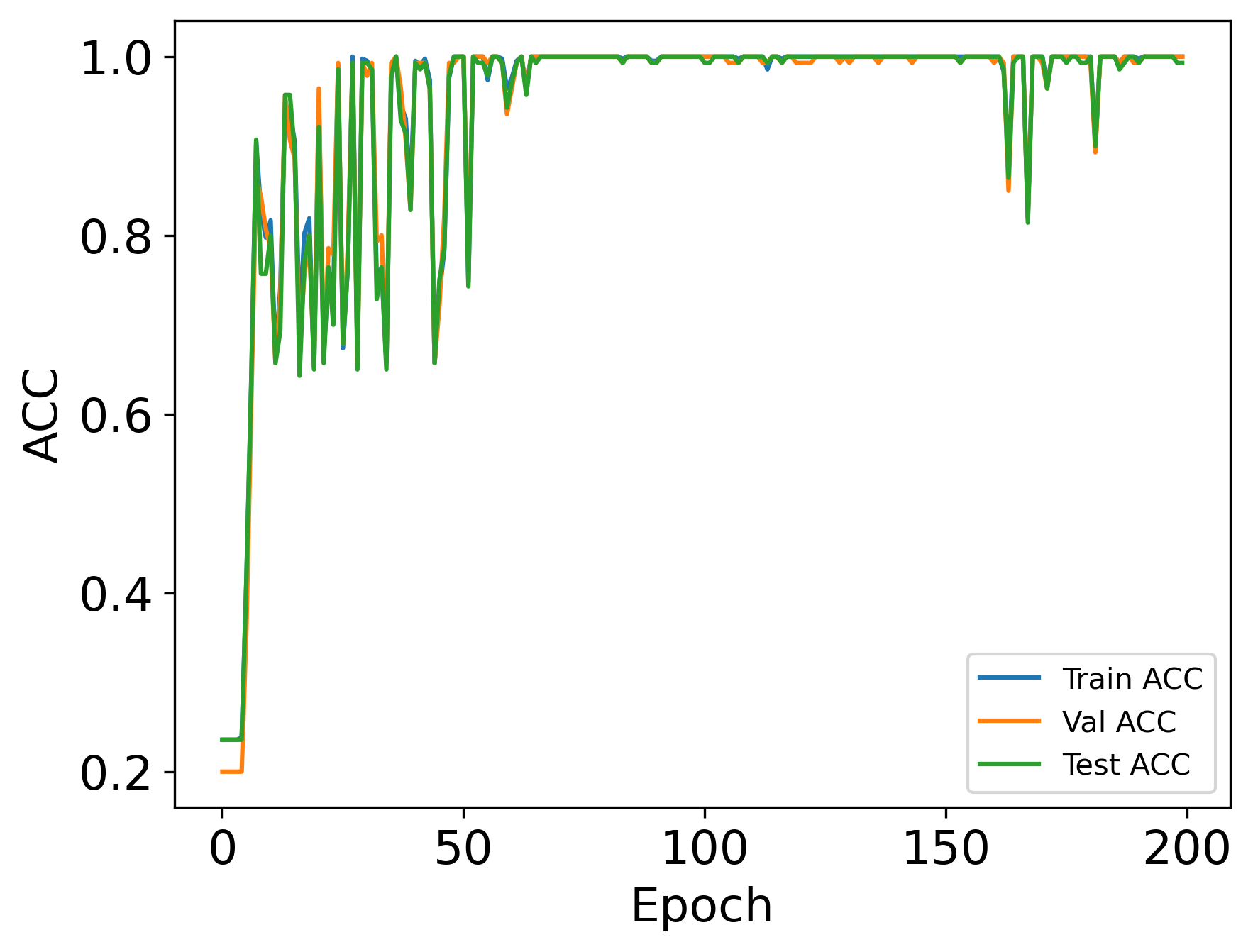}}
    \caption{no shift + color domain}
\end{subfigure}
    \caption{Metric score curves for ERM on GOOD-CBAS.}
    \label{fig:curve10}
\end{figure}

We also report the metric score curves for 11 datasets in Fig.~\ref{fig:curve1}-\ref{fig:curve10}.
Note that we only include the curves for ERM with all splits, while all curve figures for other algorithms are available at our GitHub repository.

\subsection{Comparison between training, validation and test scores}
To directly view performance gaps between training and test data, we compare training, validation, and test scores in Table~\ref{table:tvt}. These comparisons reveal the distribution shift by definition.

\begin{table*}[!htbp]
% \setlength{\abovecaptionskip}{0.1cm}
% \footnotesize
% \scalebox{0.622}{
% % \setlength{\tabcolsep}{0.3mm}{
% \centering
\centering
\resizebox{\textwidth}{!}
{
\begin{tabular}{lllcccccccccc}
\toprule[2pt]
\multicolumn{3}{l}{\multirow{2}{*}{Dataset}} & \multicolumn{1}{c}{\multirow{2}{*}{Domain}} & \multicolumn{1}{c}{\multirow{2}{*}{Shift}} & \multicolumn{4}{c}{ID validation} & \multicolumn{4}{c}{OOD validation}\\ 
\cmidrule(r){6-9} \cmidrule(r){10-13}
& & & & & \multicolumn{1}{c}{Train} & \multicolumn{1}{c}{Validation} & \multicolumn{1}{c}{ID test} & \multicolumn{1}{c}{OOD test} & \multicolumn{1}{c}{Train} & \multicolumn{1}{c}{Validation} & \multicolumn{1}{c}{ID test} & \multicolumn{1}{c}{OOD test} \\
\midrule[1pt]

\multicolumn{3}{l}{\multirow{4}{*}{GOOD-HIV\textuparrow}} & \multicolumn{1}{c}{\multirow{2}{*}{scaffold}} & covariate & 99.40 & 84.11 & 82.62 & 68.65 & 91.63 & 78.94 & 81.49 & 69.57\\
& & & & concept & 94.04 & 83.56 & 82.63 & 58.28 & 99.56 & 76.92 & 80.55 & 72.43\\
& & & \multicolumn{1}{c}{\multirow{2}{*}{size}} & covariate & 99.76 & 86.34 & 83.58 & 59.26 & 87.84 & 74.86 & 82.14 & 54.68\\
& & & & concept & 98.53 & 91.93 & 88.38 & 50.07 & 99.97 & 61.48 & 83.89 & 63.38\\
\midrule[1pt]
\multicolumn{3}{l}{\multirow{4}{*}{GOOD-PCBA\textuparrow}} & \multicolumn{1}{c}{\multirow{2}{*}{scaffold}} & covariate & 51.57 & 33.74 & 32.75 & 17.01 & 47.95 & 20.75 & 32.76 & 16.49\\
& & & & concept & 47.86 & 28.42 & 25.86 & 20.40 & 49.57 & 22.00 & 26.08 & 21.24\\
& & & \multicolumn{1}{c}{\multirow{2}{*}{size}} & covariate & 52.91 & 33.89 & 34.17 & 18.26 & 52.91 & 27.23 & 34.17 & 18.26\\
& & & & concept & 55.16 & 34.56 & 33.64 & 16.45 & 55.78 & 17.32 & 33.36 & 16.49\\
\midrule[1pt]
\multicolumn{3}{l}{\multirow{4}{*}{GOOD-ZINC\textdownarrow}} & \multicolumn{1}{c}{\multirow{2}{*}{scaffold}} & covariate & 0.1183 & 0.1224 & 0.1224 & 0.1895 & 0.1380 & 0.1421 & 0.1409 & 0.2159\\
& & & & concept & 0.1074 & 0.1138 & 0.1128 & 0.1243 & 0.1086 & 0.1232 & 0.1141 & 0.1239\\
& & & \multicolumn{1}{c}{\multirow{2}{*}{size}} & covariate & 0.1167 & 0.1215 & 0.1214 & 0.2581 & 0.1210 & 0.1313 & 0.1259 & 0.2352\\
& & & & concept & 0.1117 & 0.1142 & 0.1162 & 0.1370 & 0.1244 & 0.1286 & 0.1298 & 0.1279\\
\midrule[1pt]
\multicolumn{3}{l}{\multirow{2}{*}{GOOD-SST2\textuparrow}} & \multicolumn{1}{c}{\multirow{2}{*}{length}} & covariate & 100.00 & 90.10 & 89.81 & 76.03 & 99.79 & 85.78 & 88.74 & 80.58\\
& & & & concept & 100.00 & 94.64 & 94.39 & 67.13 & 99.90 & 74.34 & 93.96 & 72.41\\
\midrule[1pt]
\multicolumn{3}{l}{\multirow{2}{*}{GOOD-CMNIST\textuparrow}} & \multicolumn{1}{c}{\multirow{2}{*}{color}} & covariate & 93.54 & 77.83 & 77.17 & 26.39 & 97.07 & 64.39 & 76.97 & 29.49\\
& & & & concept & 92.15 & 89.46 & 90.00 & 36.68 & 99.13 & 60.01 & 89.40 & 42.60\\
\midrule[1pt]
\multicolumn{3}{l}{\multirow{4}{*}{GOOD-Motif\textuparrow}} & \multicolumn{1}{c}{\multirow{2}{*}{base}} & covariate & 92.90 & 92.70 & 92.67 & 70.43 & 91.60 & 90.53 & 91.93 & 66.57\\
& & & & concept & 93.06 & 92.74 & 91.96 & 80.37 & 93.08 & 84.45 & 92.00 & 80.78\\
& & & \multicolumn{1}{c}{\multirow{2}{*}{size}} & covariate & 92.06 & 92.87 & 92.40 & 55.63 & 91.85 & 88.57 & 92.17 & 54.33\\
& & & & concept & 91.80 & 92.63 & 91.81 & 69.93 & 91.97 & 76.67 & 91.81 & 71.35\\
\midrule[1pt]
\multicolumn{3}{l}{\multirow{4}{*}{GOOD-Cora\textuparrow}} & \multicolumn{1}{c}{\multirow{2}{*}{word}} & covariate & 83.61 & 69.02 & 70.79 & 65.38 & 83.61 & 67.98 & 70.79 & 65.38\\
& & & & concept & 84.96 & 71.22 & 65.77 & 64.84 & 84.96 & 65.92 & 65.77 & 64.84\\
& & & \multicolumn{1}{c}{\multirow{2}{*}{degree}} & covariate & 82.98 & 72.46 & 72.36 & 56.25 & 81.29 & 64.30 & 72.92 & 56.49\\
& & & & concept & 85.98 & 70.41 & 68.61 & 60.51 & 83.80 & 61.68 & 68.49 & 60.93\\
\midrule[1pt]
\multicolumn{3}{l}{\multirow{4}{*}{GOOD-Arxiv\textuparrow}} & \multicolumn{1}{c}{\multirow{2}{*}{time}} & covariate & 76.74 & 73.70 & 73.00 & 70.30 & 76.69 & 72.21 & 72.66 & 71.16\\
& & & & concept & 78.23 & 74.92 & 74.51 & 65.17 & 76.17 & 68.25 & 73.39 & 67.01\\
& & & \multicolumn{1}{c}{\multirow{2}{*}{degree}} & covariate & 79.02 & 77.50 & 77.39 & 58.27 & 78.48 & 66.89 & 76.96 & 59.03\\
& & & & concept & 79.79 & 76.43 & 75.43 & 61.85 & 77.33 & 63.77 & 74.41 & 62.75\\
\midrule[1pt]
\multicolumn{3}{l}{\multirow{2}{*}{GOOD-Twitch\textuparrow}} & \multicolumn{1}{c}{\multirow{2}{*}{language}} & covariate & 70.11 & 69.06 & 70.98 & 48.24 & 68.55 & 50.95 & 69.87 & 54.76\\
& & & &  concept & 79.67 & 80.26 & 79.27 & 48.15 & 74.84 & 54.89 & 73.81 & 56.92\\
\midrule[1pt]
\multicolumn{3}{l}{\multirow{2}{*}{GOOD-WebKB\textuparrow}} & \multicolumn{1}{c}{\multirow{2}{*}{university}} & covariate & 95.08 & 59.02 & 39.34 & 9.52 & 90.57 & 22.40 & 34.43 & 18.25\\
& & & & concept & 74.82 & 63.33 & 61.67 & 25.69 & 98.58 & 33.02 & 60.00 & 26.61\\
\midrule[1pt]
\multicolumn{3}{l}{\multirow{2}{*}{GOOD-CBAS\textuparrow}} & \multicolumn{1}{c}{\multirow{2}{*}{color}} & covariate & 94.05 & 82.86 & 82.86 & 70.00 & 99.76 & 80.00 & 91.43 & 77.14\\
& & & & concept & 100.00 & 92.14 & 87.86 & 82.86 & 100.00 & 89.29 & 90.71 & 81.43\\

\bottomrule[2pt]
\end{tabular}
}
% }
\caption{Comparison between training, validation and test scores for ERM on 11 datasets. The scores are evaluated on the final model of a random run. \textuparrow { }indicates higher values correspond to better performance while \textdownarrow { }indicates lower values for better performance.}
\label{table:tvt}
\end{table*}

\section{Complete OOD Parameter Selections}\label{sec:E}
Following Appendix~\ref{sec:B}, in this section we specify the hyperparameter tune set and selection for each algorithm on each dataset in Table~\ref{table:h1}-\ref{table:h10}.
%~\ref{table:h2}~\ref{table:h3}~\ref{table:h4}~\ref{table:h5}~\ref{table:h6}~\ref{table:h7}~\ref{table:h8}.
\begin{table*}[!htbp]
% \setlength{\abovecaptionskip}{0.05cm}
% \footnotesize
% \scalebox{0.93}{
% % \setlength{\tabcolsep}{0.34mm}{
% \centering
\centering
\resizebox{\textwidth}{!}
{
\begin{tabular}{lllccccccccc}
\toprule[2pt]
\multicolumn{3}{c}{\multirow{2}{*}{GOOD-HIV}} & \multicolumn{3}{c}{\multirow{2}{*}{tune set}} & \multicolumn{3}{c}{scaffold} & \multicolumn{3}{c}{size}\\
\cmidrule(r){7-9} \cmidrule(r){10-12}
& & & & & & \multicolumn{1}{c}{covariate} & \multicolumn{1}{c}{concept} & \multicolumn{1}{c}{no shift} & \multicolumn{1}{c}{covariate} & \multicolumn{1}{c}{concept} & \multicolumn{1}{c}{no shift} \\
\midrule[1pt]

\multicolumn{3}{l}{ERM} & -- & -- & -- & -- & -- & -- & -- & -- & -- \\
\multicolumn{3}{l}{IRM} & 10.0 & 0.1 & 1.0 & 1.0 & 0.1 & 0.1 & 10.0 & 0.1 & 0.1 \\
\multicolumn{3}{l}{VREx} & 10.0 & 1000.0 & 100.0& 100.0 & 10.0 & 100.0 & 10.0 & 1000.0 & 100.0 \\
\multicolumn{3}{l}{GroupDRO} & 0.01 & 0.1 & 0.001 & 0.1 & 0.01 & 0.001 & 0.01 & 0.001 & 0.001 \\
\multicolumn{3}{l}{DANN} & 0.1 & 1.0 & 0.01 & 1.0 & 0.1 & 0.01 & 0.01 & 1.0 & 0.01 \\
\multicolumn{3}{l}{Deep Coral} & 0.01 & 1.0 & 0.1 & 0.1 & 0.01 & 0.01 & 0.1 & 0.01 & 0.01 \\
\multicolumn{3}{l}{Mixup} & 1.0 & 2.0 & 0.4 & 2.0 & 0.4 & 2.0 & 2.0 & 0.4 & 2.0 \\
\multicolumn{3}{l}{DIR} & 0.4 & 0.6 & 0.8 & 0.8 & 0.8 & 0.8 & 0.8 & 0.8 & 0.8 \\

\bottomrule[2pt]
\end{tabular}
}
% }
\caption{
OOD hyperparameter selections on GOOD-HIV.
}
\label{table:h1}
\end{table*}
% \vspace{-0.6cm}
\begin{table*}[!htbp]
% \setlength{\abovecaptionskip}{0.05cm}
% \footnotesize
% \scalebox{0.93}{
% % \setlength{\tabcolsep}{0.34mm}{
% \centering
\centering
\resizebox{\textwidth}{!}
{
\begin{tabular}{lllccccccccc}
\toprule[2pt]
\multicolumn{3}{c}{\multirow{2}{*}{GOOD-PCBA}} & \multicolumn{3}{c}{\multirow{2}{*}{tune set}} & \multicolumn{3}{c}{scaffold} & \multicolumn{3}{c}{size}\\
\cmidrule(r){7-9} \cmidrule(r){10-12}
& & & & & & \multicolumn{1}{c}{covariate} & \multicolumn{1}{c}{concept} & \multicolumn{1}{c}{no shift} & \multicolumn{1}{c}{covariate} & \multicolumn{1}{c}{concept} & \multicolumn{1}{c}{no shift} \\
\midrule[1pt]

\multicolumn{3}{l}{ERM} & -- & -- & -- & -- & -- & -- & -- & -- & -- \\
\multicolumn{3}{l}{IRM} & 1.0 & 0.1 & 10.0 & 0.1 & 0.1 & 0.1 & 0.1 & 1.0 & 0.1 \\
\multicolumn{3}{l}{VREx} & 10.0 & 100.0 & 1.0 & 10.0 & 100.0 & 10.0 & 1.0 & 10.0 & 10.0 \\
\multicolumn{3}{l}{GroupDRO} & 0.01 & 0.001 & 0.1 & 0.01 & 0.001 & 0.1 & 0.1 & 0.01 & 0.1 \\
\multicolumn{3}{l}{DANN} & 0.01 & 0.001 & 0.1 & 0.01 & 0.01 & 0.01 & 0.01 & 0.01 & 0.01 \\
\multicolumn{3}{l}{Deep Coral}& 0.1 & 0.01 & 1.0 & 0.01 & 0.1 & 1.0 & 0.1 & 0.1 & 1.0 \\
\multicolumn{3}{l}{Mixup} & 1.0 & 2.0 & 0.4 & 1.0 & 2.0 & 1.0 & 2.0 & 1.0 & 1.0 \\
\multicolumn{3}{l}{DIR} & 0.4 & 0.6 & 0.8 & 0.8 & 0.8 & 0.8 & 0.8 & 0.8 & 0.8 \\

\bottomrule[2pt]
\end{tabular}
}
% }
\caption{
OOD hyperparameter selections on GOOD-PCBA.
}
\label{table:h2}
\end{table*}
% \vspace{-0.6cm}
\begin{table*}[!htbp]
% \setlength{\abovecaptionskip}{0.05cm}
% \footnotesize
% \scalebox{0.93}{
% % \setlength{\tabcolsep}{0.34mm}{
% \centering
\centering
\resizebox{\textwidth}{!}
{
\begin{tabular}{lllccccccccc}
\toprule[2pt]
\multicolumn{3}{c}{\multirow{2}{*}{GOOD-ZINC}} & \multicolumn{3}{c}{\multirow{2}{*}{tune set}} & \multicolumn{3}{c}{scaffold} & \multicolumn{3}{c}{size}\\
\cmidrule(r){7-9} \cmidrule(r){10-12}
& & & & & & \multicolumn{1}{c}{covariate} & \multicolumn{1}{c}{concept} & \multicolumn{1}{c}{no shift} & \multicolumn{1}{c}{covariate} & \multicolumn{1}{c}{concept} & \multicolumn{1}{c}{no shift} \\
\midrule[1pt]

\multicolumn{3}{l}{ERM} & -- & -- & -- & -- & -- & -- & -- & -- & -- \\
\multicolumn{3}{l}{IRM} & 1.0 & 0.1 & 0.01 & 0.01 & 0.01 & 0.01 & 0.01 & 0.01 & 0.01 \\
\multicolumn{3}{l}{VREx} & 100.0 & 10.0 & 1000.0 & 1000.0 & 100.0 & 100.0 & 1000.0 & 100.0 & 100.0 \\
\multicolumn{3}{l}{GroupDRO} & 0.01 & 0.1 & 0.001 & 0.1 & 0.001 & 0.1 & 0.001 & 0.001 & 0.1 \\
\multicolumn{3}{l}{DANN} & 0.01 & 0.001 & 0.1 & 0.001 & 0.001 & 0.1 & 0.01 & 0.1 & 0.1 \\
\multicolumn{3}{l}{Deep Coral} & 0.1 & 0.01 & 1.0 & 1.0 & 1.0 & 1.0 & 1.0 & 1.0 & 1.0 \\
\multicolumn{3}{l}{Mixup} & 1.0 & 0.4 & 2.0 & 0.4 & 1.0 & 1.0 & 1.0 & 0.4 & 1.0 \\
\multicolumn{3}{l}{DIR} & 0.4 & 0.6 & 0.8 & 0.8 & 0.8 & 0.8 & 0.8 & 0.8 & 0.8 \\

\bottomrule[2pt]
\end{tabular}
}
% }
\caption{
OOD hyperparameter selections on GOOD-ZINC.
}
\label{table:h3}
\end{table*}
\begin{table*}[!htbp]
% \setlength{\abovecaptionskip}{0.1cm}
% \footnotesize
% \scalebox{0.93}{
% % \setlength{\tabcolsep}{0.34mm}{
% \centering
\centering
\resizebox{0.7\textwidth}{!}
{
\begin{tabular}{lllcccccc}
\toprule[2pt]
\multicolumn{3}{c}{\multirow{2}{*}{GOOD-SST2}} & \multicolumn{3}{c}{\multirow{2}{*}{tune set}} & \multicolumn{3}{c}{length}\\
\cmidrule(r){7-9}
& & & & & & \multicolumn{1}{c}{covariate} & \multicolumn{1}{c}{concept} & \multicolumn{1}{c}{no shift}\\
\midrule[1pt]

\multicolumn{3}{l}{ERM} & -- & -- & -- & -- & -- & -- \\
\multicolumn{3}{l}{IRM} & 0.1 & 1.0 & 10.0 & 0.1 & 10.0 & 1.0 \\
\multicolumn{3}{l}{VREx} & 1000.0 & 100.0 & 10.0 & 100.0 & 100.0 & 10.0 \\
\multicolumn{3}{l}{GroupDRO} & 0.01 & 0.001 & 0.1 & 0.01 & 0.001 & 0.001 \\
\multicolumn{3}{l}{DANN} & 0.1 & 0.01 & 1.0 & 0.01 & 0.1 & 0.01 \\
\multicolumn{3}{l}{Deep Coral} & 0.1 & 1.0 & 0.01 & 1.0 & 1.0 & 0.1 \\
\multicolumn{3}{l}{Mixup} & 0.4 & 2.0 & 1.0 & 1.0 & 1.0 & 1.0 \\
\multicolumn{3}{l}{DIR} & 0.6 & 0.7 & 0.8 & 0.8 & 0.7 & 0.8 \\

\bottomrule[2pt]
\end{tabular}
}
% }
\caption{
OOD hyperparameter selections on GOOD-SST2.
}
\label{table:h35}
\end{table*}
\begin{table*}[!htbp]
% \setlength{\abovecaptionskip}{0.1cm}
% \footnotesize
% \scalebox{0.93}{
% % \setlength{\tabcolsep}{0.34mm}{
% \centering
\centering
\resizebox{0.7\textwidth}{!}
{
\begin{tabular}{lllcccccc}
\toprule[2pt]
\multicolumn{3}{c}{\multirow{2}{*}{GOOD-CMNIST}} & \multicolumn{3}{c}{\multirow{2}{*}{tune set}} & \multicolumn{3}{c}{color}\\
\cmidrule(r){7-9}
& & & & & & \multicolumn{1}{c}{covariate} & \multicolumn{1}{c}{concept} & \multicolumn{1}{c}{no shift}\\
\midrule[1pt]

\multicolumn{3}{l}{ERM} & -- & -- & -- & -- & -- & -- \\
\multicolumn{3}{l}{IRM} & 0.1 & 1.0 & 0.01 & 0.1 & 0.1 & 1.0 \\
\multicolumn{3}{l}{VREx} & 0.01 & 0.1 & 1.0 & 1.0 & 0.01 & 0.1 \\
\multicolumn{3}{l}{GroupDRO} & 0.001 & 0.01 & 0.1 & 0.1 & 0.01 & 0.1 \\
\multicolumn{3}{l}{DANN} & 0.1 & 0.01 & 0.001 & 0.1 & 0.01 & 0.001 \\
\multicolumn{3}{l}{Deep Coral} & 0.1 & 0.01 & 0.001 & 0.1 & 0.0001 & 0.001 \\
\multicolumn{3}{l}{Mixup} & 1.0 & 2.0 & 0.4 & 1.0 & 0.4 & 0.4 \\
\multicolumn{3}{l}{DIR} & 0.4 & 0.6 & 0.8 & 0.6 & 0.6 & 0.6 \\

\bottomrule[2pt]
\end{tabular}
}
% }
\caption{
OOD hyperparameter selections on GOOD-CMNIST.
}
\label{table:h4}
\end{table*}
\begin{table*}[!htbp]
% \setlength{\abovecaptionskip}{0.05cm}
% \footnotesize
% \scalebox{0.93}{
% % \setlength{\tabcolsep}{0.34mm}{
% \centering
\centering
\resizebox{\textwidth}{!}
{
\begin{tabular}{lllccccccccc}
\toprule[2pt]
\multicolumn{3}{c}{\multirow{2}{*}{GOOD-Motif}} & \multicolumn{3}{c}{\multirow{2}{*}{tune set}} & \multicolumn{3}{c}{base} & \multicolumn{3}{c}{size}\\
\cmidrule(r){7-9} \cmidrule(r){10-12}
& & & & & & \multicolumn{1}{c}{covariate} & \multicolumn{1}{c}{concept} & \multicolumn{1}{c}{no shift} & \multicolumn{1}{c}{covariate} & \multicolumn{1}{c}{concept} & \multicolumn{1}{c}{no shift} \\
\midrule[1pt]

\multicolumn{3}{l}{ERM} & -- & -- & -- & -- & -- & -- & -- & -- & -- \\
\multicolumn{3}{l}{IRM} & 1.0 & 10.0 & 0.1 & 0.1 & 0.1 & 0.1 & 0.1 & 0.1 & 0.1 \\
\multicolumn{3}{l}{VREx} & 1000.0 & 100.0 & 10.0 & 1000.0 & 1000.0 & 1000.0 & 100.0 & 10.0 & 1000.0 \\
\multicolumn{3}{l}{GroupDRO} & 0.001 & 0.01 & 0.1 & 0.001 & 0.001 & 0.1 & 0.1 & 0.01 & 0.1 \\
\multicolumn{3}{l}{DANN} & 0.1 & 1.0 & 0.01 & 0.01 & 0.1 & 0.01 & 0.1 & 0.1 & 0.01 \\
\multicolumn{3}{l}{Deep Coral} & 1.0 & 0.1 & 0.01 & 1.0 & 0.1 & 0.1 & 1.0 & 0.01 & 0.1 \\
\multicolumn{3}{l}{Mixup} & 1.0 & 2.0 & 0.4 & 2.0 & 0.4 & 0.4 & 1.0 & 0.4 & 0.4 \\
\multicolumn{3}{l}{DIR} & 0.2 & 0.25 & 0.3 & 0.25 & 0.3 & 0.25 & 0.25 & 0.3 & 0.25 \\

\bottomrule[2pt]
\end{tabular}
}
% }
\caption{
OOD hyperparameter selections on GOOD-Motif.
}
\label{table:h5}
\end{table*}
\begin{table*}[!htbp]
% \setlength{\abovecaptionskip}{0.05cm}
% \footnotesize
% \scalebox{0.93}{
% % \setlength{\tabcolsep}{0.34mm}{
% \centering
\centering
\resizebox{\textwidth}{!}
{
\begin{tabular}{lllccccccccc}
\toprule[2pt]
\multicolumn{3}{c}{\multirow{2}{*}{GOOD-Cora}} & \multicolumn{3}{c}{\multirow{2}{*}{tune set}} & \multicolumn{3}{c}{word} & \multicolumn{3}{c}{degree}\\
\cmidrule(r){7-9} \cmidrule(r){10-12}
& & & & & & \multicolumn{1}{c}{covariate} & \multicolumn{1}{c}{concept} & \multicolumn{1}{c}{no shift} & \multicolumn{1}{c}{covariate} & \multicolumn{1}{c}{concept} & \multicolumn{1}{c}{no shift} \\
\midrule[1pt]

\multicolumn{3}{l}{ERM} & -- & -- & -- & -- & -- & -- & -- & -- & -- \\
\multicolumn{3}{l}{IRM} & 0.1 & 1.0 & 10.0 & 10.0 & 0.1 & 0.1 & 1.0 & 10.0 & 0.1 \\
\multicolumn{3}{l}{VREx} & 100.0 & 10.0 & 1.0 & 100.0 & 10.0 & 1.0 & 10.0 & 10.0 & 1.0 \\
\multicolumn{3}{l}{GroupDRO} & 0.001 & 0.01 & 0.1 & 0.1 & 0.01 & 0.01 & 0.1 & 0.01 & 0.01 \\
\multicolumn{3}{l}{DANN} & 0.01 & 0.1 & 0.001 & 0.001 & 0.1 & 0.001 & 0.01 & 0.1 & 0.001 \\
\multicolumn{3}{l}{Deep Coral} & 0.01 & 0.1 & 1.0 & 0.01 & 0.01 & 1.0 & 1.0 & 0.1 & 0.01 \\
\multicolumn{3}{l}{Mixup} & 0.4 & 2.0 & 1.0 & 0.4 & 2.0 & 0.4 & 0.4 & 2.0 & 2.0 \\
\multicolumn{3}{l}{EERM} & \multicolumn{3}{l}{1/3, 1e-2/5e-3/1e-2/1e-4} & 1,5e-3 & 3,1e-3 & 3,1e-2 & 3,5e-3 & 3,1e-3 & 3,1e-2 \\
\multicolumn{3}{l}{SRGNN} & 1e-5 & 1e-6 & 1e-4 & 1e-4 & 1e-5 & 1e-6 & 1e-5 & 1e-6 & 1e-6 \\

\bottomrule[2pt]
\end{tabular}
}
% }
\caption{
OOD hyperparameter selections on GOOD-Cora.
}
\label{table:h6}
\end{table*}
\begin{table*}[!htbp]
% \setlength{\abovecaptionskip}{0.05cm}
% \footnotesize
% \scalebox{0.93}{
% % \setlength{\tabcolsep}{0.34mm}{
% \centering
\centering
\resizebox{\textwidth}{!}
{
\begin{tabular}{lllccccccccc}
\toprule[2pt]
\multicolumn{3}{c}{\multirow{2}{*}{GOOD-Arxiv}} & \multicolumn{3}{c}{\multirow{2}{*}{tune set}} & \multicolumn{3}{c}{time} & \multicolumn{3}{c}{degree}\\
\cmidrule(r){7-9} \cmidrule(r){10-12}
& & & & & & \multicolumn{1}{c}{covariate} & \multicolumn{1}{c}{concept} & \multicolumn{1}{c}{no shift} & \multicolumn{1}{c}{covariate} & \multicolumn{1}{c}{concept} & \multicolumn{1}{c}{no shift} \\
\midrule[1pt]

\multicolumn{3}{l}{ERM} & -- & -- & -- & -- & -- & -- & -- & -- & -- \\
\multicolumn{3}{l}{IRM} & 0.1 & 1.0 & 10.0 & 0.1 & 1.0 & 1.0 & 0.1 & 1.0 & 1.0 \\
\multicolumn{3}{l}{VREx} & 1.0 & 100.0 & 10.0 & 100.0 & 1.0 & 100.0 & 1.0 & 100.0 & 1.0 \\
\multicolumn{3}{l}{GroupDRO} & 0.001 & 0.01 & 0.1 & 0.01 & 0.001 & 0.1 & 0.1 & 0.001 & 0.1 \\
\multicolumn{3}{l}{DANN} & 0.1 & 0.001 & 0.01 & 0.001 & 0.001 & 0.001 & 0.01 & 0.001 & 0.1 \\
\multicolumn{3}{l}{Deep Coral} & 0.1 & 0.01 & 1.0 & 0.1 & 1.0 & 1.0 & 0.1 & 1.0 & 0.1 \\
\multicolumn{3}{l}{Mixup} & 2.0 & 1.0 & 0.4 & 1.0 & 0.4 & 0.4 & 2.0 & 1.0 & 1.0 \\
\multicolumn{3}{l}{EERM} & -- & -- & -- & -- & -- & -- & -- & -- & -- \\
\multicolumn{3}{l}{SRGNN} & 1e-5 & 1e-6 & 1e-4 & 1e-6 & 1e-6 & 1e-6 & 1e-6 & 1e-5 & 1e-6 \\

\bottomrule[2pt]
\end{tabular}
}
% }
\caption{
OOD hyperparameter selections on GOOD-Arxiv.
}
\label{table:h7}
\end{table*}
\begin{table*}[!htbp]
% \setlength{\abovecaptionskip}{0.1cm}
% \footnotesize
% \scalebox{0.93}{
% % \setlength{\tabcolsep}{0.34mm}{
% \centering
\centering
\resizebox{0.7\textwidth}{!}
{
\begin{tabular}{lllcccccc}
\toprule[2pt]
\multicolumn{3}{c}{\multirow{2}{*}{GOOD-Twitch}} & \multicolumn{3}{c}{\multirow{2}{*}{tune set}} & \multicolumn{3}{c}{language}\\
\cmidrule(r){7-9}
& & & & & & \multicolumn{1}{c}{covariate} & \multicolumn{1}{c}{concept} & \multicolumn{1}{c}{no shift}\\
\midrule[1pt]

\multicolumn{3}{l}{ERM} & -- & -- & -- & -- & -- & -- \\
\multicolumn{3}{l}{IRM} & 10.0 & 0.1 & 1.0 & 10.0 & 1.0 & 0.1 \\
\multicolumn{3}{l}{VREx} & 100.0 & 10.0 & 1.0 & 100.0 & 100.0 & 10.0 \\
\multicolumn{3}{l}{GroupDRO} & 0.001 & 0.1 & 0.01 & 0.1 & 0.1 & 0.001 \\
\multicolumn{3}{l}{DANN} & 0.1 & 0.01 & 0.001 & 0.01 & 0.001 & 0.01 \\
\multicolumn{3}{l}{Deep Coral} & 0.01 & 1.0 & 0.1 & 0.01 & 0.01 & 0.1 \\
\multicolumn{3}{l}{Mixup} & 2.0 & 0.4 & 1.0 & 0.4 & 0.4 & 0.4 \\
\multicolumn{3}{l}{EERM} & \multicolumn{3}{l}{1/3, 1e-2/5e-3/1e-2/1e-4} & 1,1e-2 & 3,5e-3 & 3,1e-2 \\
\multicolumn{3}{l}{SRGNN} & 1e-5 & 1e-6 & 1e-4 & 1e-5 & 1e-6 & 1e-6 \\

\bottomrule[2pt]
\end{tabular}
}
% }
\caption{
OOD hyperparameter selections on GOOD-Twitch.
}
\label{table:h8}
\end{table*}
\begin{table*}[!htbp]
% \setlength{\abovecaptionskip}{0.1cm}
% \footnotesize
% \scalebox{0.93}{
% % \setlength{\tabcolsep}{0.34mm}{
% \centering
\centering
\resizebox{0.7\textwidth}{!}
{
\begin{tabular}{lllcccccc}
\toprule[2pt]
\multicolumn{3}{c}{\multirow{2}{*}{GOOD-WebKB}} & \multicolumn{3}{c}{\multirow{2}{*}{tune set}} & \multicolumn{3}{c}{university}\\
\cmidrule(r){7-9}
& & & & & & \multicolumn{1}{c}{covariate} & \multicolumn{1}{c}{concept} & \multicolumn{1}{c}{no shift}\\
\midrule[1pt]

\multicolumn{3}{l}{ERM} & -- & -- & -- & -- & -- & -- \\
\multicolumn{3}{l}{IRM} & 10.0 & 1.0 & 0.1 & 10.0 & 10.0 & 10.0 \\
\multicolumn{3}{l}{VREx} & 10.0 & 100.0 & 1.0 & 10.0 & 10.0 & 1.0 \\
\multicolumn{3}{l}{GroupDRO} & 0.01 & 0.001 & 0.1 & 0.001 & 0.01 & 0.1 \\
\multicolumn{3}{l}{DANN} & 0.001 & 0.01 & 0.1 & 0.001 & 0.01 & 0.001 \\
\multicolumn{3}{l}{Deep Coral} & 0.1 & 1.0 & 0.01 & 0.01 & 0.01 & 1.0 \\
\multicolumn{3}{l}{Mixup} & 0.4 & 1.0 & 2.0 & 2.0 & 0.4 & 2.0 \\
\multicolumn{3}{l}{EERM} & \multicolumn{3}{l}{1/3, 1e-2/5e-3/1e-2/1e-4} & 3,1e-3 & 3,5e-3 & 3,1e-3 \\
\multicolumn{3}{l}{SRGNN} & 1e-5 & 1e-6 & 1e-4 & 1e-6 & 1e-5 & 1e-4 \\

\bottomrule[2pt]
\end{tabular}
}
% }
\caption{
OOD hyperparameter selections on GOOD-WebKB.
}
\label{table:h9}
\end{table*}
\begin{table*}[!htbp]
% \setlength{\abovecaptionskip}{0.1cm}
% \footnotesize
% \scalebox{0.93}{
% % \setlength{\tabcolsep}{0.34mm}{
% \centering
\centering
\resizebox{0.7\textwidth}{!}
{
\begin{tabular}{lllcccccc}
\toprule[2pt]
\multicolumn{3}{c}{\multirow{2}{*}{GOOD-CBAS}} & \multicolumn{3}{c}{\multirow{2}{*}{tune set}} & \multicolumn{3}{c}{color}\\
\cmidrule(r){7-9}
& & & & & & \multicolumn{1}{c}{covariate} & \multicolumn{1}{c}{concept} & \multicolumn{1}{c}{no shift}\\
\midrule[1pt]

\multicolumn{3}{l}{ERM} & -- & -- & -- & -- & -- & -- \\
\multicolumn{3}{l}{IRM} & 10.0 & 1.0 & 0.1 & 10.0 & 1.0 & 10.0 \\
\multicolumn{3}{l}{VREx} & 100.0 & 1.0 & 10.0 & 100.0 & 100.0 & 1.0 \\
\multicolumn{3}{l}{GroupDRO} & 0.1 & 0.01 & 0.001 & 0.1 & 0.001 & 0.01 \\
\multicolumn{3}{l}{DANN} & 0.01 & 0.001 & 0.1 & 0.1 & 0.1 & 0.01 \\
\multicolumn{3}{l}{Deep Coral} & 0.01 & 0.1 & 0.001 & 0.01 & 0.001 & 0.01 \\
\multicolumn{3}{l}{Mixup} & 0.4 & 1.0 & 2.0 & 0.4 & 2.0 & 0.4 \\
\multicolumn{3}{l}{EERM} & \multicolumn{3}{l}{1/3, 1e-2/5e-3/1e-2/1e-4} & 1,5e-3 & 1,1e-2 & 1,1e-2 \\
\multicolumn{3}{l}{SRGNN} & 1e-5 & 1e-6 & 1e-4 & 1e-5 & 1e-5 & 1e-6 \\

\bottomrule[2pt]
\end{tabular}
}
% }
\caption{
OOD hyperparameter selections on GOOD-CBAS.
}
\label{table:h10}
\end{table*}

\section{GOOD Usage Guidelines and Maintenance Schedule}\label{sec:F}
We provide the open-source GOOD project to reproduce all reported results and extend OOD datasets and algorithms. The GOOD project enables automatic dataset downloads, easy data loading, and handy start-up code to work with any GOOD dataset or method. Meanwhile, we provide various modular utilities for OOD method development.
Reproduction is available and effortless with given test scripts and automatic re-loading of our best checkpoints. Please refer to our GitHub repository for installation details, along with more documentation and usage information at {\url{https://github.com/divelab/GOOD/}}. The code of GOOD uses the GPL3.0 license, while the datasets follow the MIT license. Please refer to the GOOD GitHub repository for license details.

We provide simple and standardized examples for dataset loading and training/evaluation procedures. 

\subsection{GOOD dataset loading}
Code listing~\ref{lst:pipeline} shows two ways to import a GOOD dataset and specify the domain selection and shift split. 

\subsection{GOOD taining/test pipeline}
Code listing~\ref{lst:pipline} provides a script to use the main function of the training/evaluation pipeline, following the three steps of loading the config, specifying the model, and executing the task.

\subsection{Maintenance schedule}
GOOD is maintained on GitHub, with CI tests hosted by CircleCI.
We welcome public use of the community. Any issues or discussions regarding technical or other concerns can be submitted to the GitHub repository, and we will reply as soon as possible.
GOOD benchmark is a growing project and expects to include more datasets, splits, and methods along with the development of the field.
We expect to include more methods in future work, especially graph-related ones. We will also include datasets and domain selections of a larger quantity and variety.
In addition, the current benchmark does not consider link prediction tasks~\cite{zhou2022ood}, which will be added as the project develops.

GOOD provides simple APIs for loading OOD algorithms, graph neural networks, and datasets, taking only several lines of code to start.
The full OOD split generalization code is provided for extensions and any new graph OOD dataset contributions. OOD algorithm base class can be easily overwritten to create new OOD methods.
In addition to playing as a package, GOOD is also an integrated and well-organized project ready to be further developed. All algorithms, models, and datasets can be easily registered by the register and automatically embedded into the designed pipeline without much effort. The only thing the user needs to do is write their own OOD algorithm class, model class, or new dataset class. Then they can compare their results with the leaderboard.
We provide insightful comparisons from multiple perspectives. Any research and studies can use our leaderboard results for comparison. Note that this is a growing project, so we will include new OOD algorithms gradually. Besides, we welcome researchers to include their algorithms in the leaderboard. We welcome and will assist with any contributions to this project. We expect GOOD as a graph OOD research, study, and development toolkit of easy use.

\begin{listing}[htbp]
\begin{tcolorbox}
\begin{minted}{Python}
# Directly import
from GOOD.data.good_datasets.good_hiv import GOODHIV
hiv_datasets, hiv_meta_info = GOODHIV.load(
        dataset_root, 
        domain='scaffold', 
        shift='covariate', 
        generate=False
)
# Or use register
from GOOD import register as good_reg
hiv_datasets, hiv_meta_info = good_reg.datasets['GOODHIV'].load(
        dataset_root, 
        domain='scaffold', 
        shift='covariate', 
        generate=False
)
cmnist_datasets, cmnist_meta_info = ood_reg.datasets['GOODCMNIST'].load(
        dataset_root, 
        domain='color', 
        shift='concept', 
        generate=False
)
\end{minted}
\end{tcolorbox}
\caption{\textbf{GOOD dataset loader}}
\label{lst:pipeline}
\end{listing}

\begin{listing}[htbp]
\begin{tcolorbox}
\begin{minted}{Python}
# Load a config
from GOOD import config_summoner
from GOOD.utils.args import args_parser
from GOOD.utils.logger import load_logger
args = args_parser()
config = config_summoner(args)
load_logger(config)

# Load a GNN, a dataloader, and an OOD algorithm
from GOOD.kernel.pipeline import initialize_model_dataset
from GOOD.ood_algorithms.ood_manager import load_ood_alg
model, loader = initialize_model_dataset(config)
ood_algorithm = load_ood_alg(config.ood.ood_alg, config)

# Start training
from GOOD.kernel.train import train
train(model, loader, ood_algorithm, config)
# Or start a test
from GOOD.kernel.evaluation import evaluate
test_stat = evaluate(model, loader, ood_algorithm, 'test', config)
\end{minted}
\end{tcolorbox}
\caption{\textbf{GOOD taining/test pipeline}}
\label{lst:pipline}
\end{listing}

\end{document}